\documentclass[10pt,journal,cspaper,compsoc]{IEEEtran}
\usepackage{epsfig}
\usepackage{graphicx}
\usepackage{amsmath}
\usepackage{amssymb}
\usepackage{epstopdf}
\usepackage{tablefootnote}
\usepackage{caption}
\usepackage{subcaption}
\usepackage[font={small}]{caption}
\usepackage{wrapfig}
\usepackage{array}
\newcolumntype{C}[1]{>{\centering\let\newline\\\arraybackslash\hspace{0pt}}m{#1}}
\usepackage{adjustbox}
\newcolumntype{R}[2]{%
    >{\adjustbox{angle=#1,lap=\width-(#2)}\bgroup}%
    l%
    <{\egroup}%
}

\usepackage{multirow}

\ifCLASSOPTIONcompsoc
  \usepackage[nocompress]{cite}
\else
\fi

\begin{document}

\title{Shape Distributions of Nonlinear Dynamical Systems for Video-based Inference}

\author{Vinay~Venkataraman,~\IEEEmembership{Student Member,~IEEE},
        and Pavan~Turaga,~\IEEEmembership{Senior Member,~IEEE}
\IEEEcompsocitemizethanks{\IEEEcompsocthanksitem V. Venkataraman and P. Turaga are with the School of Electrical, Computer and Energy Engineering and School of Arts, Media
and Engineering, Arizona State University, Tempe, USA.\protect\\

E-mail: vvenka18@asu.edu and pturaga@asu.edu 


}
\thanks{}}

\IEEEcompsoctitleabstractindextext{%
\begin{abstract}
This paper presents a shape-theoretic framework for dynamical analysis of nonlinear dynamical systems which appear frequently in several video-based inference tasks. Traditional approaches to dynamical modeling have included linear and nonlinear methods with their respective drawbacks. A novel approach we propose is the use of descriptors of the shape of the dynamical attractor as a feature representation of nature of dynamics. The proposed framework has two main advantages over traditional approaches: a) representation of the dynamical system is derived directly from the observational data, without any inherent assumptions, and b) the proposed features show stability under different time-series lengths where traditional dynamical invariants fail. We illustrate our idea using nonlinear dynamical models such as Lorenz and Rossler systems, where our feature representations (shape distribution) support our hypothesis that the local shape of the reconstructed phase space can be used as a discriminative feature. Our experimental analyses on these models also indicate that the proposed framework show stability for different time-series lengths, which is useful when the available number of samples are small/variable. The specific applications of interest in this paper are: 1) activity recognition using motion capture and RGBD sensors, 2) activity quality assessment for applications in stroke rehabilitation, and 3) dynamical scene classification. We provide experimental validation through action and gesture recognition experiments on motion capture and Kinect datasets. 
In all these scenarios, we show experimental evidence of the favorable properties of the proposed representation.
\end{abstract}

}

\maketitle

\IEEEdisplaynotcompsoctitleabstractindextext

\IEEEpeerreviewmaketitle

\section{Introduction}
\IEEEPARstart{D}{ynamical} modeling methods for understanding signals from various sensing platforms have been the cornerstone of many applications in the computer vision community, such as human activity analysis \cite{aggarwal2011human} and dynamical natural scene recognition \cite{movingvistas}. Recent advances in sensing platforms like motion capture systems and the Kinect have opened doors to several applications including home-based health monitoring, gaming and entertainment. Take for instance, the task of developing algorithms for understanding the dynamics in human activities. This problem is non-trivial due to the complexity of natural human movement, which is a result of interactions between multiple body joints having high degrees of freedom. In addition, the task of recognizing human actions is challenging due to several factors including inter-class similarities between actions (e.g., running and walking), intra-class variations due to multiple strategies for an action (e.g., dance) and inter-subject variations. Natural human movements (such as walking, running) are composed of periodic action sequences in the form of repetitions, with some variability \cite{stergiou2011human}. These inherent attributes of human movement (periodicity with variability) descriptive of a complex nonlinear chaotic system has motivated researchers to employ tools from nonlinear dynamical systems theory to model human movement \cite{ali2007chaotic,junejo2011view,stergiou2011human,perc2005dynamics,dingwell2000nonlinear,dingwell2007differences,harbourne2009movement,miller2006improved}. Dynamical modeling of spatio-temporal evolution of human activities are traditionally accomplished by defining a state space and learning a function that maps the current state to the next state \cite{ralaivola2003dynamical,bissacco2001recognition}. A recent alternate approach has attempted to derive a representation for the dynamical system directly from the observation data using tools from chaos theory \cite{ali2007chaotic}. The main idea here is that, by using a top-down approach of dynamical modeling, one would only approximate the true-dynamics of the system with attempts to fit a model to the observational data. Whereas, in the bottom-up approach \cite{ali2007chaotic}, the dynamical system parameters such as the number of independent variables, degrees of freedom and other unknown parameters are estimated from the data. Such an approach can be seen as a generalized representation without any strong assumptions, suitable for analyzing a wide range of dynamical phenomenon. 

\section{Related Work}
Several approaches have been proposed in literature for modeling the dynamics in an observed time-series and we list the prior works in the specific applications of interest in a) activity recognition, b) activity quality assessment, and c) natural scene recognition.

\subsection{Classical Dynamical Invariants}
The largest Lyapunov exponent is a widely used dynamical invariant (measure of chaos), which quantifies the rate of divergence of initially closely-spaced trajectories \cite{ali2007chaotic,movingvistas}. 
A practical method for estimating the largest Lyapunov exponent from an observational time-series was first proposed by Wolf \textit{et al}. \cite{wolf1985determining}. Several other approaches were also proposed in literature to \textit{quantify} chaos \cite{eckmann1985ergodic,sano1985measurement,farmer1987predicting}, which were found to suffer from at least one of these drawbacks: (a) unreliable for small datasets, (b) computationally intensive, (c) relatively difficult to implement \cite{lyaprosen}. An improved method for estimation of the largest Lyapunov exponent to overcome the above mentioned drawbacks was later proposed by Rosenstein \textit{et al}. \cite{lyaprosen}. However, experimental results on nonlinear dynamical models have shown that the suggested number of data samples for accurate estimation of the largest Lyapunov exponent is $10\textsuperscript{$m$}$ (where $m$ is the embedding dimension) \cite{tenbroek2007lyapunov,lyaprosen}. In recent years, these methods have been applied to model various visual dynamical phenomenon such as video-based recognition of human activities \cite{ali2007chaotic} as well as recognition of dynamical scenes \cite{movingvistas}. However, when one needs to make inferences from short videos, or for instance when the activity of interest lasts only a few seconds, the classical approaches have significant drawbacks. While quantification of chaos using the largest Lyapunov exponent have been used to monitor varying chaos levels (level of complexity of the system) for recognition or prediction purposes \cite{iasemidis2003adaptive}, experimental studies for modeling human activities have not reported any evidence for different levels of chaos in human activities. Hence, we believe that a representation for level of chaos may not be a suitable approach to model human activities. In this paper, we propose an alternative approach to model human activities by extracting dynamical features representative of the \textit{shape} of the reconstructed phase space instead of quantifying chaos. We also demonstrate through experiments that the framework for estimation of dynamical features show stability across different time-series lengths and compare the performance with traditional chaotic invariants. 

\subsection{Activity Recognition}
Human activity analysis has attracted the attention of many researchers providing extensive literature on the subject. A detailed review of the approaches in literature for modeling and recognition of human activities are discussed in \cite{aggarwal2011human,gavrila1999visual}. Since our present work is related to non-parametric approaches for dynamical system analysis for action modeling, we restrict our discussion to related methods. 

Human actions have been modeled using dynamical system theory in computer vision \cite{ali2007chaotic, bissacco2001recognition} and biomechanics \cite{dingwell2000nonlinear, perc2005dynamics, stergiou2011human}. Differential equations can be used to model such a system, which requires access to all independent variables of the system. This approach would facilitate an understanding of the system behavior and also allow for the prediction of future states using present and past state information. However, this is not realizable in practice, as it is extremely hard to determine the independent variables and the interactions governing the dynamics of human actions. 

Dynamical modeling of human actions can be broadly categorized into parametric and nonparametric methods. Furthermore, human actions have been modeled with the assumption that the underlying dynamical system is linear \cite{bissacco2001recognition} or nonlinear \cite{ali2007chaotic,ralaivola2003dynamical}. In parametric modeling approaches, the dynamics of a system is represented by imposing a model and learning the model parameters from training data. Hidden Markov Models (HMMs) \cite{rabiner1989tutorial} and Linear Dynamical Systems (LDSs) \cite{casti1986linear} are the most popular parametric modeling approaches employed for action recognition \cite{yamato1992recognizing,wilson1995learning,vaswani2005shape,cuntoor2007epitomic} and gait analysis \cite{kale2004identification,liu2006improved,bissacco2001recognition}. Nonlinear parametric modeling approaches like Switching Linear Dynamical Systems (SLDSs) have been utilized to model complex activities composed of sequences of short segments modeled by LDS \cite{bregler1997learning}. While, nonlinear approaches can provide a more accurate model, it is difficult to precisely learn the model parameters. In addition, one would only approximate the true-dynamics of the system with attempts to fit a model to the experimental data. An alternative nonparametric action modeling approach is based on tools from chaos theory, with no assumptions on the underlying dynamical system. Traditional chaotic measures, like the largest Lyapunov exponent, correlation dimension and correlation integral, have been extensively used to model human actions \cite{ali2007chaotic,dingwell2000nonlinear,perc2005dynamics,stergiou2011human}. However, \cite{lyaprosen} and \cite{tenbroek2007lyapunov} have shown that these nonlinear dynamical measures need large amounts of data to produce stable results ($10^m$, where $m$ is the embedding dimension). Junejo \textit{et al}. \cite{junejo2011view} used a self-similarity matrix, a graphical representation of distinct recurrent behavior of nonlinear dynamical systems, to learn an action descriptor. In this paper, through illustrative examples and experimental validation, we show that our framework works better than traditional chaotic invariants for action modeling. 

\subsection{Activity Quality for Stroke Rehabilitation}
Recently researchers from various backgrounds have shown interest in the development of computational frameworks for quantification of \textit{quality} of movement, for possible applications in health monitoring and rehabilitation \cite{chen2011computational,stergiou2011human,tenbroek2007lyapunov,venkataraman2013attractor}. Stroke being the most common neurological disorder, leaves millions disabled every year who are unable to undergo long-term therapy treatment due to insufficient coverage by insurance. Recent directions in rehabilitation research has been towards development of portable systems for therapy treatment. Traditional quantitative scales such as the Fugl Meyer Test \cite{fugl1975post} and the Wolf Motor Function Test (WMFT) \cite{wolf2001assessing}, have proven to be effective in evaluating movement quality. However, these approaches involve visual monitoring which would greatly benefit from the development of an objective computational framework for movement quality assessment. The aim here is to develop standardized methods to describe the level of impairment across subjects. We show the utility of the proposed action modeling framework for quantifying the quality of reaching tasks using a single marker on the wrist, and obtain comparable results to a heavy marker-based setup ($14$ markers placed on arm, shoulder and torso \cite{chen2011computational}).

The focus of existing approaches for movement quality assessment has been towards finding typical patterns in kinematics which differ between healthy and impaired subjects. While these approaches are successful in giving an insight into understanding human movement, they fail to utilize the inherent dynamical nature of the movement. Rehabilitation therapies are composed of repetitive movements (e.g., reach to a target) that are strongly periodic with inherent variability. Traditional methods have assumed that this variability arises from noise in the system. However, it is evident that variability is an integral part of repetitive movements due to the availability of multiple strategies for the movement. Also, it is believed that variability produced in human movement is a result of nonlinear interactions and have deterministic origin \cite{stergiou2011human}. Extensive research has been carried out to model this variability using nonlinear dynamical system theory \cite{dingwell2000nonlinear,perc2005dynamics,stergiou2011human}. In this paper, we utilize the action modeling framework for movement quality assessment using a single wrist marker.

\subsection{Natural Scene Classification}
Natural scene classification has been an active area of research in computer vision with applications in automated image and video understanding. Much research has been focused around scene classification using single still images \cite{fei2005bayesian,xiao2010sun}, thereby neglecting dynamical motion information available in videos. Recently, the problem of dynamical modeling of natural scenes was introduced by Shroff \textit{et al.} \cite{movingvistas} who utilized tools from chaos theory along with GIST \cite{oliva2006building,oliva2001modeling} to model the spatio-temporal evolution in natural scenes in an unconstrained setting. 

Dynamic texture representation using LDS proposed by Soatto \textit{et al}. have been used to recognize and synthesize dynamic textures such as sea-waves, smoke, traffic \cite{soatto2001dynamic,doretto2003dynamic}. Such low-dimensional models have been used to capture complex natural phenomena. However, experimental results reported in \cite{movingvistas} show that these simple models might not be effective for dynamic scene classification in an unconstrained setting. Shroff \textit{et al}. utilized traditional chaotic invariants to model the dynamics and have shown that dynamical attributes augmented with spatial attributes (GIST \cite{oliva2001modeling}) can be effectively used for categorization of dynamic scenes \cite{movingvistas}. Another recent approach utilized spatio-temporal oriented energy filters for dynamic natural scene classification \cite{derpanis2012dynamic}. In this paper, we test the generality of the proposed action modeling framework for dynamic scene classification application.

\textbf{Contributions:}
In this paper, we present a computational framework for analysis of dynamical systems by combining the theoretical concepts of dynamical system analysis and ideas in shape theory. 
We extract dynamical shape features from the reconstructed phase space in the form of shape distributions to achieve improved results. We show the utility of the proposed framework in action and gesture recognition, movement quality assessment and dynamical scene recognition and evaluate the performance by comparing it with traditional chaotic invariants. We also propose two new shape functions to encode local dynamical evolution as opposed to global shape functions proposed by Osada \textit{et al}. \cite{osada2002shape}.

\begin{figure*}
\centering
\begin{subfigure}[b]{0.3\textwidth}
	\includegraphics[width=\textwidth]{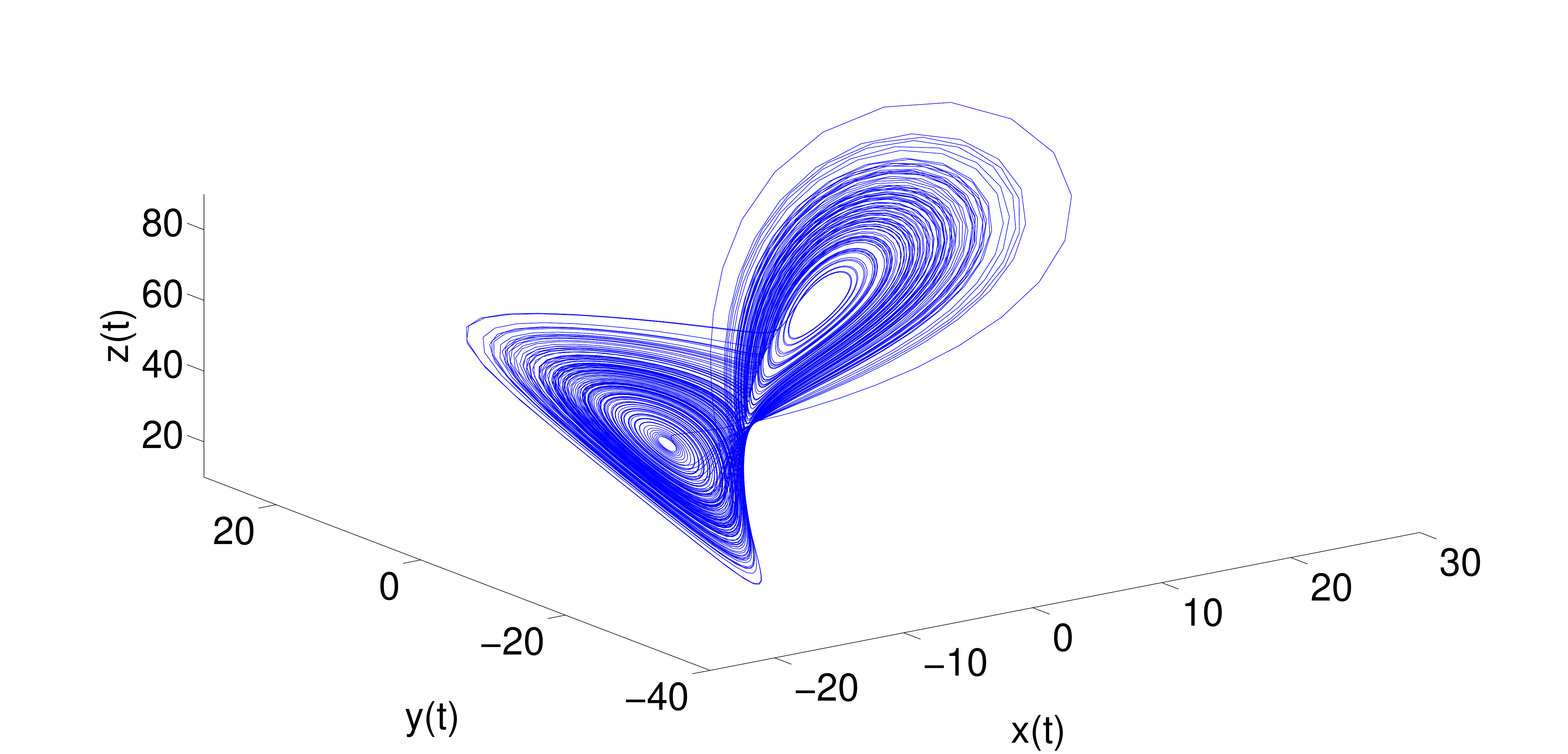}
	\caption{Lorenz attractor}
\end{subfigure}
\quad
\begin{subfigure}[b]{0.3\textwidth}
\centering
\includegraphics[width=\textwidth]{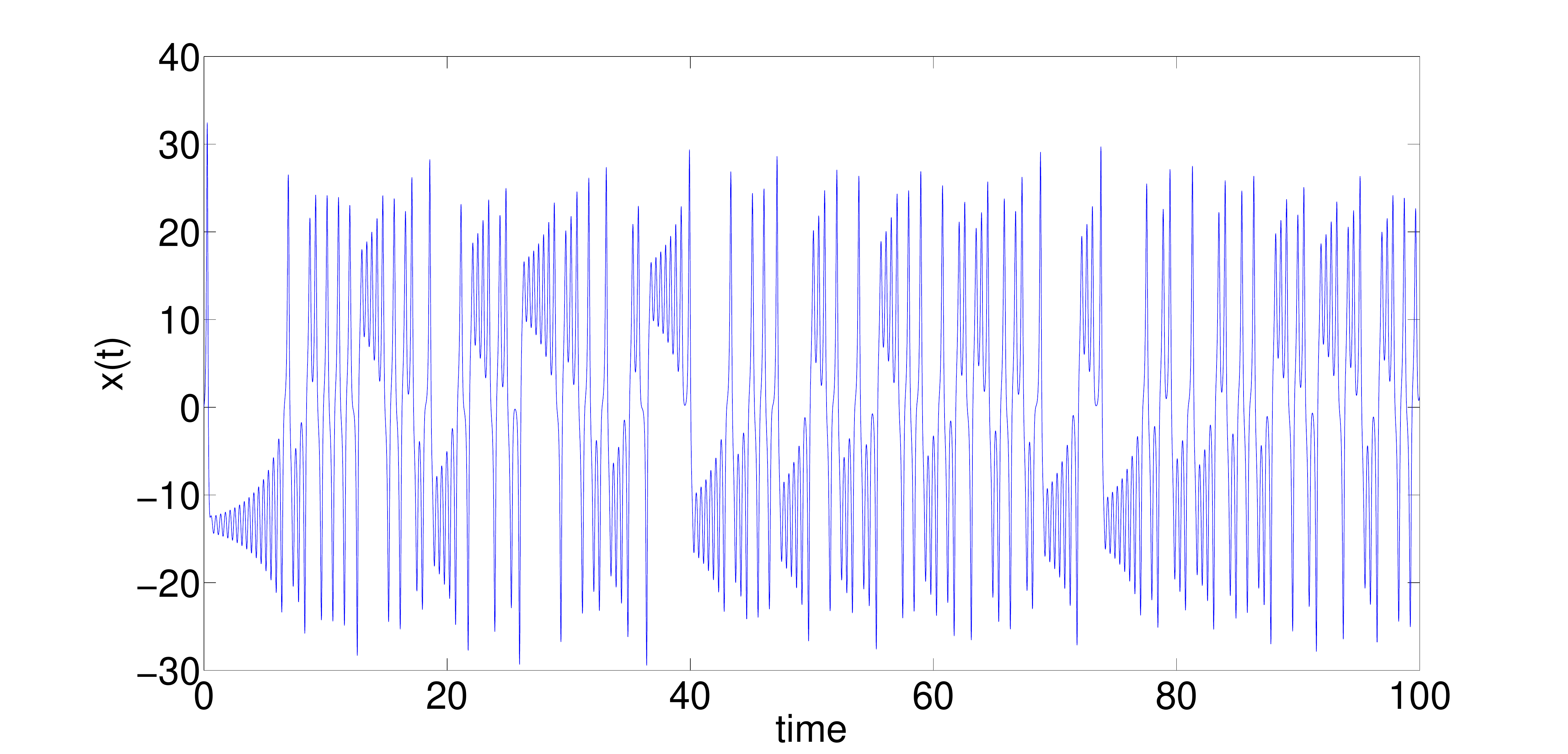}	
	\caption{One-dimensional time series \\* of Lorenz attractor (x(t))}
\end{subfigure}
\quad
\begin{subfigure}[b]{0.3\textwidth}
	\includegraphics[width=\textwidth]{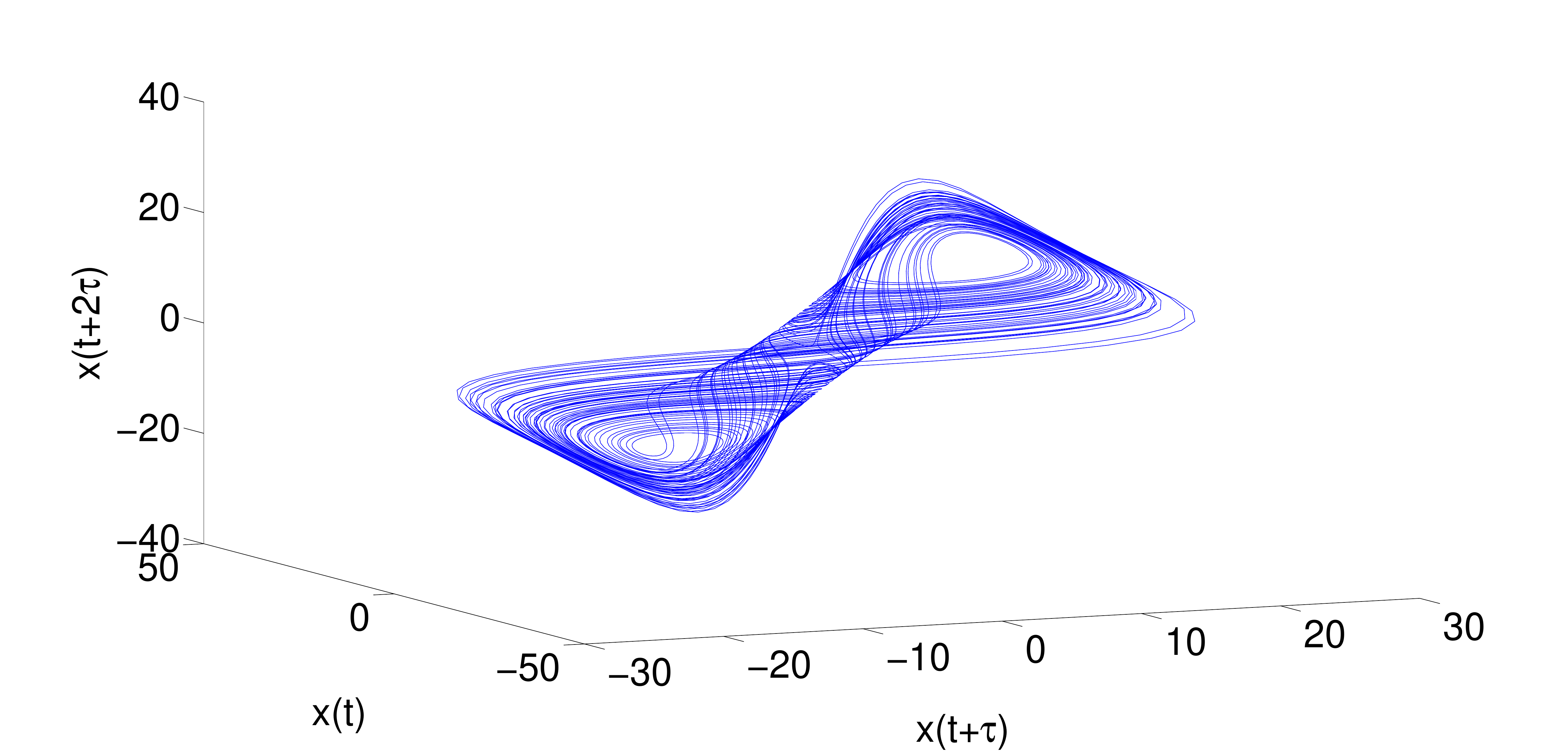}
	\caption{Reconstructed phase space by delay embedding}
\end{subfigure}
\caption{Phase space reconstruction of Lorenz attractor by delay embedding. (a) shows the $3$D view of trajectories of Lorenz attractor with control parameters $\rho = 45.92, \sigma = 16.0$ and $\beta = 4.0$. We can see that trajectories of Lorenz system settle down and are confined within the attractor. The one-dimensional time series (observed) of the Lorenz system is shown in (b). We see that a low-dimensional nonlinear system can generate such complex and chaotic signal. (c) shows the reconstructed phase space from observed time series of the Lorenz system using delay embedding ($\tau = 11$). The above example illustrates that the reconstructed phase space preserves certain topological properties of the original Lorenz attractor.}
\label{fig:lorenzattr}
\end{figure*}

\section{Preliminaries}
In this section, we introduce the background necessary to develop an understanding of nonlinear dynamical system analysis and chaos theory for applications in activity analysis, activity quality assessment and natural scene analysis.  

\subsection{Dynamical System Analysis}
Dynamical systems are governed by a set of functions defining the variations in the behavior of the system over time. A dynamical system is termed linear or nonlinear if the function defining the behavior of the system is linear or nonlinear respectively. 
Dynamical systems can be represented using state variables defining the state of the system at a given time $t$. A dynamical system is termed deterministic if there exists a unique future state for a given current state and is termed stochastic if the future state is derived from a probability distribution of possible states. Chaos theory is the field of study of such deterministic dynamical systems that show high sensitivity to initial conditions. A chaotic system is a dynamical system with deterministic behavior showing sensitivity to initial conditions. 

The states of a chaotic system are generally considered to be in an $n$-dimensional manifold also called \textit{phase space}. A chaotic system evolves over time in its phase space according to the system variables governing the dynamics. The path traversed by the system over time is called a \textit{trajectory} and the region of the phase space where the trajectories settle down as time approaches infinity is denoted as an \textit{attractor}. 

One would intend to have access to all independent variables of the system and their interactions for a complete understanding of the system. In a real world scenario, the data recorded is of low-dimension and is insufficient to model the dynamics of the system. In addition, model-based (parametric) approaches, such as LDS assume an underlying mapping function $f$ to describe the dynamics of the system. It has been established that such approaches may not be suitable for modeling the dynamics of complex systems such as human movements due to the simplifying assumptions \cite{bissacco2005modeling}. The theory of chaotic systems allows for determining certain invariants of the dynamical system function $f$ without making any assumptions about the system.

\subsection{Phase Space Reconstruction}
\label{sec:sec2}
The \textit{phase space} is defined as the space with all possible states of a system \cite{williams1997chaos,abarbanel1996analysis}. In a deterministic dynamical system that can be mathematically modeled, future states of the system can be determined using present and past state information. However, for applications such as human activity understanding and dynamical scene understanding, the system equations are complex. Furthermore, sensing systems in the real-world do not allow us to observe all variables of the system (e.g., the home-based setting for stroke rehabilitation with single marker on the wrist). To address these problems, we have to employ methods for reconstructing the attractor to obtain a phase space which preserves the important topological properties of the original dynamical system. This process is required to find the mapping function between the one-dimensional observed time series and the $m$-dimensional attractor, with the assumption that all variables of the system influence one another. The concept of phase space reconstruction was expounded in the embedding theorem proposed by Takens, called Takens' embedding theorem \cite{Takens} and an example of the procedure is shown in Fig. \ref{fig:lorenzattr}. For a discrete dynamical system with a multidimensional phase space, time-delay vectors (or embedding vectors) are obtained by concatenation of time-delayed samples given by 
\begin{equation}
\textbf{x}_{i}(n) = [x_{i}(n),x_{i}(n+\tau),\cdots,x_{i}(n+(m-1)\tau)]^T, 
\label{EmVec}
\end{equation}
where `$m$' is the embedding dimension and `$\tau$' is the embedding delay. These parameters should be carefully selected in order to facilitate a good phase space reconstruction. For a sufficiently large `$m$', the important topological properties of the unknown multidimensional system are reproduced in the reconstructed phase space \cite{abarbanel1996analysis}. The embedding method has proven to be useful, particularly for time series generated from low-dimensional deterministic dynamical systems, by providing a way to apply theoretical concepts of nonlinear dynamical systems onto observed time series. The embedding theorem does not suggest methods to estimate the optimal values for `$m$' and `$\tau$'. We use the false nearest neighbors \cite{kennel1992determining} approach to estimate $m$ and the first zero crossing of the autocorrelation function \cite{small2005applied} to estimate $\tau$. Fig. \ref{fig:lorenzattr} shows an example of phase space reconstruction from a one-dimensional observed time-series of a Lorenz system. 

\subsection{Embedding Dimension}
The embedding dimension refers to the number of time-delayed samples concatenated to form the time-delay vector (see (\ref{EmVec})). The aim here is to estimate an integer embedding dimension which can \textit{unfold} the attractor thereby removing any self-overlaps due to projection of the attractor onto lower dimensional space. Hence, the embedding dimension can be defined as the minimum dimension required to unfold the attractor completely. The false nearest neighbor approach finds this minimum embedding dimension to remove any \textit{false} nearest neighbors (neighbors due to projection onto lower dimension) \cite{abarbanel1996analysis}. Consider a vector in reconstructed phase space in dimension $m$ given by
\begin{subequations}
\begin{equation}
\textbf{x}(k) = [x(k),x(k+\tau),\cdots,x(k+(m-1)\tau)]^T, 
\end{equation}
and a nearest neighbor in the phase space given by 
\begin{equation}
\footnotesize
\textbf{x}^{NN}(k) = [x^{NN}(k),x^{NN}(k+\tau),\cdots,x^{NN}(k+(m-1)\tau)]^T. 
\end{equation}
\end{subequations}
If the vector $\textbf{x}^{NN}(k)$ is a true neighbor of $\textbf{x}(k)$, then it  should be because of the underlying dynamics. The vector $\textbf{x}^{NN}(k)$ can be a false neighbor of $\textbf{x}(k)$ when dimension $m$ is unable to unfold the attractor. Hence, moving to the next dimension $m+1$ may move this false neighbor out of the neighborhood of $\textbf{x}(k)$. This process of finding false neighbors to every vector $\textbf{x}_{i}(k)$ sequentially removes self-overlaps and identifies $m$ where the attractor is completely unfolded. The embedding dimension $m$ suggested by the false nearest neighbor algorithm for exemplar trajectories of human actions was either $3$ or $4$. We select a constant embedding dimension $m = 3$ to reconstruct all relevant phase space. Even with this fixed value of $m$, we obtain excellent results as shown in our experiments.

\subsection{Embedding Delay}
\label{TauCalc}
Embedding delay refers to the choice of integer time delay used to construct the time-delay vector. Theoretically, the embedding process allows any value of $\tau$ if one has access to infinitely accurate data (\cite{abarbanel1996analysis}, chap. 3). Since this is practically impossible, we try to find a value $\tau$ which makes the components of the vector [$x(k)$, $x(k+\tau)$, $x(k+2\tau)$]$^T$ in the embedding sufficiently independent. A low value of $\tau$ makes adjacent components to be correlated and hence they cannot be considered as independent variables. On the other hand, a high value of $\tau$ may make the adjacent components uncorrelated (almost independent) and cannot be considered as part of the system that supposedly generated them. The shape of the embedded time series will critically depend on the choice of $\tau$ \cite{small2005applied}. A good selection of $\tau$ should ensure that the data are maximally spread in phase space resulting in smooth phase space reconstruction. We use the first zero-crossing of the autocorrelation function as an estimate of $\tau$ as suggested in \cite{small2005applied} for strongly periodic data, which is a suitable choice for our experiments.  

\subsection{Phase Space Reconstruction of the Lorenz Attractor}
The Lorenz attractor is the steady state of a nonlinear chaotic system of three coupled nonlinear ordinary differential equations \cite{tucker1999lorenz} as given below:
\begin{subequations}
\begin{equation}
\dot{x} = \sigma(y-x),
\end{equation}
\begin{equation}
\dot{y} = x(\rho-z)-y,
\end{equation}
\begin{equation}
\dot{z} = xy-\beta z,
\end{equation}
\label{LorEq}
\end{subequations}
where $x$, $y$, $z$ are the state variables and $\sigma$, $\rho$ and $\beta$ are non-negative and dimensionless parameters. These equations were defined by Lorenz in $1963$ \cite{williams1997chaos} to represent a simplified model of thermal convection in the lower atmosphere. Lorenz showed that this relatively simple-looking set of equations could have highly erratic dynamics for a range of defined control parameters, for which the dynamics are chaotic. The dynamics of the Lorenz system in the $3$-dimensional state space generated from these set of equations is illustrated in Fig. \ref{fig:lorenzattr}(a). Lorenz attractor also illustrates that deterministic nonlinear models of low dimension can produce signal with complex dynamics. Furthermore, Fig. \ref{fig:lorenzattr} illustrates that it is possible to recreate an approximate attractor generated by a multidimensional system (such as Lorenz) using only a one-dimensional observed time series.

In the next section, we propose dynamical shape feature extraction from reconstructed phase space which is more suitable for action modeling than traditional chaotic invariants. We also show the stability of the proposed dynamical shape features for different time-series lengths using nonlinear dynamical models (Lorenz and Rossler systems).

\begin{figure*}
\centering
\begin{subfigure}[p]{0.3\textwidth}
\includegraphics[width=\textwidth]{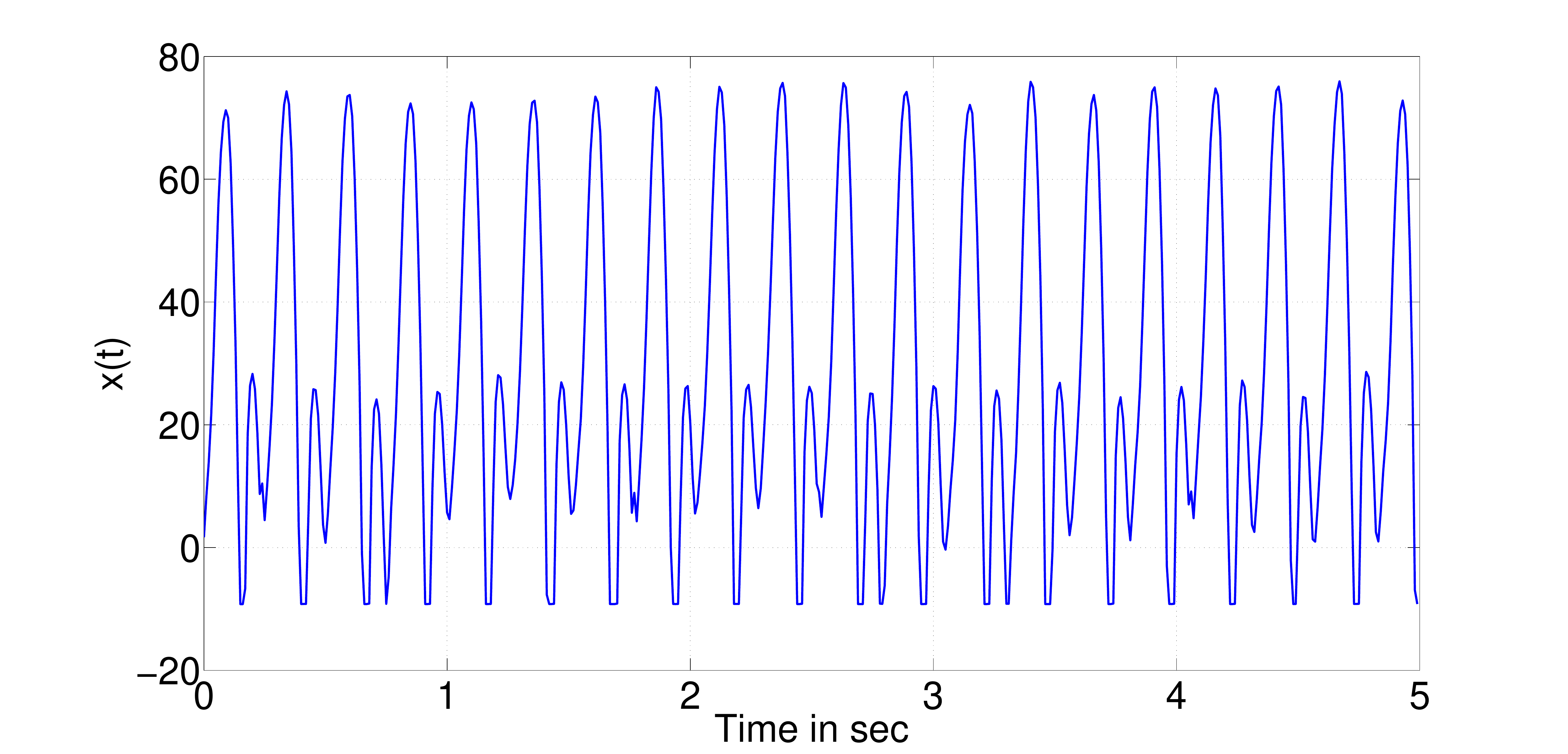}
\caption{Time series data}
\end{subfigure}
\begin{subfigure}[p]{0.3\textwidth}
\includegraphics[width=\textwidth]{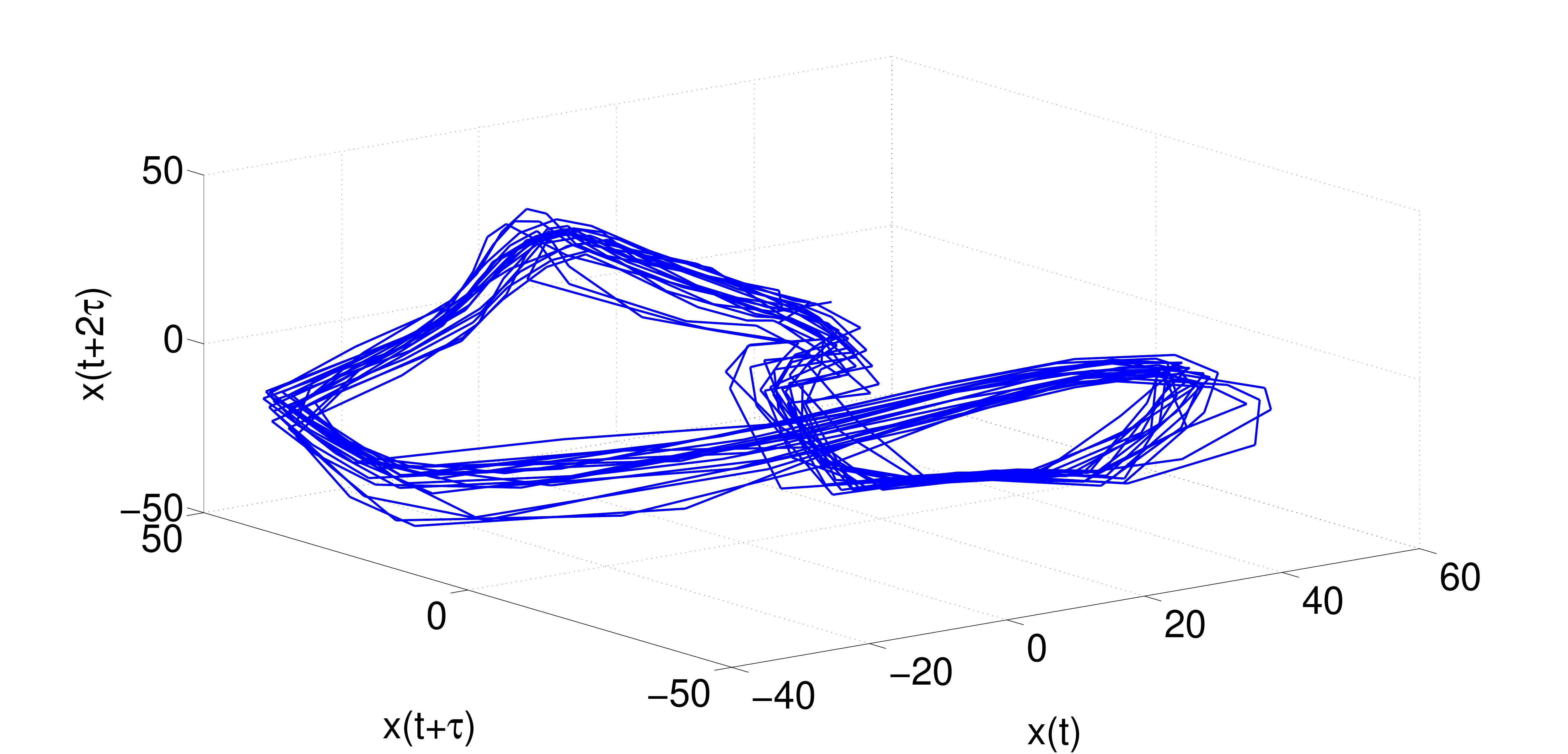}
\caption{Reconstructed Phase Space}
\end{subfigure}
\begin{subfigure}[p]{0.3\textwidth}
\includegraphics[width=\textwidth]{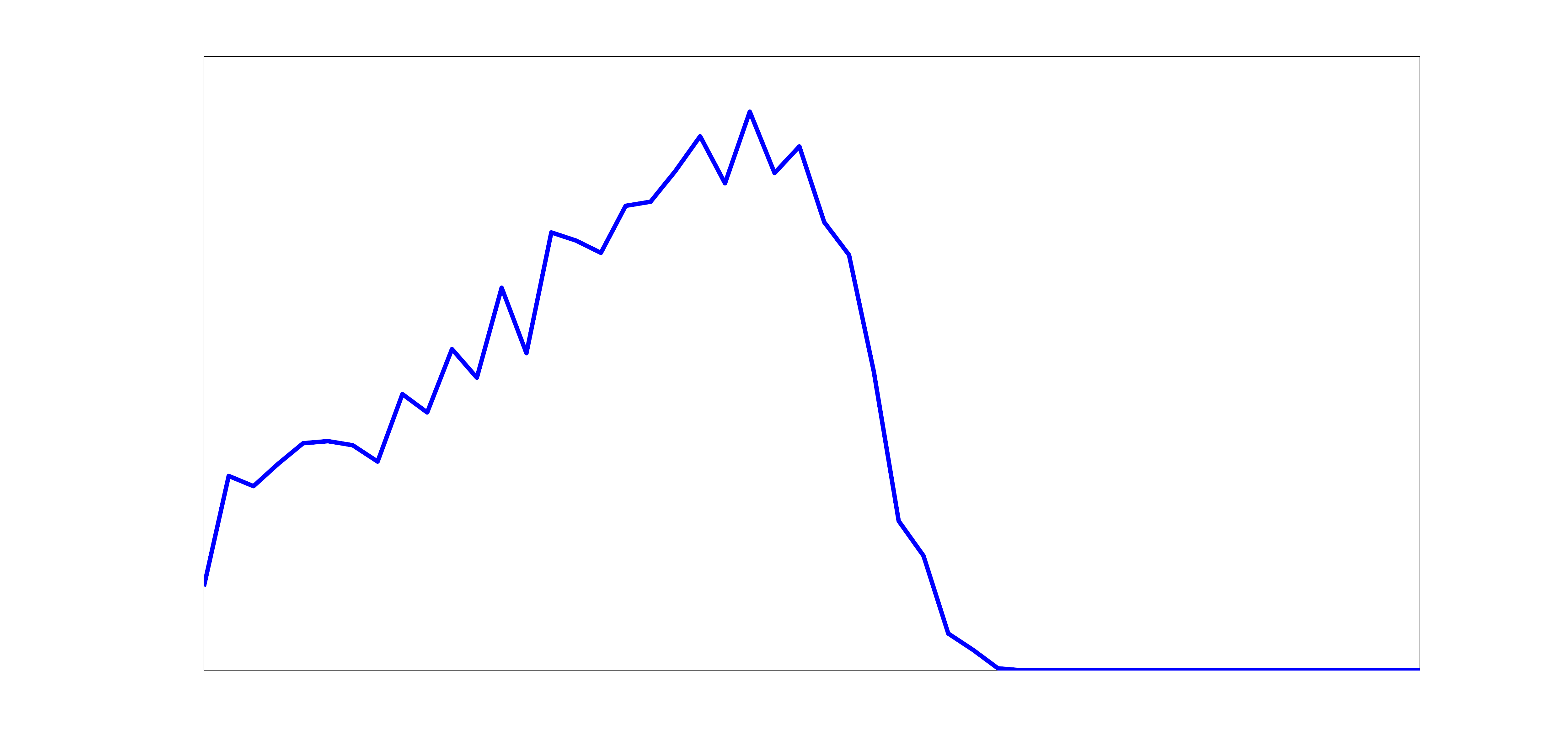}
\caption{Shape Distribution}
\end{subfigure}
\begin{subfigure}[p]{0.3\textwidth}
\includegraphics[width=\textwidth]{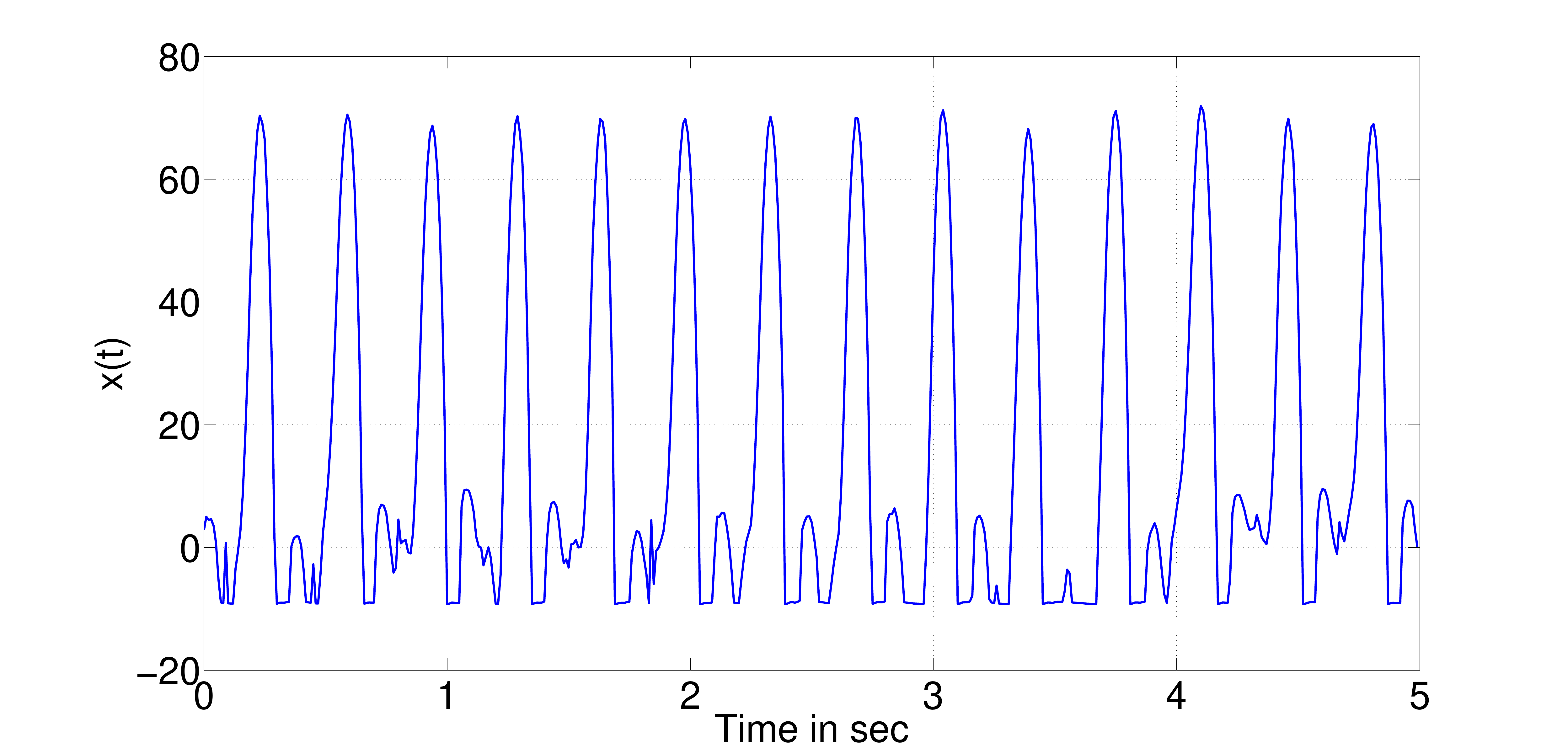}
\caption{Time series data}
\end{subfigure}
\begin{subfigure}[p]{0.3\textwidth}
\includegraphics[width=\textwidth]{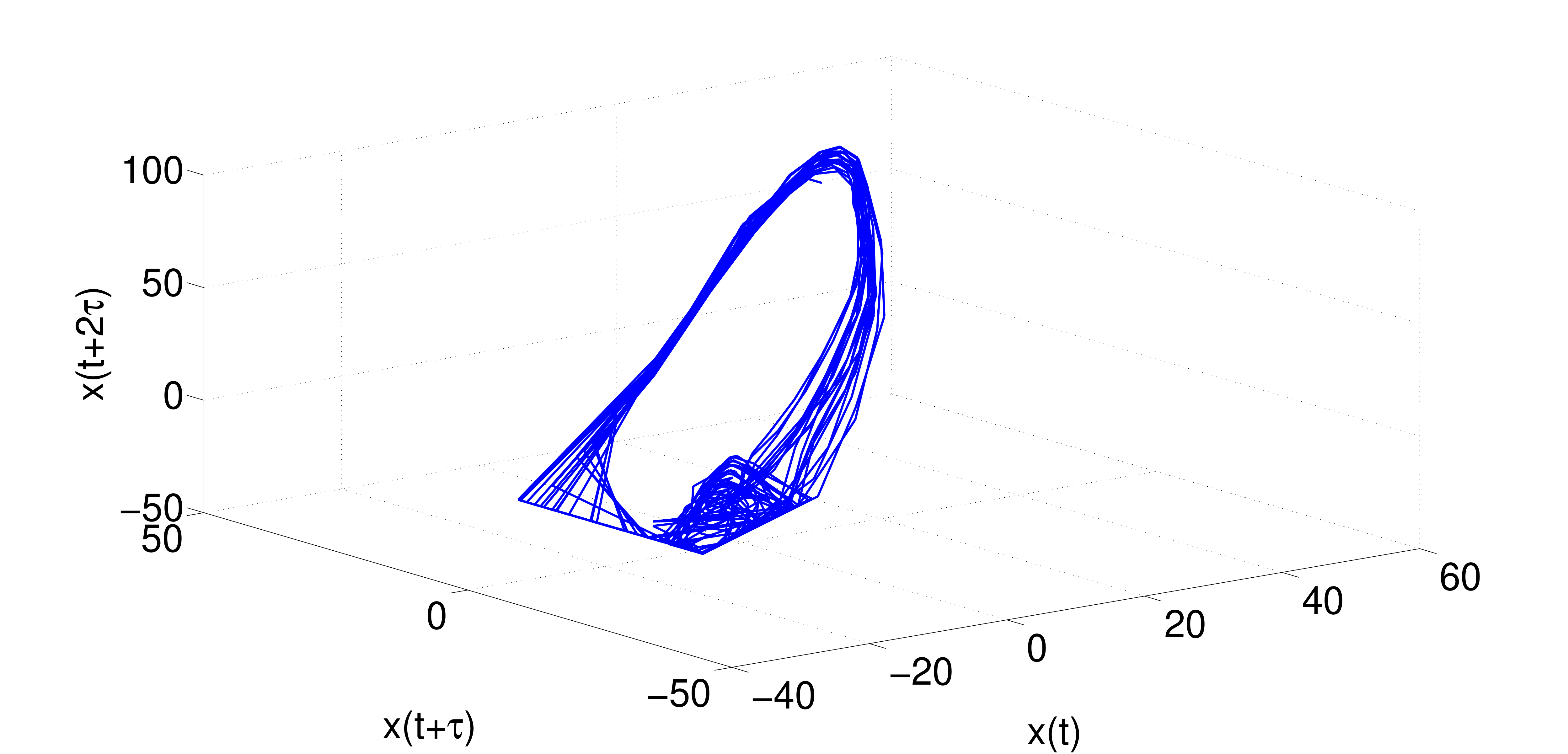}
\caption{Reconstructed Phase Space}
\end{subfigure}
\begin{subfigure}[p]{0.3\textwidth}
\includegraphics[width=\textwidth]{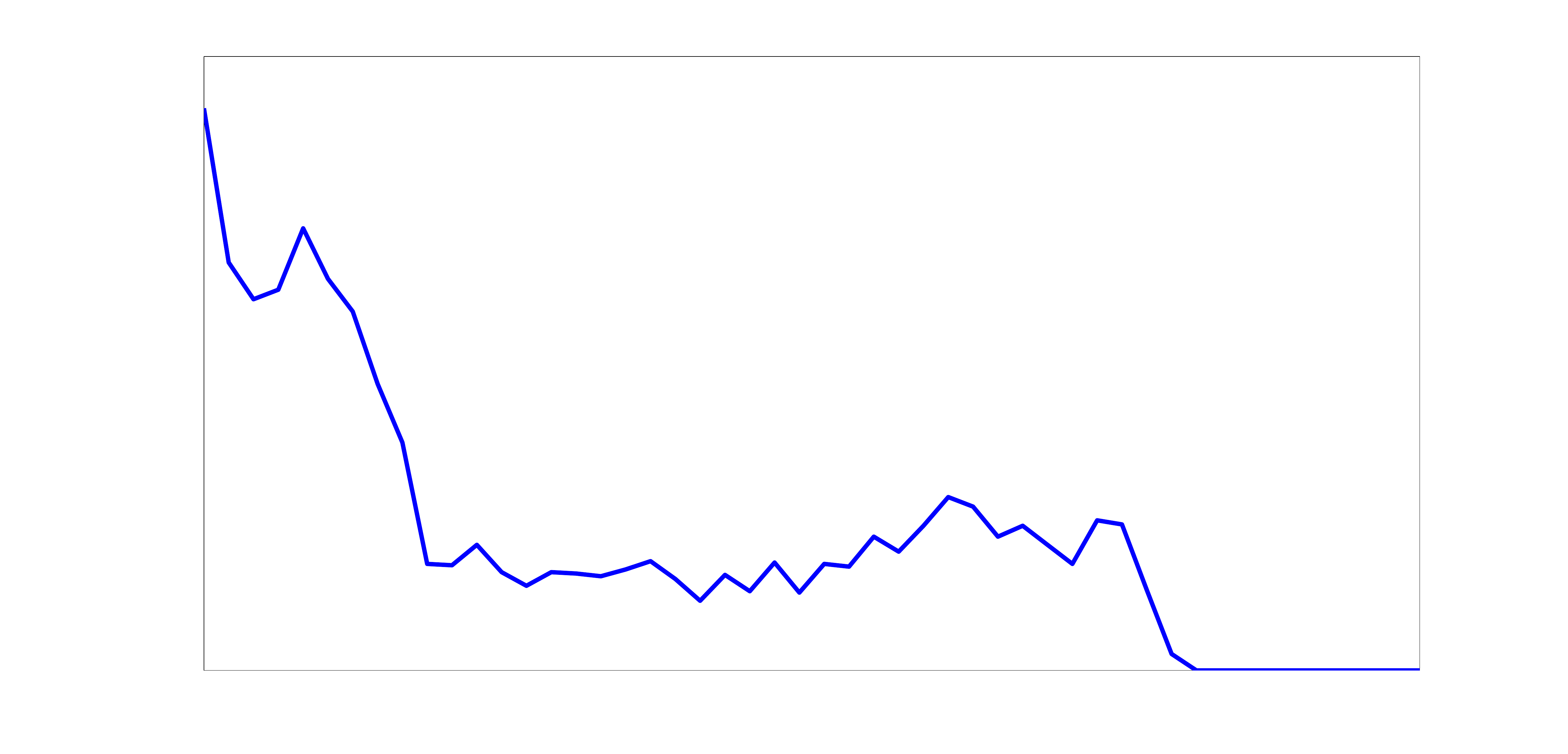}
\caption{Shape Distribution}
\end{subfigure}
\caption{Examples of phase space reconstruction of corresponding time series data of a subject performing \textit{Run} and \textit{Walk} action respectively. The embedding parameters were selected as $m=3$ and $\tau$ as described in section \ref{TauCalc}. This example illustrates that the $shape$ of the reconstructed phase space can be seen as a discriminative feature for classification of actions. We use shape distributions proposed by Osada \textit{et al}. \cite{osada2002shape} as a representation for shape of phase space. (c) and (f) together support our hypothesis that shape distribution (\textbf{D2}) can be used for classification of actions.}
\label{fig:ShapeofPhaseSpace}
\end{figure*}

\section{Attractor Shape Distributions}
In this section, we present a framework which combines the strong theoretical concepts of nonlinear dynamical analysis and ideas in shape theory to effectively represent the nature of dynamics. From Fig. \ref{fig:ShapeofPhaseSpace}, we see that the `\textit{shape}' of the reconstructed phase space can be seen as a discriminative feature for classification between $Run$ and $Walk$ action classes. Hence, our aim will be to extract feature representations for the shape of the reconstructed phase space. 
It is important to note here that the process of phase space reconstruction preserves certain topological properties and global shape is not a topological invariant, while local shape properties are. However, our goal here is to suggest a shape-based descriptor (both global and local) which possess sufficient discriminatory properties and robustness. 

We consider the attractor as having its own characteristic shape in the high-dimensional phase space. 
Shape analysis of $3$D surfaces is a well-studied problem in the computer vision community. In \cite{osada2002shape}, Osada \textit{et al}. present a method for finding a similarity measure between $3$D shapes by computing shape distributions of the $3$D surface sampled from the shape function by measuring their global geometric properties. We use the shape distribution of the reconstructed phase space as the dynamical feature representation in our experiments. While the shape distributions was originally proposed to measure similarity between $3$D shapes, we believe that shape distributions can be used as feature representations for any $n$-dimensional phase space. In addition, it is said that any function can be used to extract the shape distribution \cite{osada2002shape}, but we adopt simpler shape functions based on geometric properties (distance and area) which are listed below:\\
(a) \textit{Global Shape Functions}:

\begin{itemize}
  \item \textbf{D1}: measures the distance between one fixed point and one random point sampled from the reconstructed phase space. The fixed point is selected as the centroid of the attractor.
  \item \textbf{D2}: measures the distance between two random points in the phase space represented as ${||\textbf{x}_{i} - \textbf{x}_{j}||}_{2}$.
  \item \textbf{D3}: measures the square root of the area of the triangle formed by three random points on the attractor.
\end{itemize}

For example, the \textbf{D2} shape function can be represented as 
\begin{equation}
\textbf{D2}_{ij} = {||\textbf{x}_{i} - \textbf{x}_{j}||}_{2},
\end{equation}
where $\textbf{x}_{i}$ and $\textbf{x}_{j}$ are points (embedding vectors) in the reconstructed phase space. A set of these distances for randomly chosen embedding vector pairs are computed. From this set, we construct a histogram by counting the number of samples which fall into each of \textit{B}=$50$ fixed sized bins to obtain the attractor's shape distribution. 

These shape functions encode global geometric properties of the phase space, lacking information about local shape and dynamical evolution in the phase space. While previous investigation shows that global geometric shape function (\textbf{D2}) performs sufficiently better than the traditional nonlinear dynamical measures (largest Lyapunov exponent, correlation dimension and correlation integral) \cite{venkataraman2013attractor}, we hypothesize that a shape function which encodes local geometry and dynamical evolution information of phase space should improve the performance. In this direction, we propose new shape functions defined as, \\
(b) \textit{Local Shape Functions}:

\begin{itemize}
  \item \textbf{DT1}: It is similar to \textbf{D2}, with an additional constraint that the time separation between two random points in reconstructed phase space is $\le \delta$, thereby encoding only the local shape information. 
	
  \item \textbf{DT2}: encodes dynamical evolution of the phase space by exponential weighting given by
\begin{equation}
	\textbf{DT2}_{ij} = e^{- \gamma \left|t_i-t_j\right|} *{||\textbf{x}_{i} - \textbf{x}_{j}||}_{2},
\end{equation}
where $t_{i}$ and $t_{j}$ are the time indexes of the randomly selected pair of embedding vectors in the reconstructed phase space. `$\delta$' and `$\gamma$' are empirically determined parameters such that $\delta, \gamma \ge 0$.
\end{itemize}

\textbf{Local vs Global:} The main idea behind proposing these local shape functions is that, a global shape function would consider data samples from independent repetitions (well separated in time) of a movement. Also, repetitive human movements (such as \textit{running} and \textit{walking}) result in trajectories which wraps around itself in reconstructed phase space, creating an artifact of having closely spaced trajectories in phase space. We believe that such an approach would not provide a robust feature representation, and we suggest the use of local shape functions instead which only considers data samples close in time. 

\textbf{Metric on Shape Distributions:} Several metrics exist in literature to calculate the distance between histograms including chi-squared statistic ($\chi^{2}$ distance), Bhattacharyya distance \cite{bhattacharyya1943measure}, Riemannian analysis \cite{srivastava2007riemannian} and Earth Mover's Distance (EMD) \cite{rubner1998metric}. In our experiments, we provide results using Euclidean distance and chi-squared distance metrics for comparison due to their simplicity.

\begin{figure*}
\centering
\begin{subfigure}[p]{0.19\textwidth}
                \includegraphics[width=\textwidth]{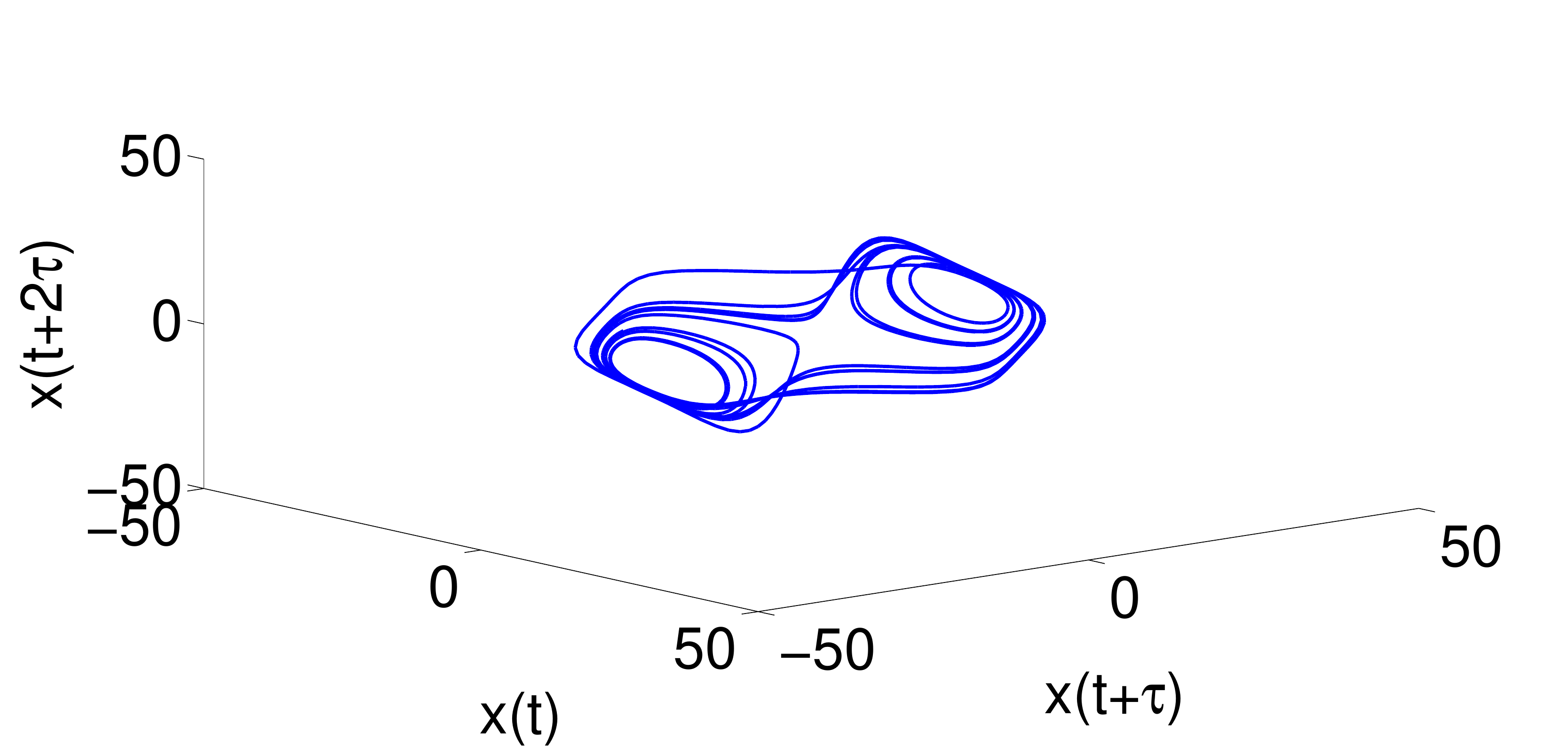}
        \end{subfigure}        
\begin{subfigure}[p]{0.19\textwidth}
                \includegraphics[width=\textwidth]{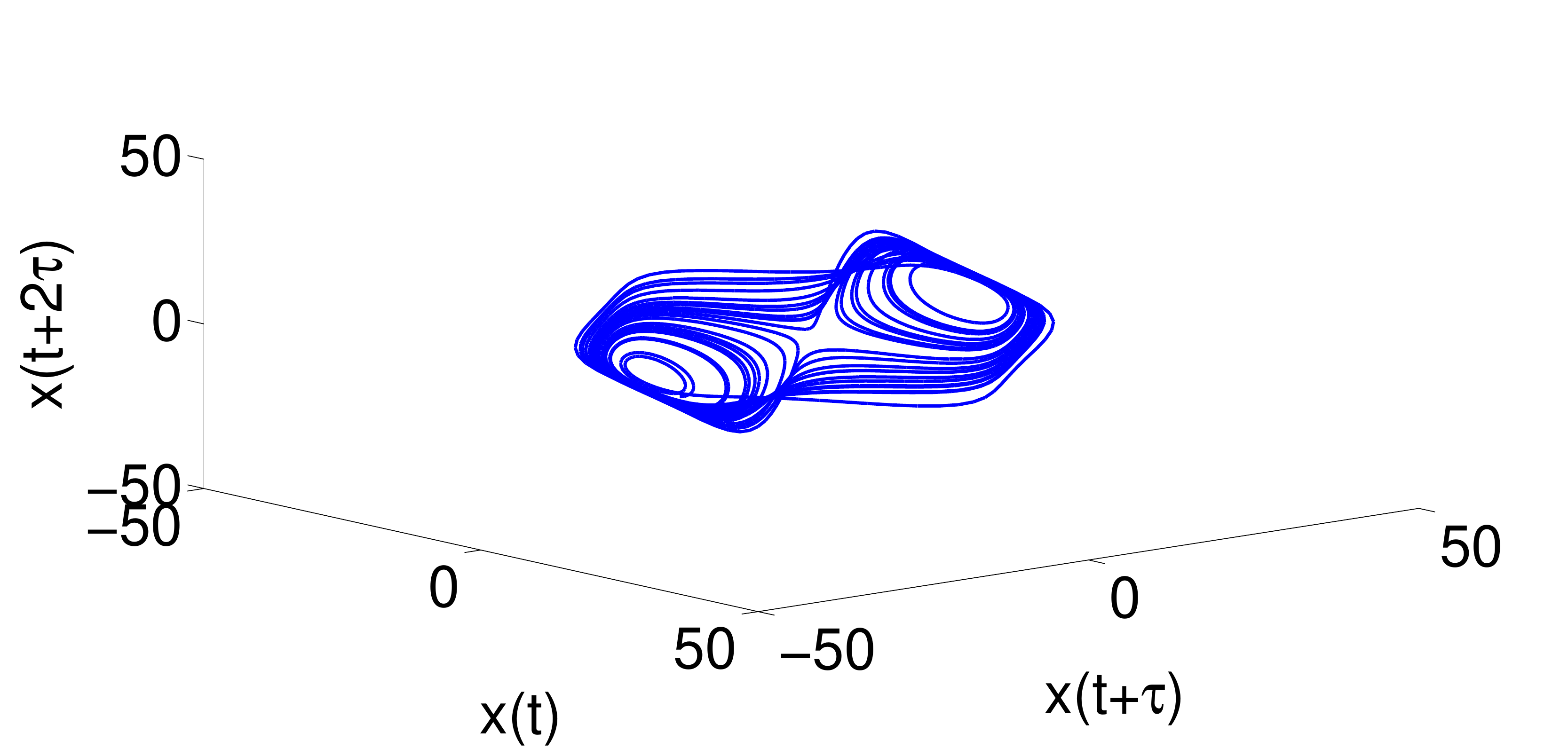}
        \end{subfigure}        
\begin{subfigure}[p]{0.19\textwidth}
                \includegraphics[width=\textwidth]{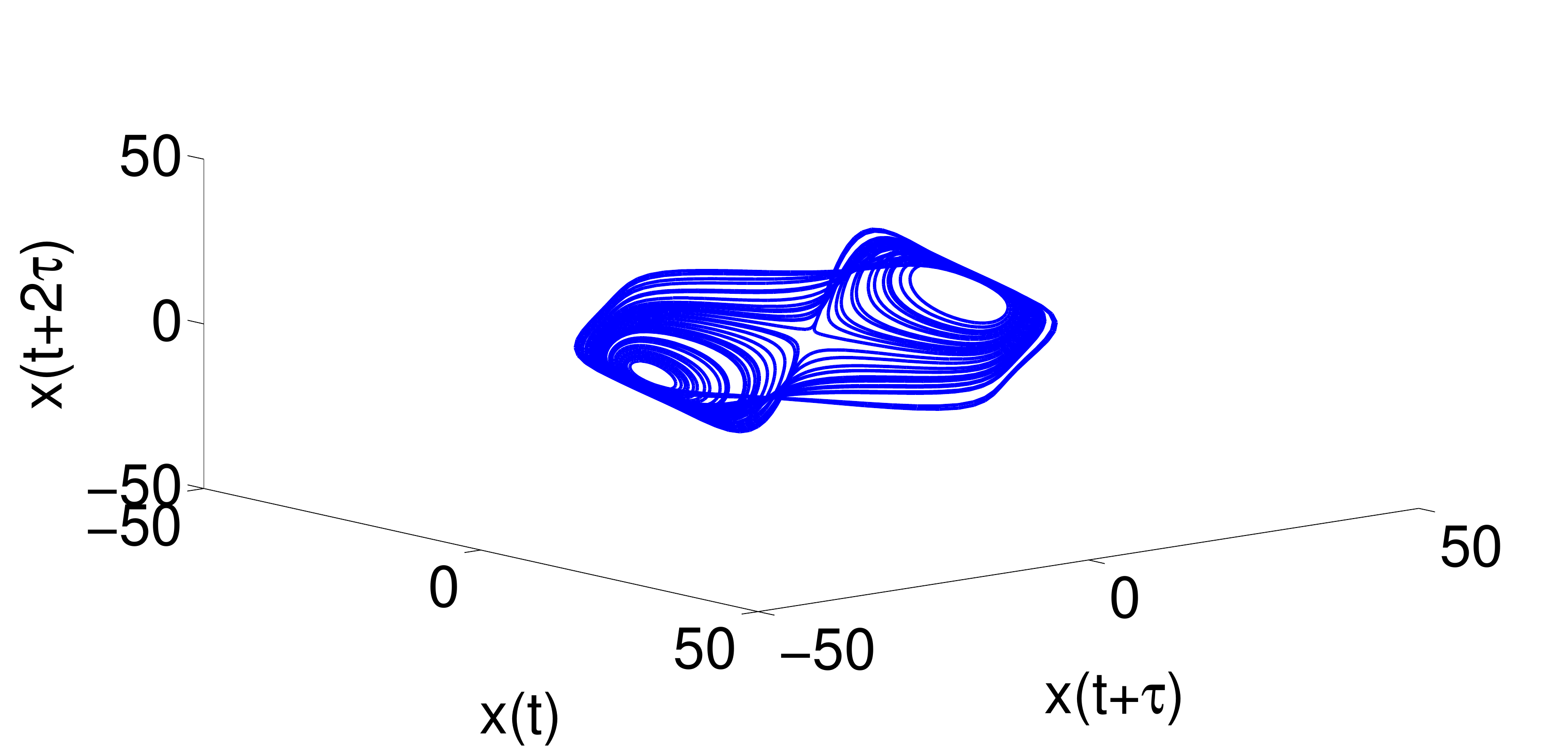}
        \end{subfigure}        
\begin{subfigure}[p]{0.19\textwidth}
                \includegraphics[width=\textwidth]{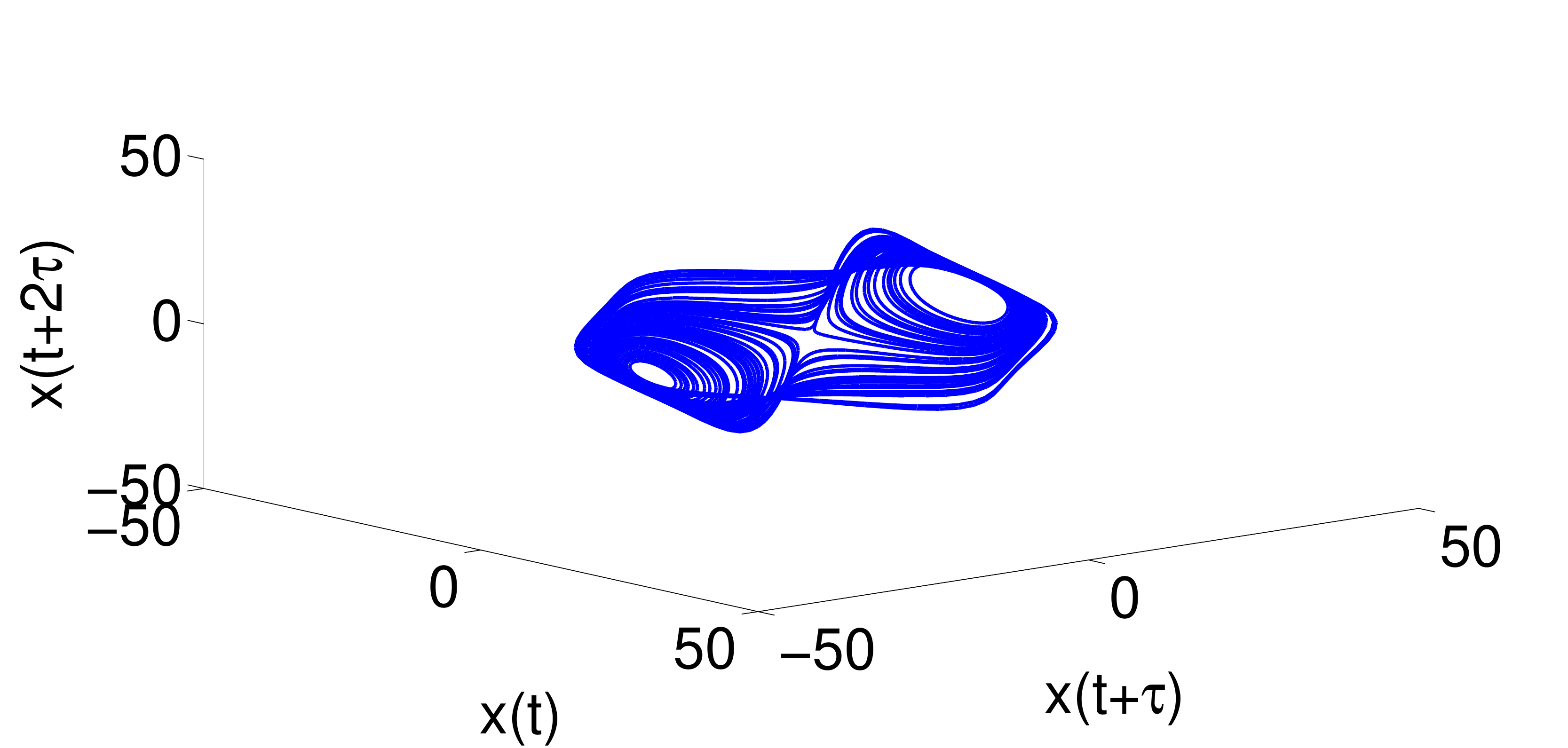}
        \end{subfigure}        
\begin{subfigure}[p]{0.19\textwidth}
                \includegraphics[width=\textwidth]{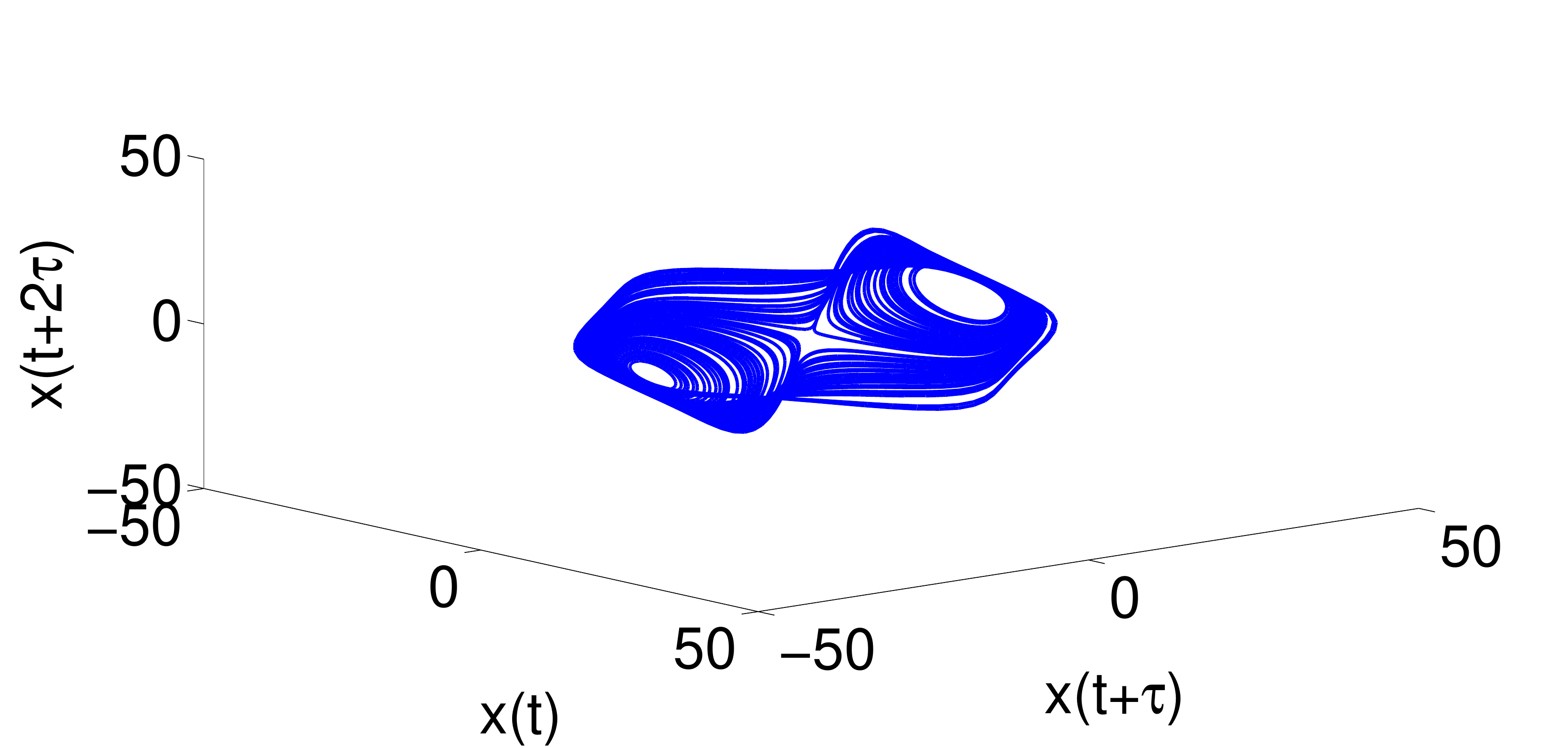}
        \end{subfigure}
        
                \vspace{13pt}        
        		\begin{picture}(0,0)
				\put(-150,0){{\mbox{\small Reconstructed phase space of Lorenz system for different time-series lengths}}}
				\end{picture}								
        \vspace{5pt}
        
\begin{subfigure}[p]{0.19\textwidth}
                \includegraphics[width=\textwidth]{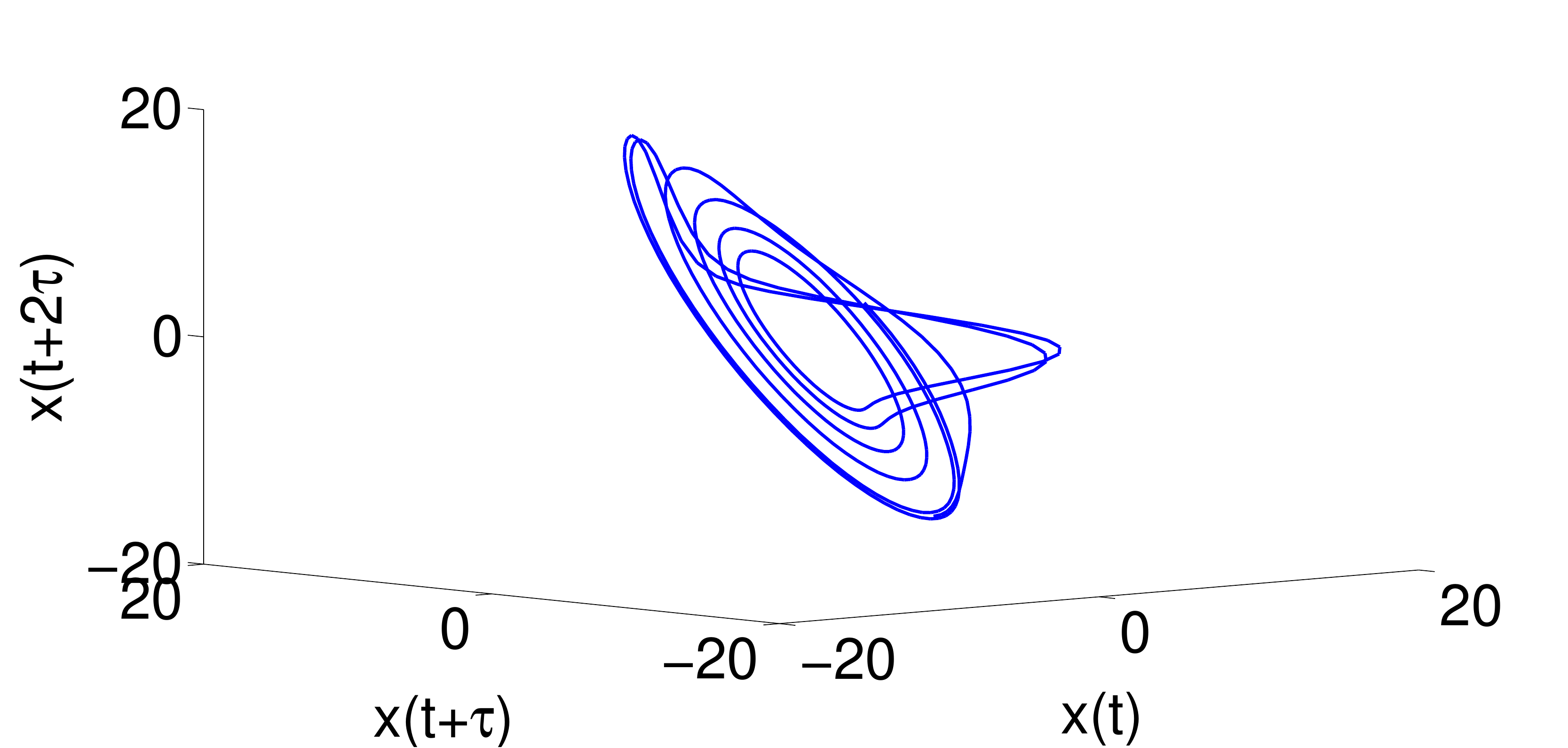}
        \end{subfigure}        
\begin{subfigure}[p]{0.19\textwidth}
                \includegraphics[width=\textwidth]{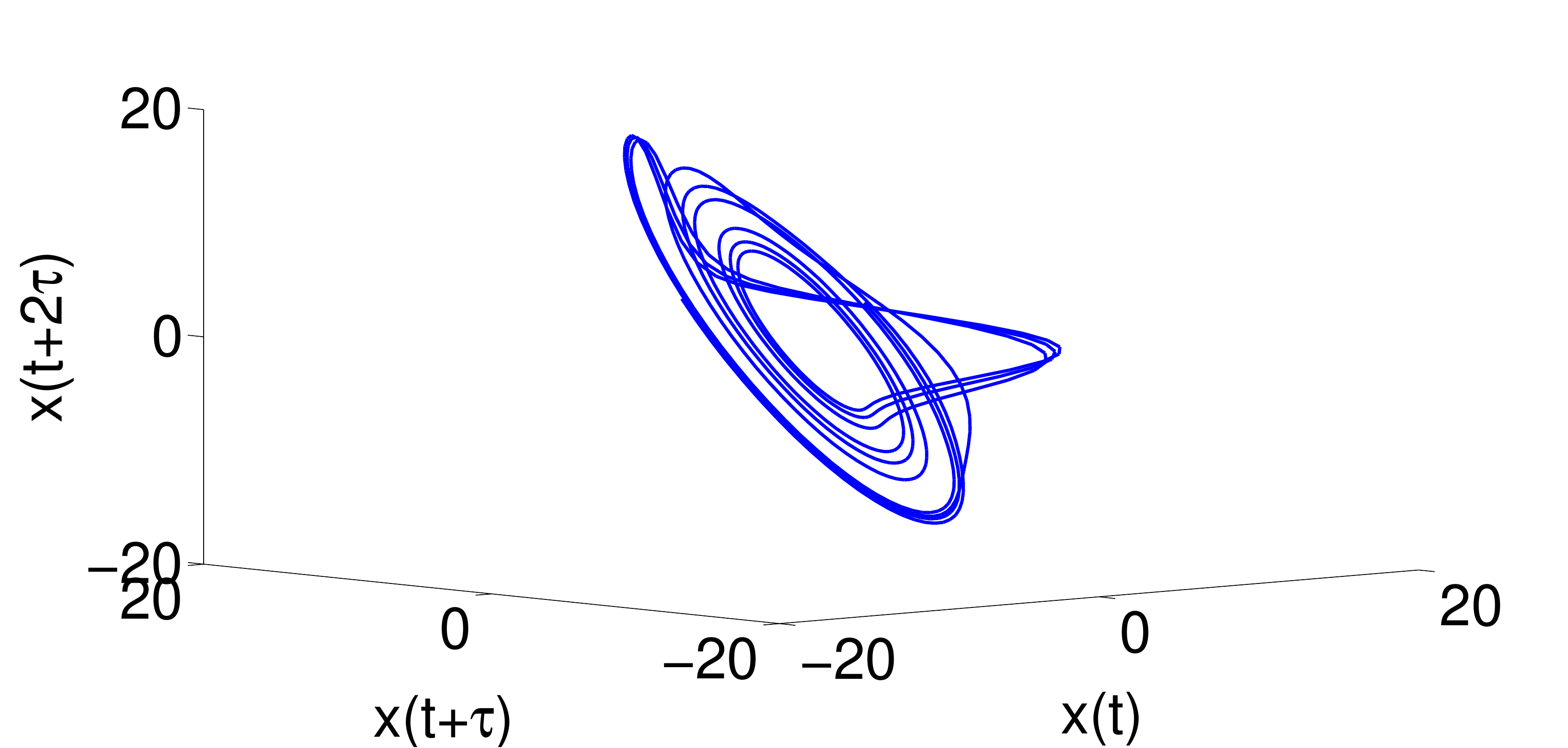}
        \end{subfigure}        
\begin{subfigure}[p]{0.19\textwidth}
                \includegraphics[width=\textwidth]{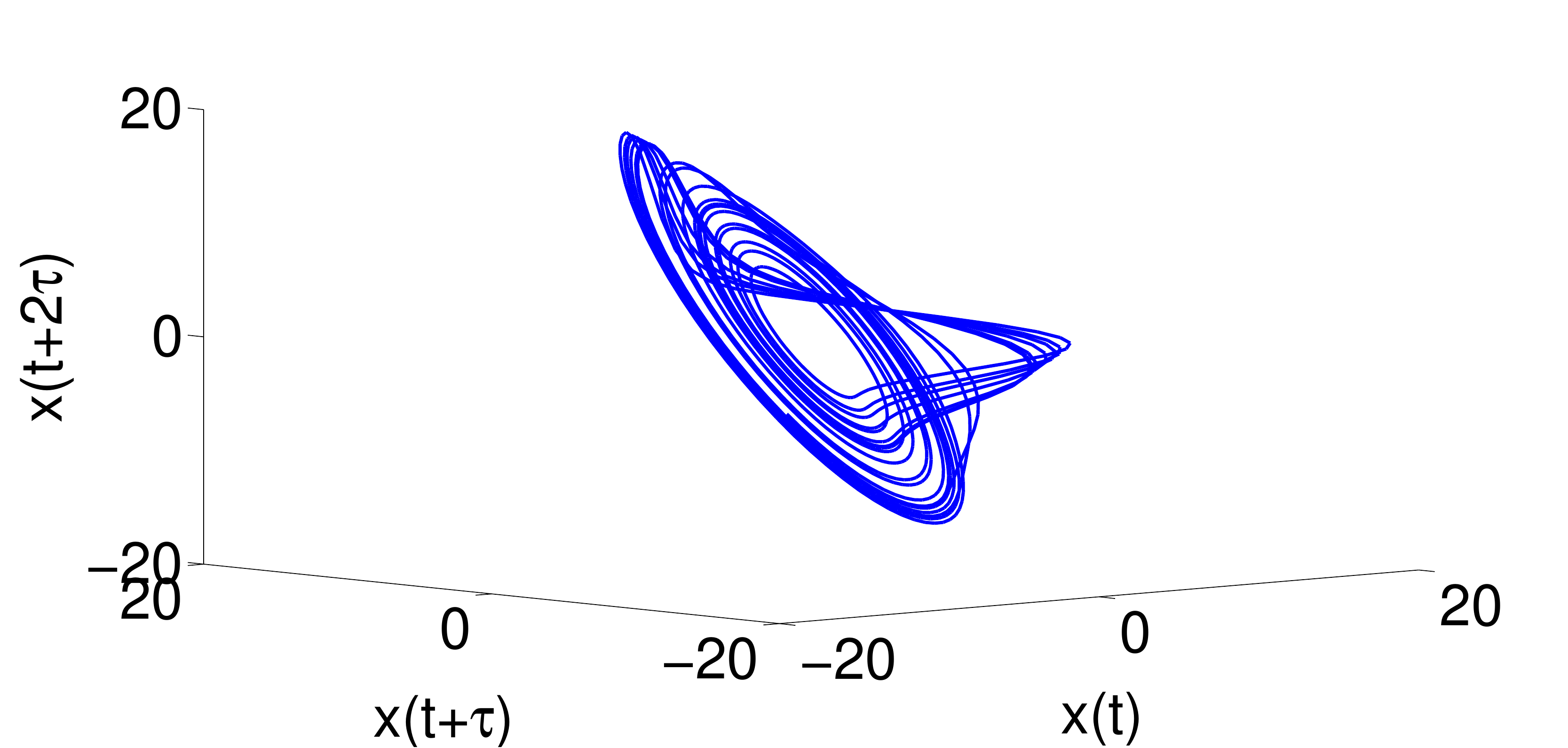}
        \end{subfigure}        
\begin{subfigure}[p]{0.19\textwidth}
                \includegraphics[width=\textwidth]{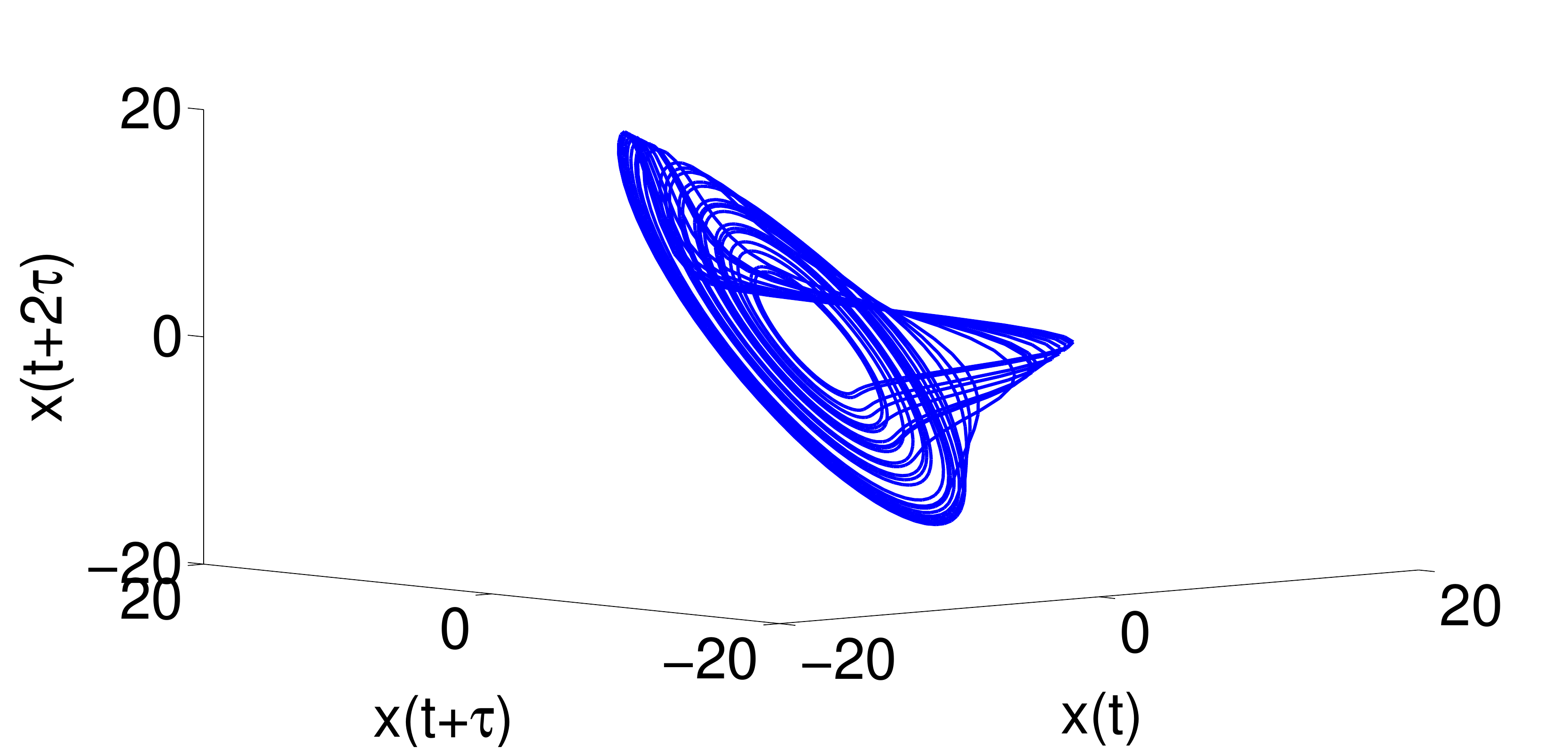}
        \end{subfigure}        
\begin{subfigure}[p]{0.19\textwidth}
                \includegraphics[width=\textwidth]{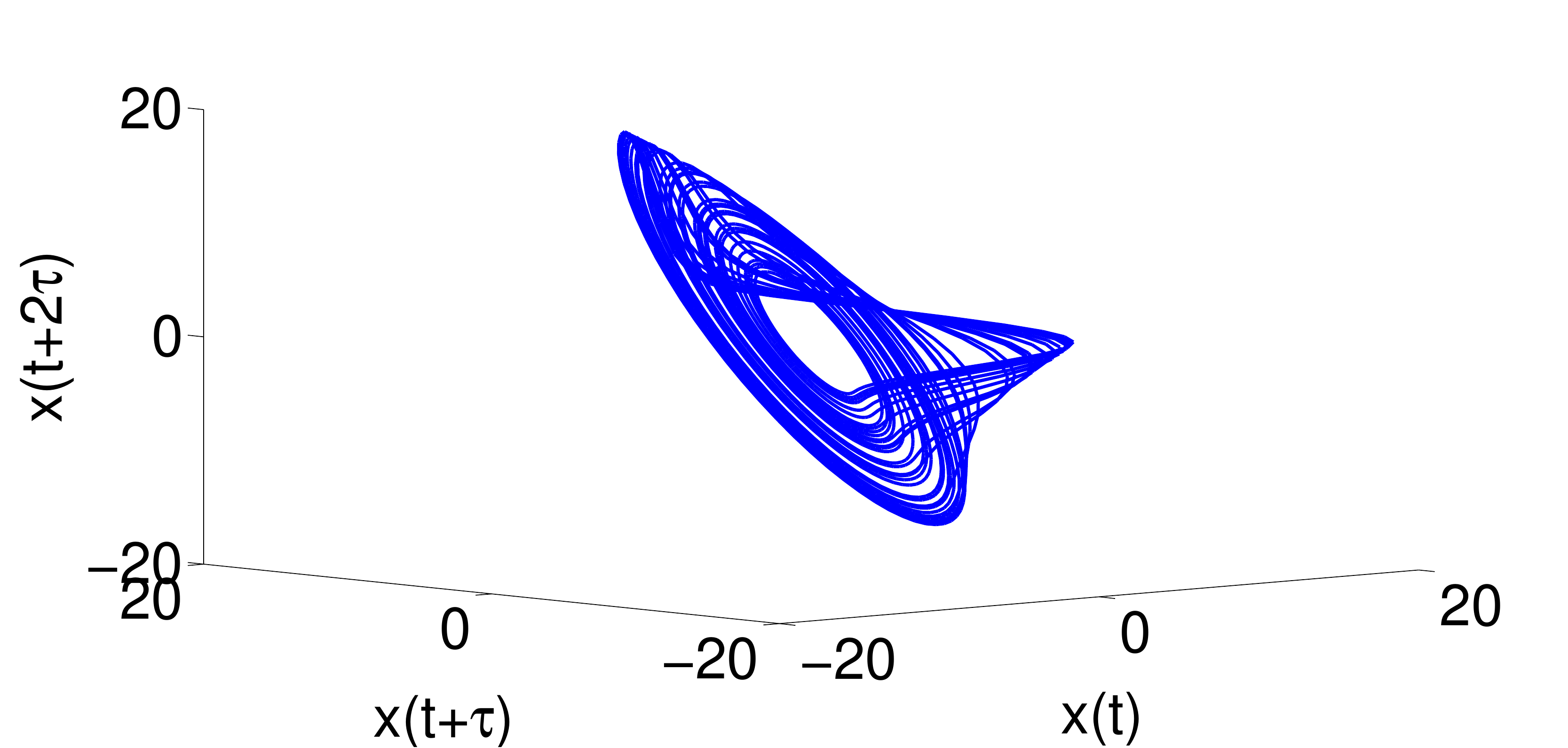}
        \end{subfigure}        
        
                \vspace{13pt}        
        		\begin{picture}(0,0)
				\put(-150,0){{\mbox{\small Reconstructed phase space of Rossler system for different time-series lengths}}}
				\end{picture}								
        \vspace{5pt}
        
\begin{subfigure}[p]{0.19\textwidth}
                \includegraphics[width=\textwidth]{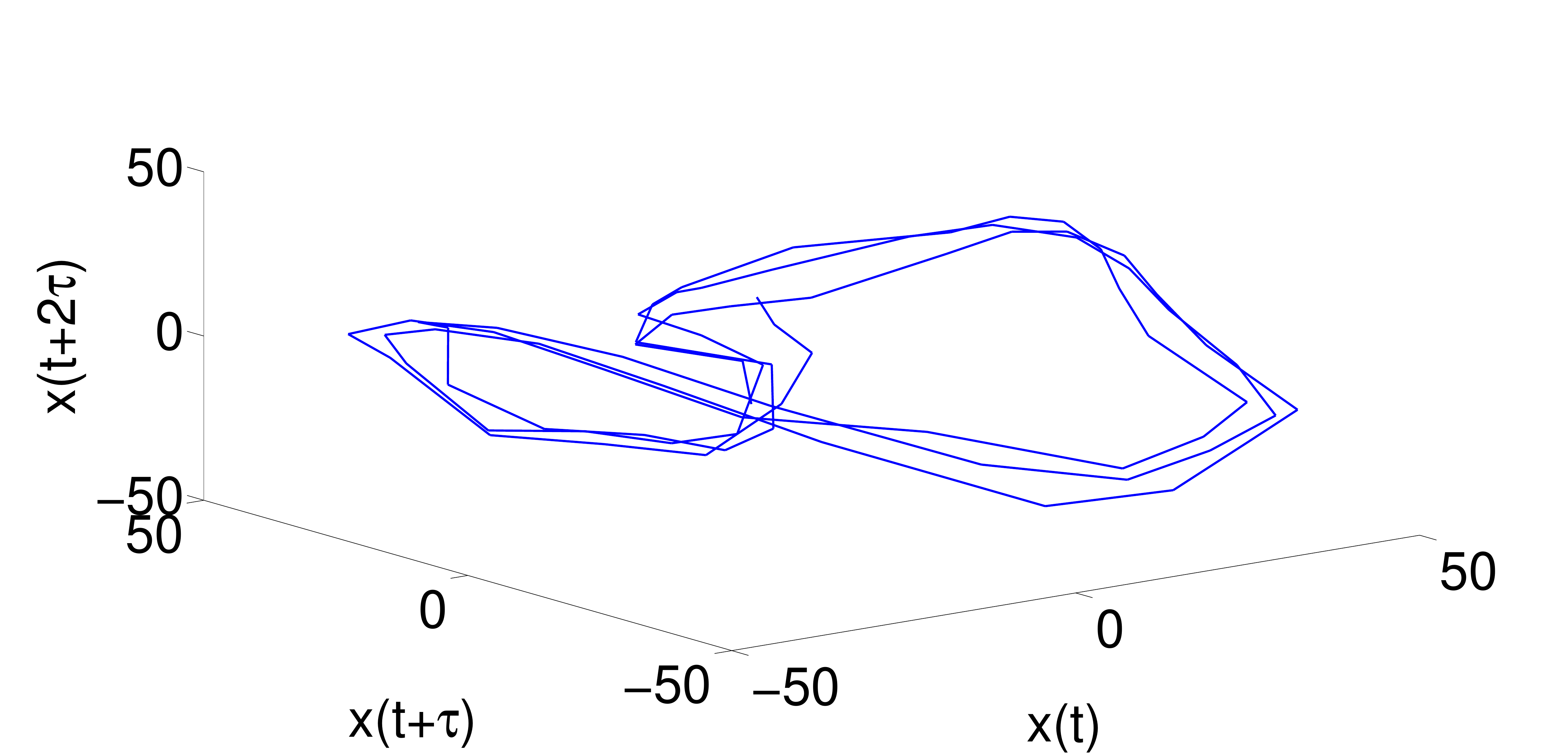}
        \end{subfigure}        
\begin{subfigure}[p]{0.19\textwidth}
                \includegraphics[width=\textwidth]{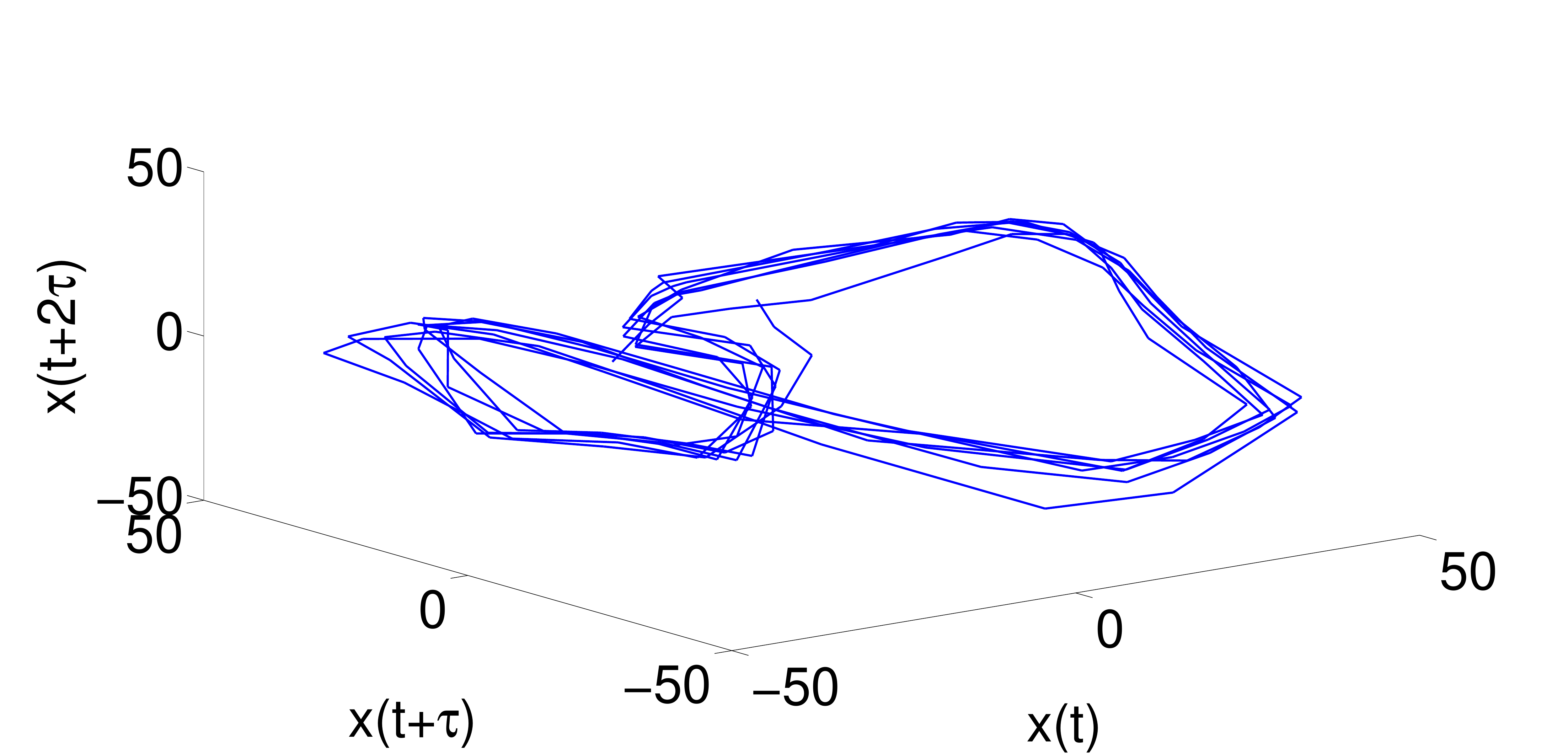}
        \end{subfigure}        
\begin{subfigure}[p]{0.19\textwidth}
                \includegraphics[width=\textwidth]{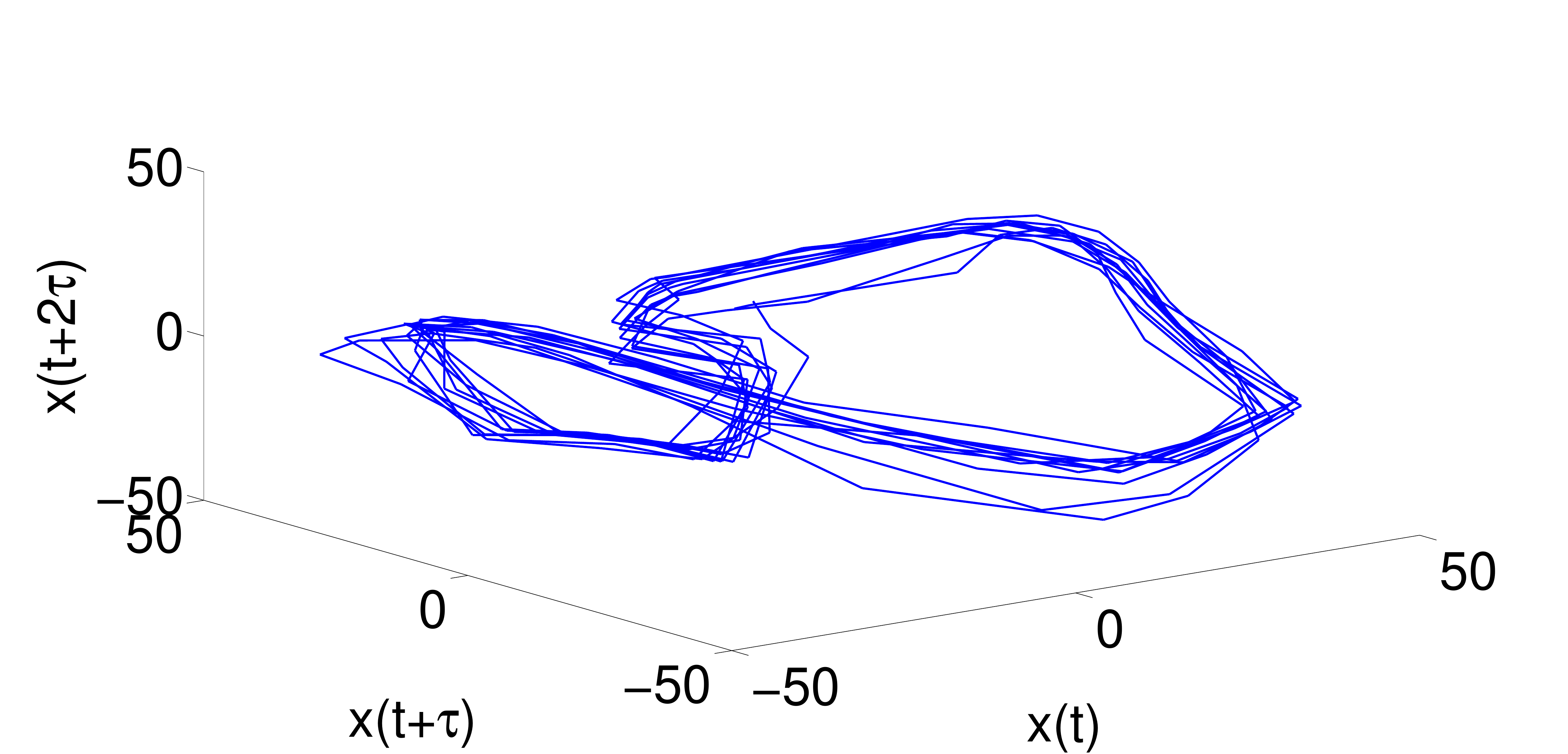}
        \end{subfigure}        
\begin{subfigure}[p]{0.19\textwidth}
                \includegraphics[width=\textwidth]{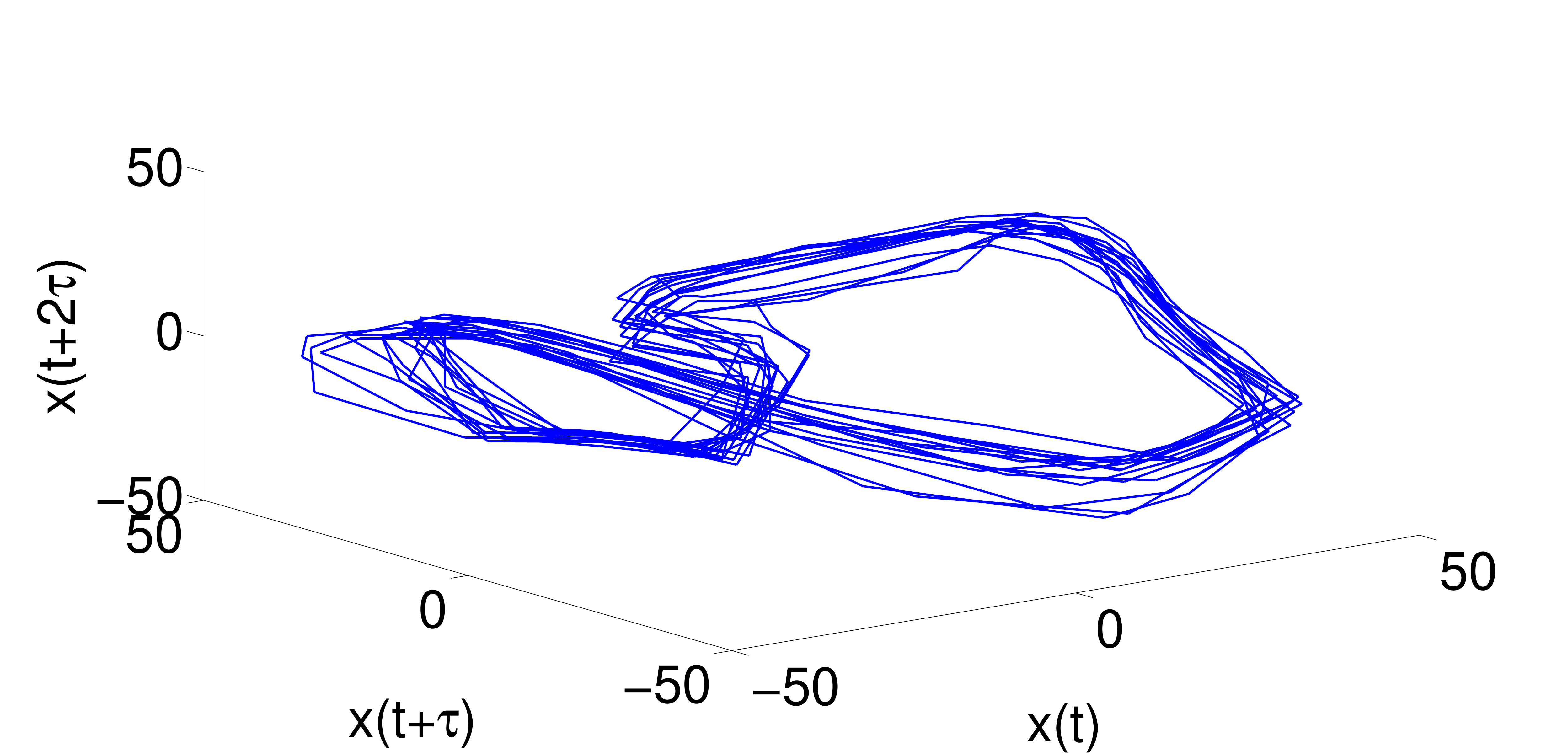}
        \end{subfigure}        
\begin{subfigure}[p]{0.19\textwidth}
                \includegraphics[width=\textwidth]{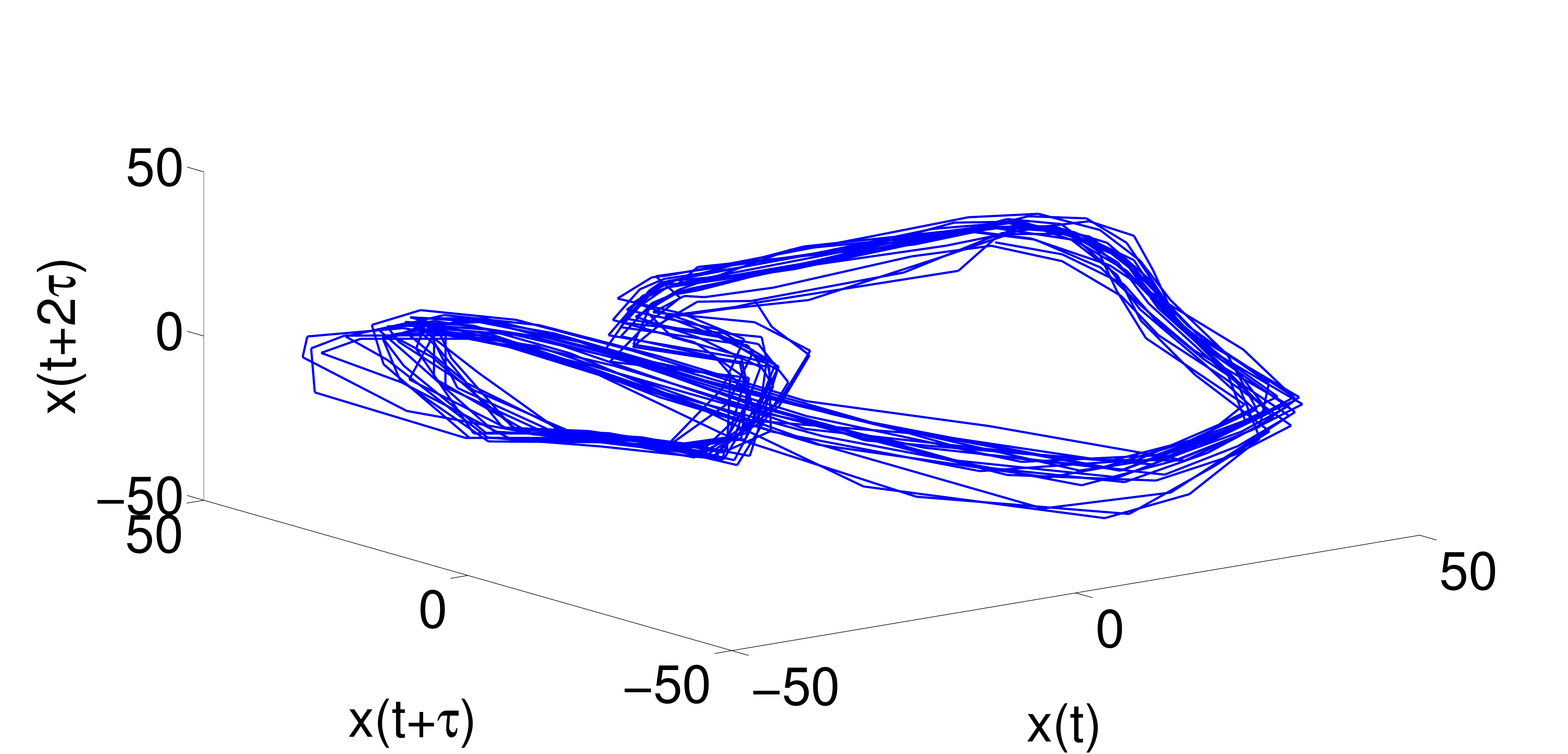}
        \end{subfigure}        
                
                \vspace{13pt}        
        		\begin{picture}(0,0)
				\put(-140,0){{\mbox{\small Reconstructed phase space of $Run$ action for different time-series lengths}}}
				\end{picture}								
        \vspace{5pt}
        \caption{Illustration of the effect of time-series lengths on reconstructed phase space for nonlinear dynamical models like Lorenz and Rossler systems, and right-foot trajectory of a subject performing $Run$ action. These examples clearly indicate that the $shape$ of the reconstructed phase space does not change with time-series length, motivating feature extraction representative of the $shape$ of the reconstructed phase space (as reported in Fig. \ref{D2stability}).}
        \label{PhaseSpace}
\end{figure*}
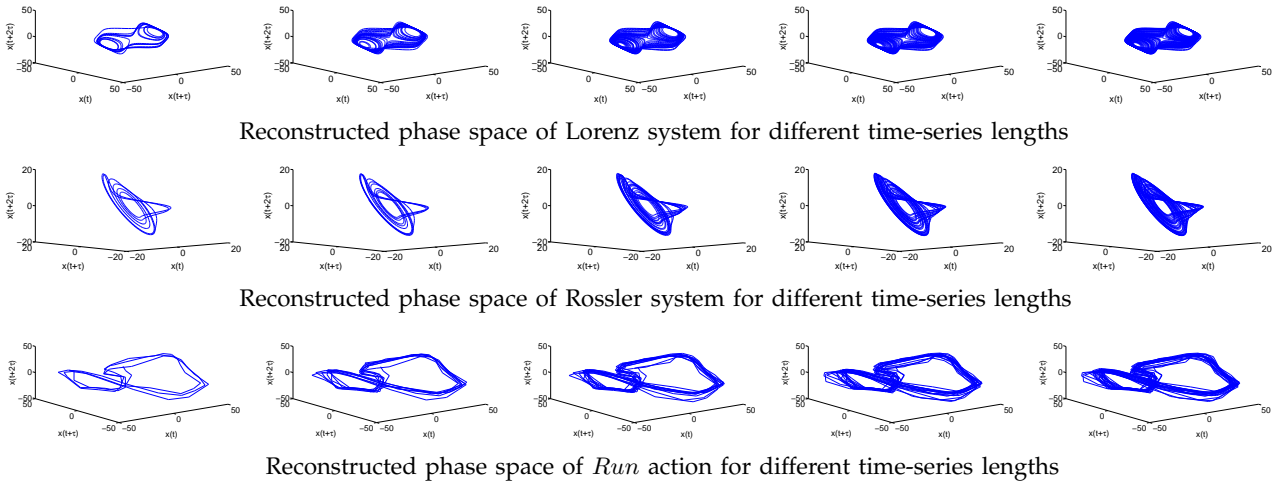
        
        \begin{table}
\centering
\caption{Experimental results on Lorenz and Rossler models for given embedding parameters ($m_L = 3$, $\tau_L = 11$, $m_R = 3$, $\tau_R = 8$) and different time-series lengths. The true value of $\lambda_1$ for Lorenz and Rossler models are $1.50$ and $0.09$ respectively \cite{wolf1985determining}.}
\begin{tabular}{| c | c | c | c |}
\hline
\textbf{System} & \textbf{N} & \textbf{Calculated $\lambda_1$} & \textbf{\% error} \\ \hline \hline
\multirow{5}{*}{\textbf{Lorenz}} & 1000 & 1.751 & 16.7 \\
 & 2000 & 1.345 & -10.3 \\ 
 & 3000 & 1.372 & -8.5 \\
 & 4000 & 1.392 & -7.2 \\
 & 5000 & 1.523 & 1.5  \\ \hline 
\multirow{5}{*}{\textbf{Rossler}} & 400 & 0.0351 & -61.0 \\
 & 800 & 0.0655 & -27.2 \\
 & 1200 & 0.0918 & 2.0 \\
 & 1600 & 0.0984 & 9.3 \\
 & 2000 & 0.0879 & -2.3 \\ \hline
\end{tabular}
\label{tab:LLEonModels}
\end{table}

\begin{figure*}     
\centering   
\begin{subfigure}[p]{0.4\textwidth}
		\includegraphics[width=\textwidth]{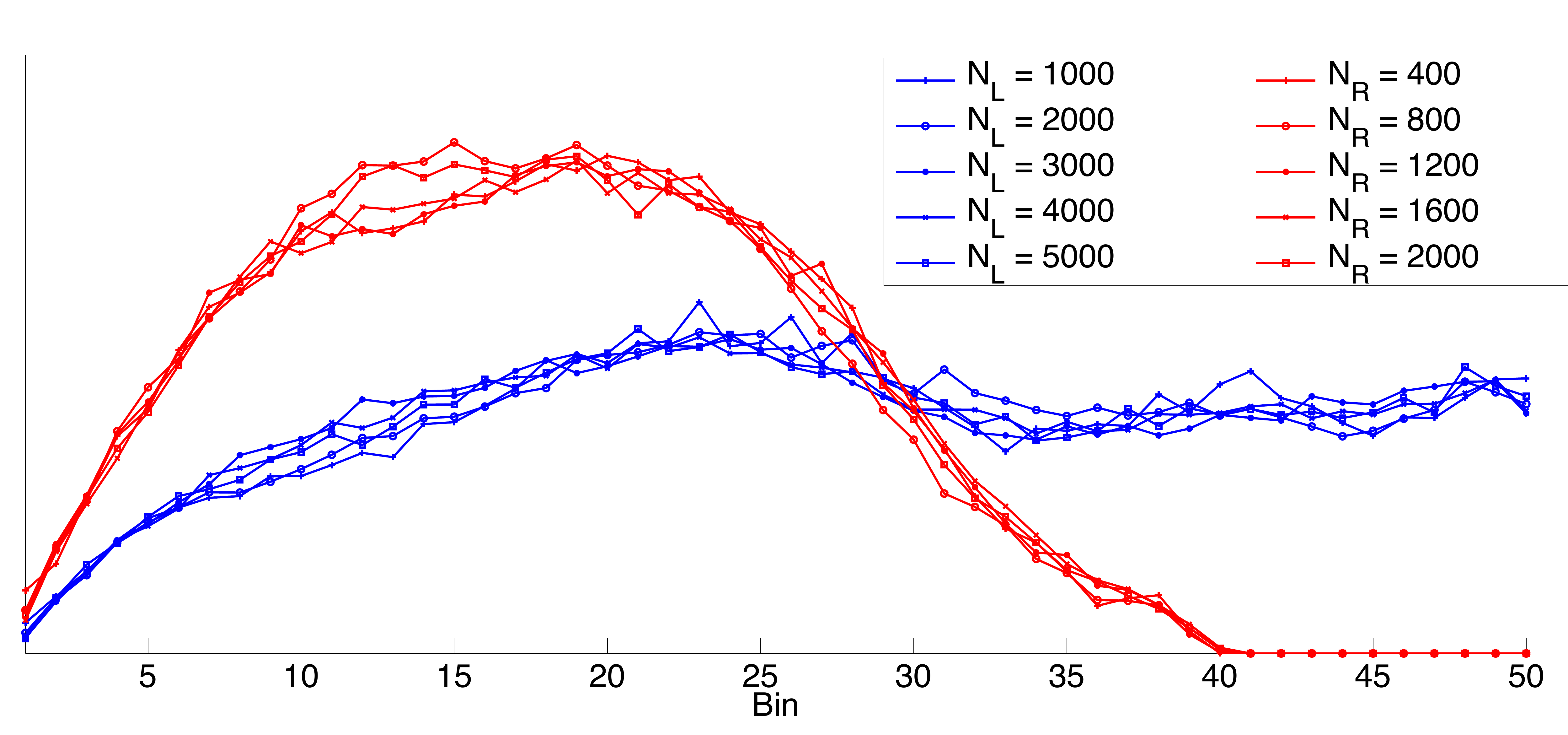}      
		\caption{Shape distribution (\textbf{D2}) of reconstructed phase space for Lorenz (blue) and Rossler (red) models for different time-series length \textit{N} ($N_{L}$ and $N_{R}$ represent time-series lengths of Lorenz and Rossler systems respectively).}         
        \end{subfigure}        
        \quad \quad
\begin{subfigure}[p]{0.4\textwidth}
 		\includegraphics[width=\textwidth]{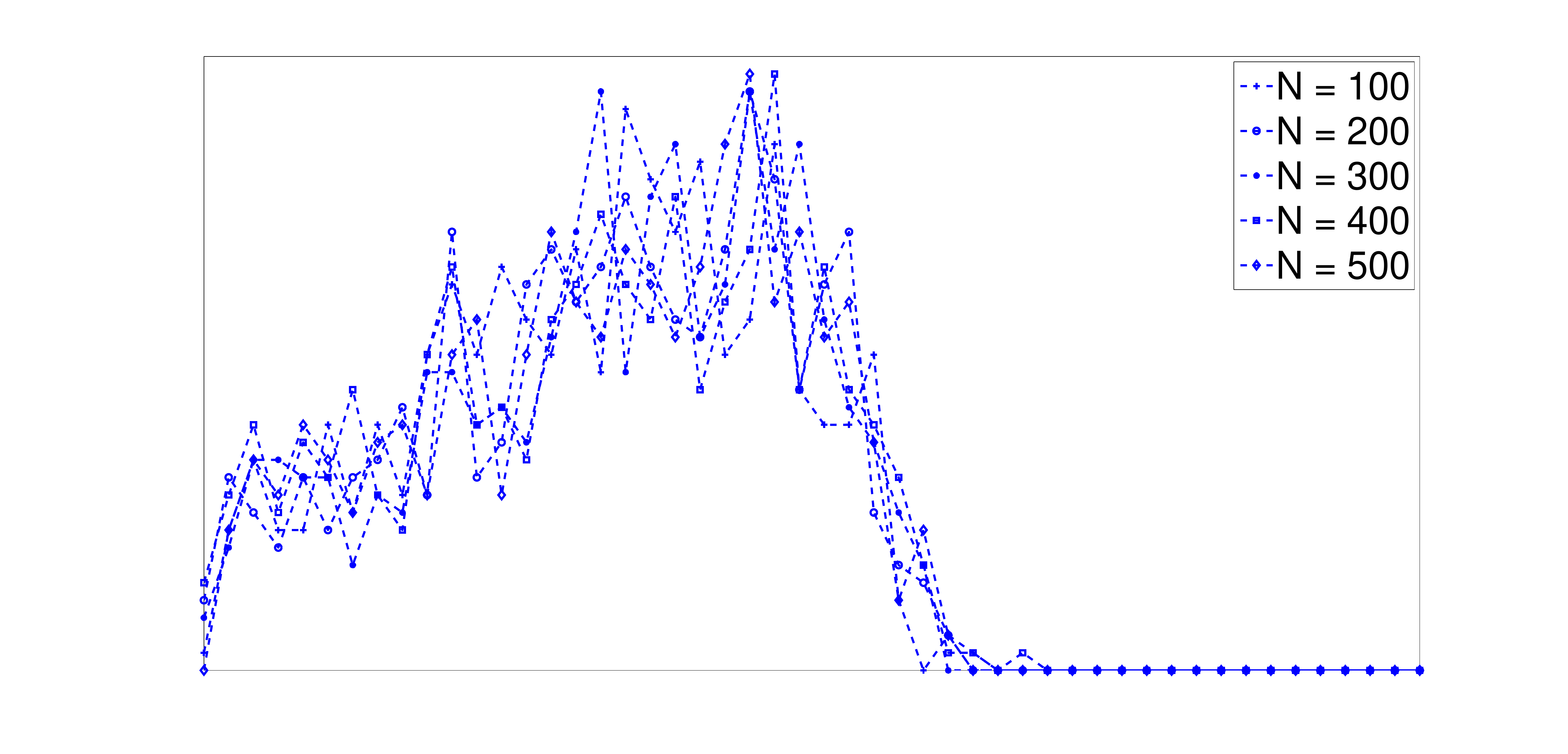}
 		\caption{Shape distribution (\textbf{D2}) of reconstructed phase space from right-foot trajectory of a subject performing $Run$ action for different time-series length.\\*}
        \end{subfigure}   
        \caption{Illustration of stability of the dynamical shape distribution (\textbf{D2}) extracted from reconstructed phase space for different time-series length. (a) shows the stability of \textbf{D2} distribution on Lorenz and Rossler systems while studies have reported significant error in estimation of largest Lyapunov exponent on these models (refer TABLE \ref{tab:LLEonModels}). (b) depicts the stability of \textbf{D2} distribution for trajectory data collected from right-foot of a subject performing $Run$ action.}    
        \label{D2stability}          
\end{figure*}

\subsection{Test on Models}
The framework was tested on the Lorenz and Rossler models to determine whether the shape feature can be effectively used to classify differences in shape of reconstructed phase space of nonlinear dynamical systems. We compare the performance of the proposed framework with that of largest Lyapunov exponent. The effect of time-series length on estimation of largest Lyapunov exponent was revealed by Rosenstein \textit{et al}. \cite{lyaprosen}, by evaluating the performance of the algorithm they proposed for estimation of $\lambda_1$ for various time-series lengths. The simulation results on Lorenz and Rossler models are shown in TABLE \ref{tab:LLEonModels}. Their findings indicate that the estimation error increases with reduction in time-series length ($N$). Fig. \ref{PhaseSpace} depicts the variations in reconstructed phase space for different time-series length with defined embedding parameters. It is evident from these plots that the $shape$ of the reconstructed phase space remain sufficiently similar and can be used as a discriminative feature for classification purposes. Also, from Fig. \ref{D2stability}, the shape distribution (using \textbf{D2} shape function) was found to be stable for different time-series lengths. This striking ability of our feature representations to be robust to changes in data length will be useful in applications related to human activity analysis, where the signal observation time is small/variable. 

\section{Experiments and Results}
\label{sec:exp}
The proposed framework for representation of dynamics was evaluated on the following video-based inference tasks: \\
(1) Action recognition on a motion capture dataset \cite{ali2007chaotic}. \\
(2) Action recognition on the MSR Action3D dataset released by Microsoft Research \cite{li2010action}. \\
(3) Action quality estimation on stroke rehabilitation datasets collected in hospital and home environments \cite{baran2011design,chen2011computational}. \\
(4) Dynamic scene classification on the Maryland ``in-the-wild'' natural scene dataset \cite{movingvistas} and the Yupenn ``stabilized'' scene dataset \cite{derpanis2012dynamic}.

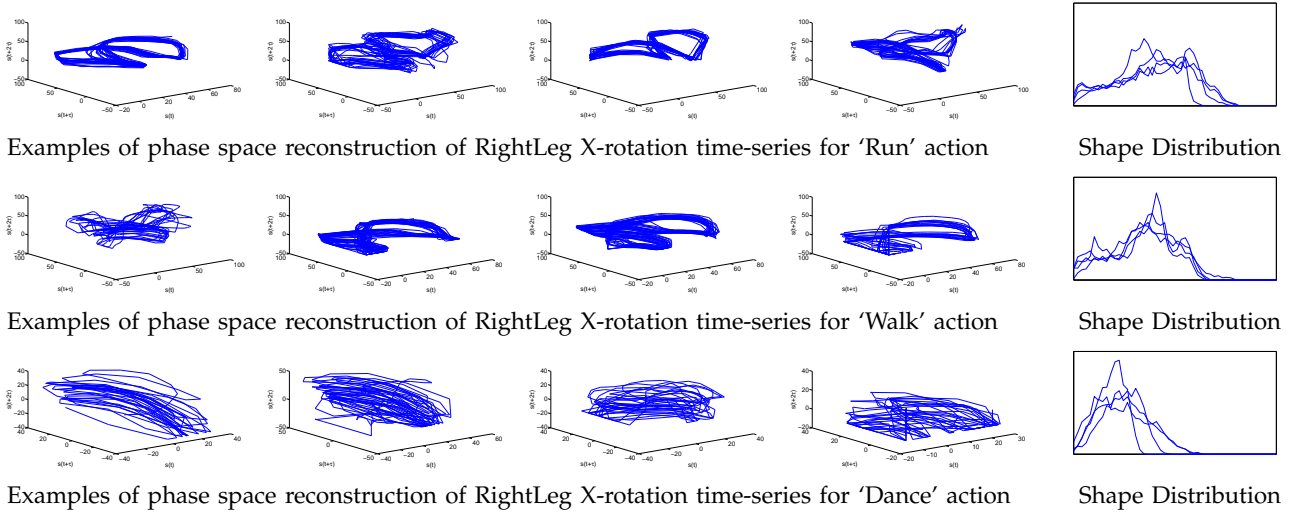
\begin{figure*}
		\begin{subfigure}[p]{0.2\textwidth}
                \includegraphics[width=\textwidth]{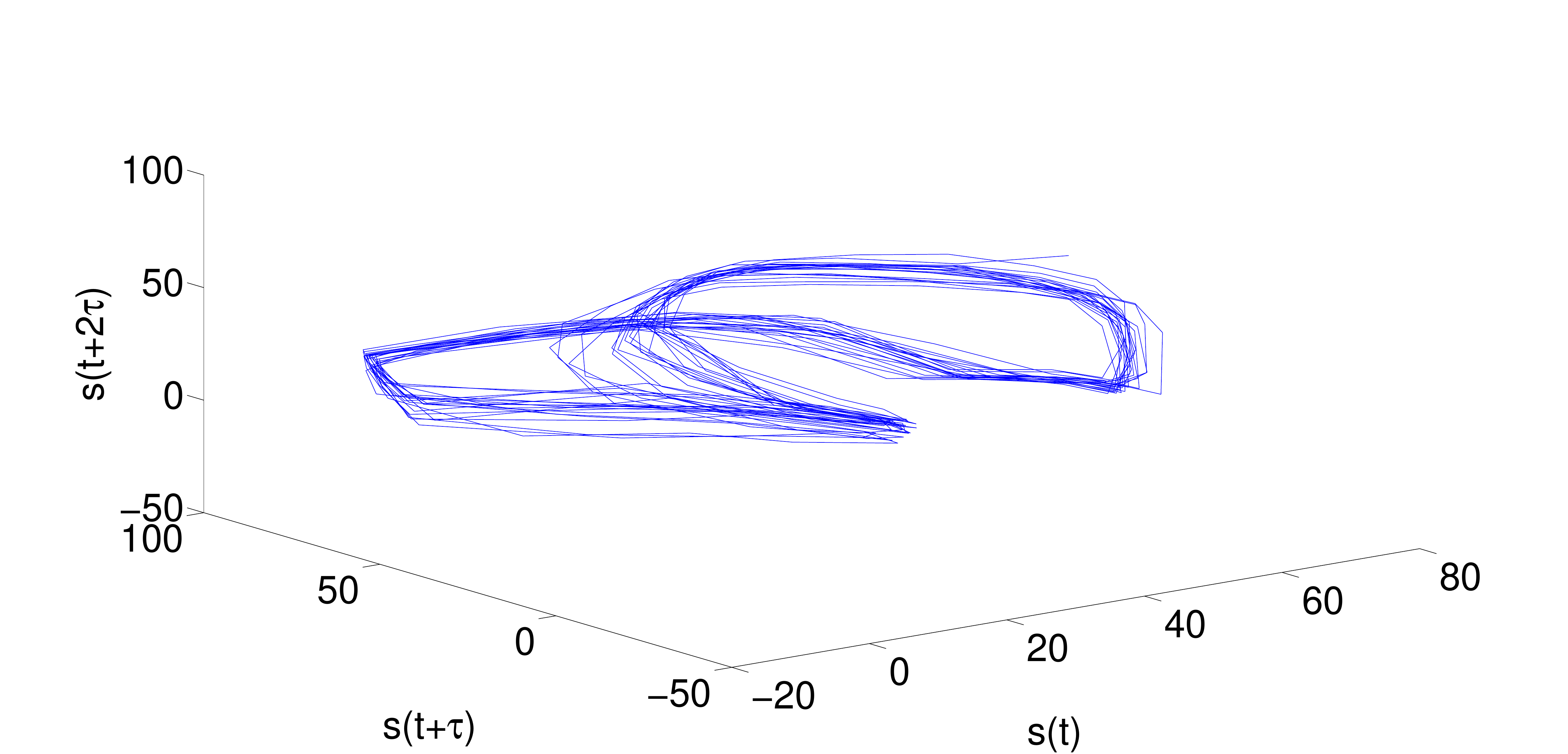}
        \end{subfigure}%
        		\begin{subfigure}[p]{0.2\textwidth}
                \includegraphics[width=\textwidth]{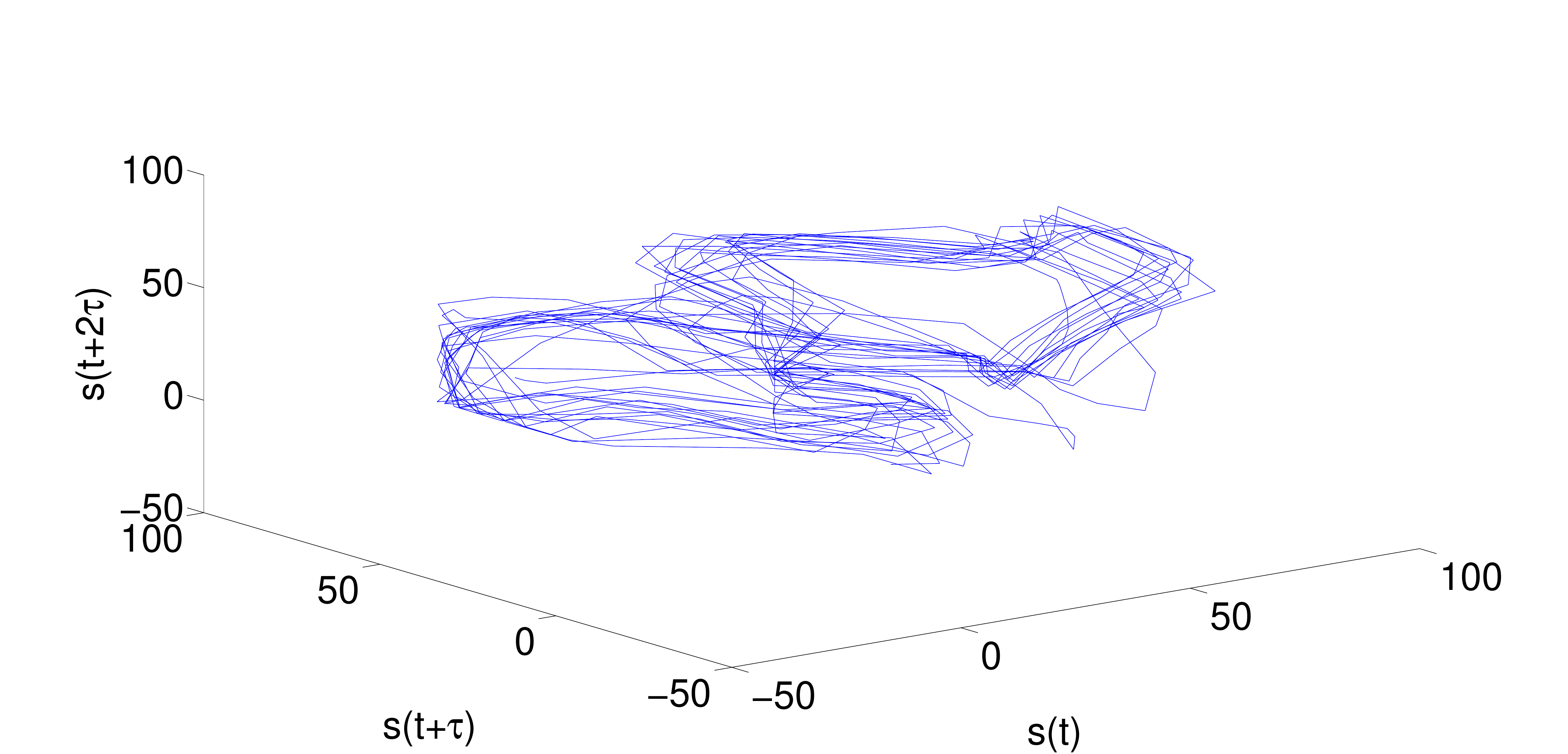}
        \end{subfigure}%
        		\begin{subfigure}[p]{0.2\textwidth}
                \includegraphics[width=\textwidth]{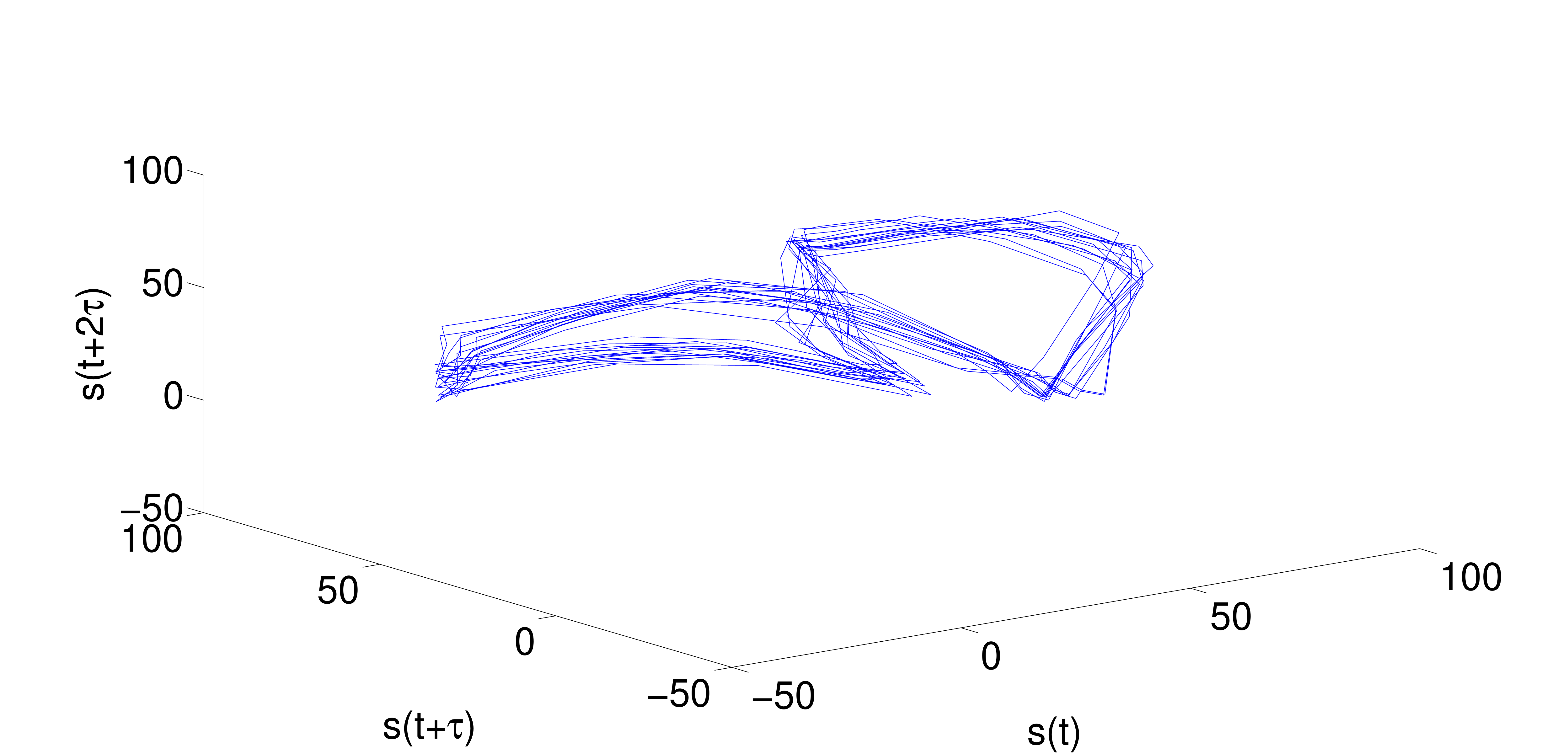}
        \end{subfigure}%
        		\begin{subfigure}[p]{0.2\textwidth}
                \includegraphics[width=\textwidth]{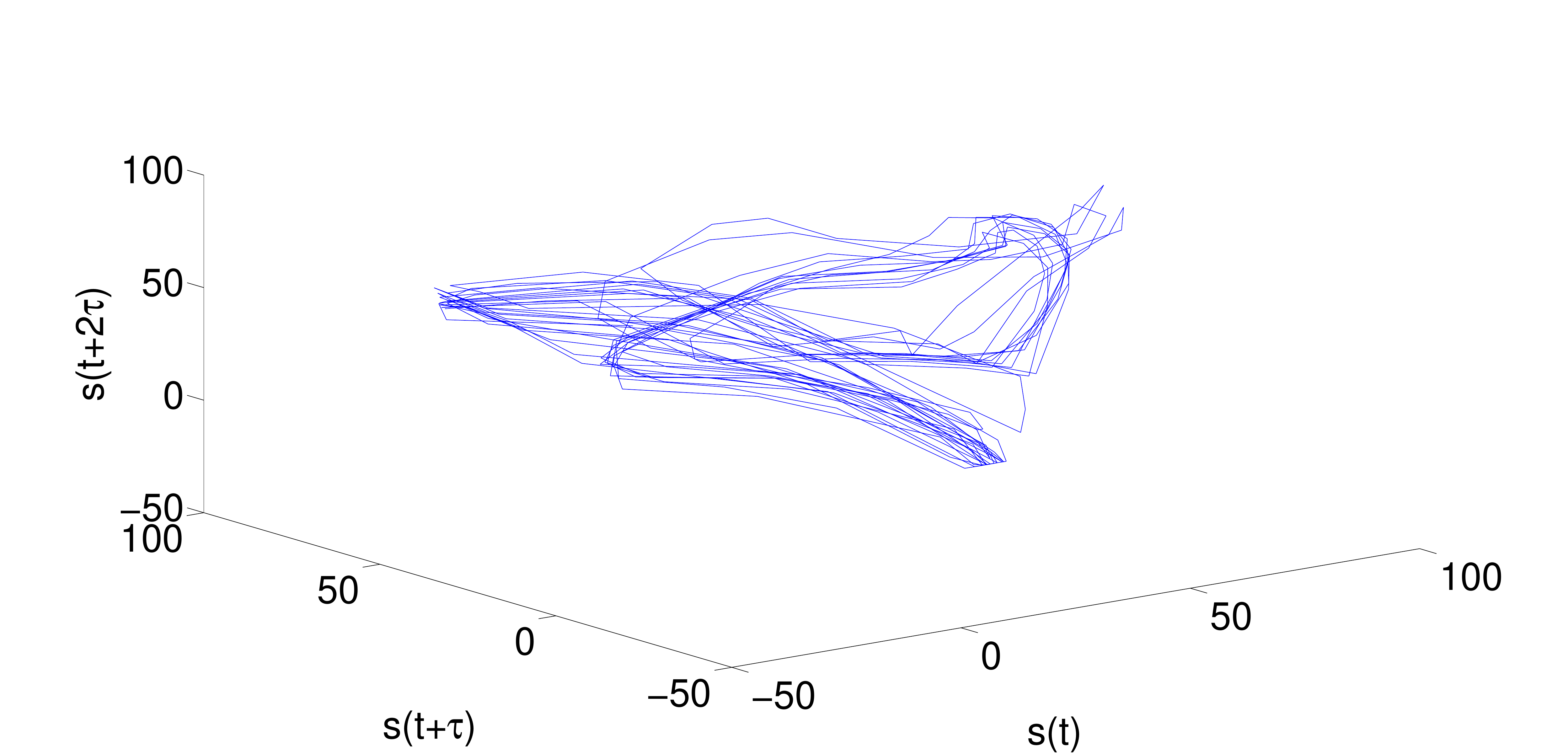}
        \end{subfigure}%
                		\begin{subfigure}[p]{0.2\textwidth}
                \includegraphics[width=\textwidth]{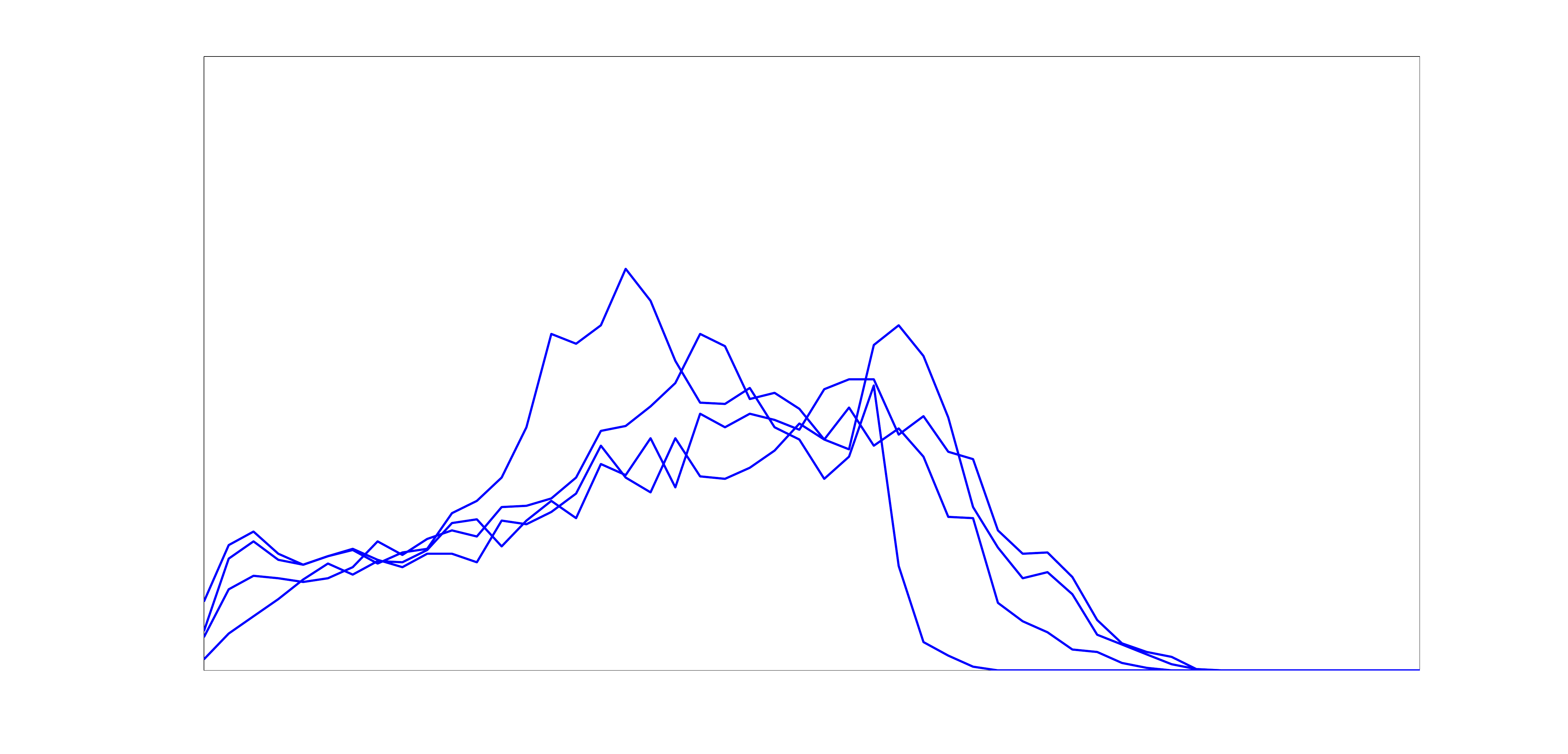}
        \end{subfigure}%
        
        \vspace{13pt}        
        		\begin{picture}(0,0)
				\put(5,0){{\mbox{\small Examples of phase space reconstruction of RightLeg X-rotation time-series for `Run' action}}}
				\end{picture}				
				\begin{picture}(0,0)
				\put(405,0){{\mbox{\small Shape Distribution}}}
				\end{picture}
        \vspace{5pt}
        
        		\begin{subfigure}[p]{0.2\textwidth}
                \includegraphics[width=\textwidth]{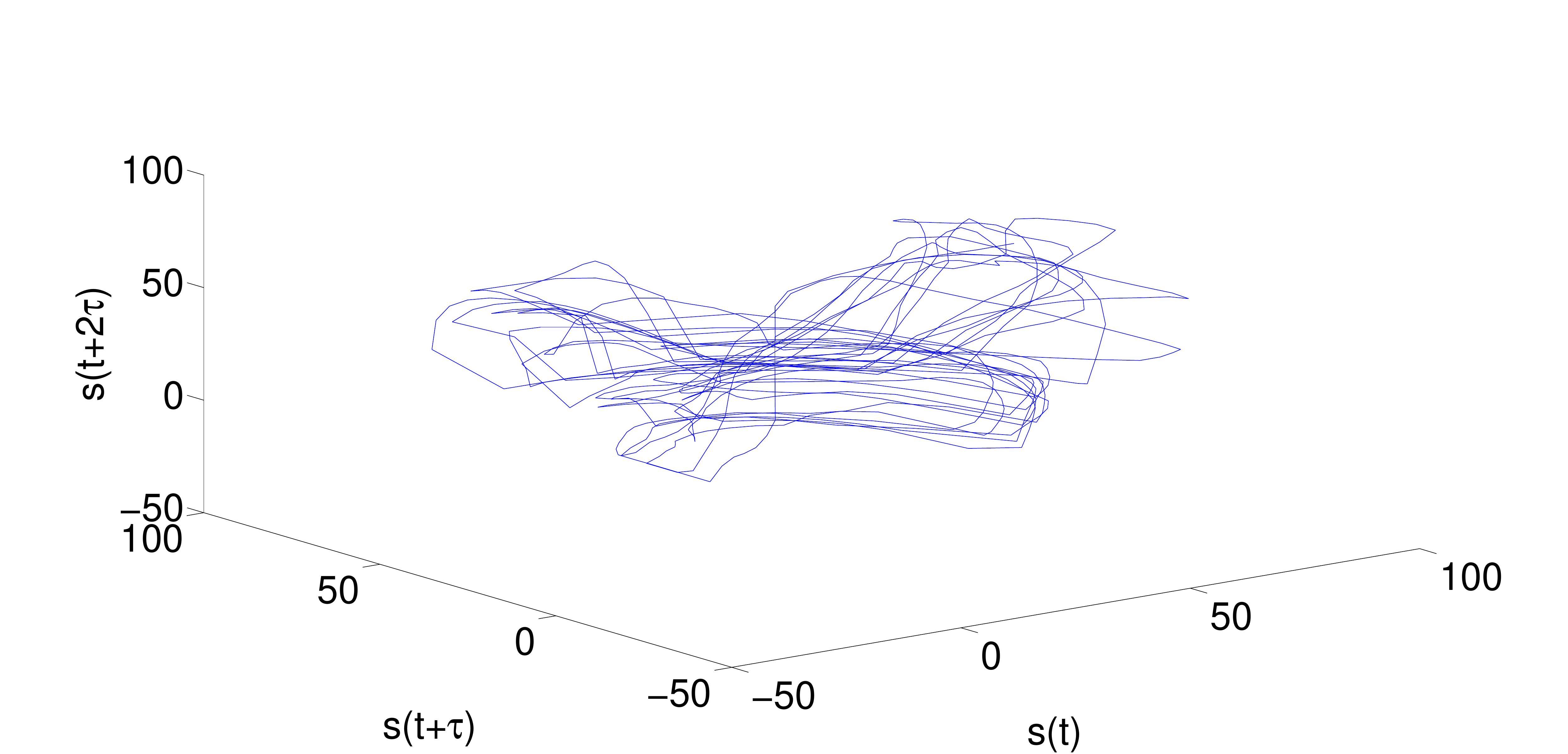}
        \end{subfigure}%
        		\begin{subfigure}[p]{0.2\textwidth}
                \includegraphics[width=\textwidth]{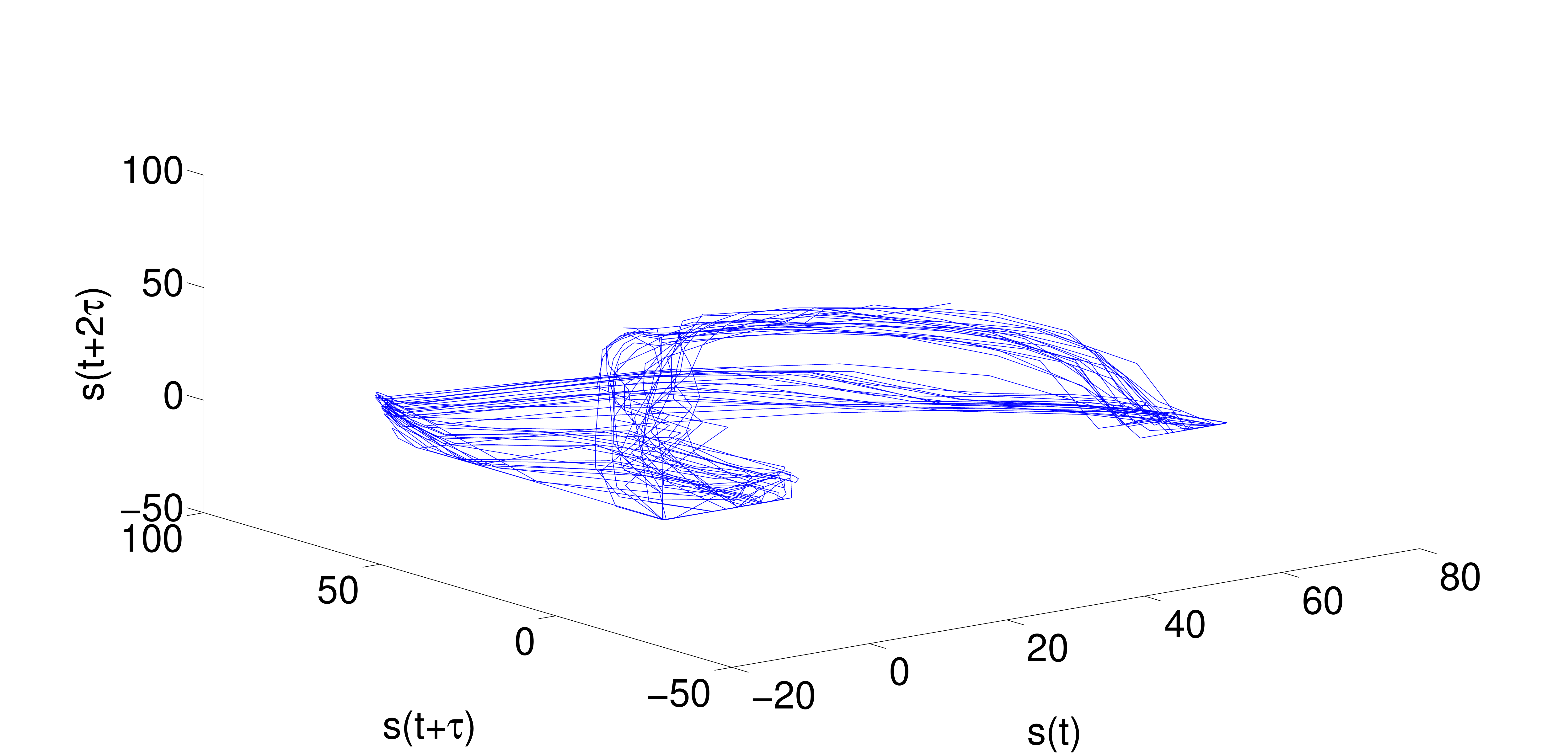}
        \end{subfigure}%
        		\begin{subfigure}[p]{0.2\textwidth}
                \includegraphics[width=\textwidth]{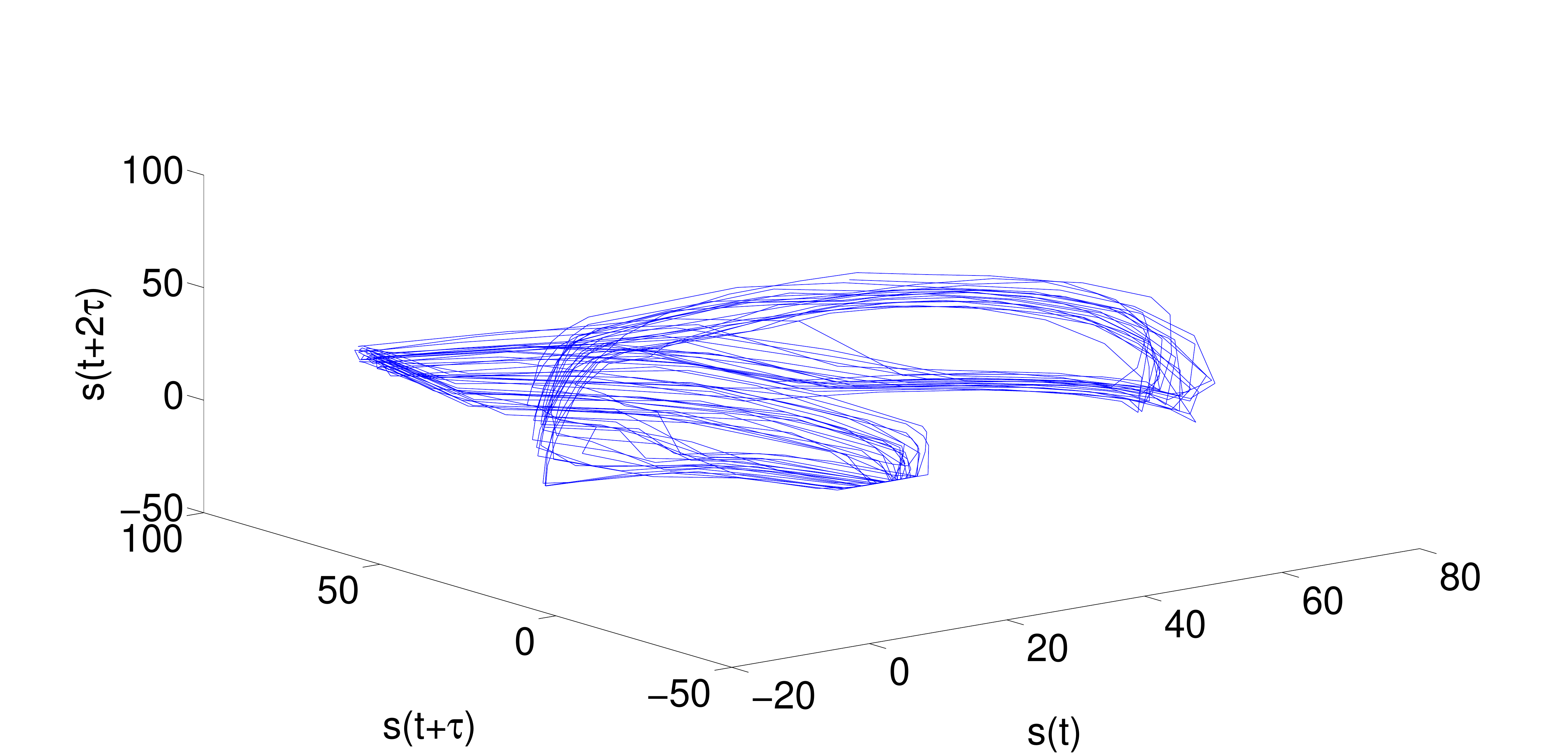}
        \end{subfigure}%
        		\begin{subfigure}[p]{0.2\textwidth}
                \includegraphics[width=\textwidth]{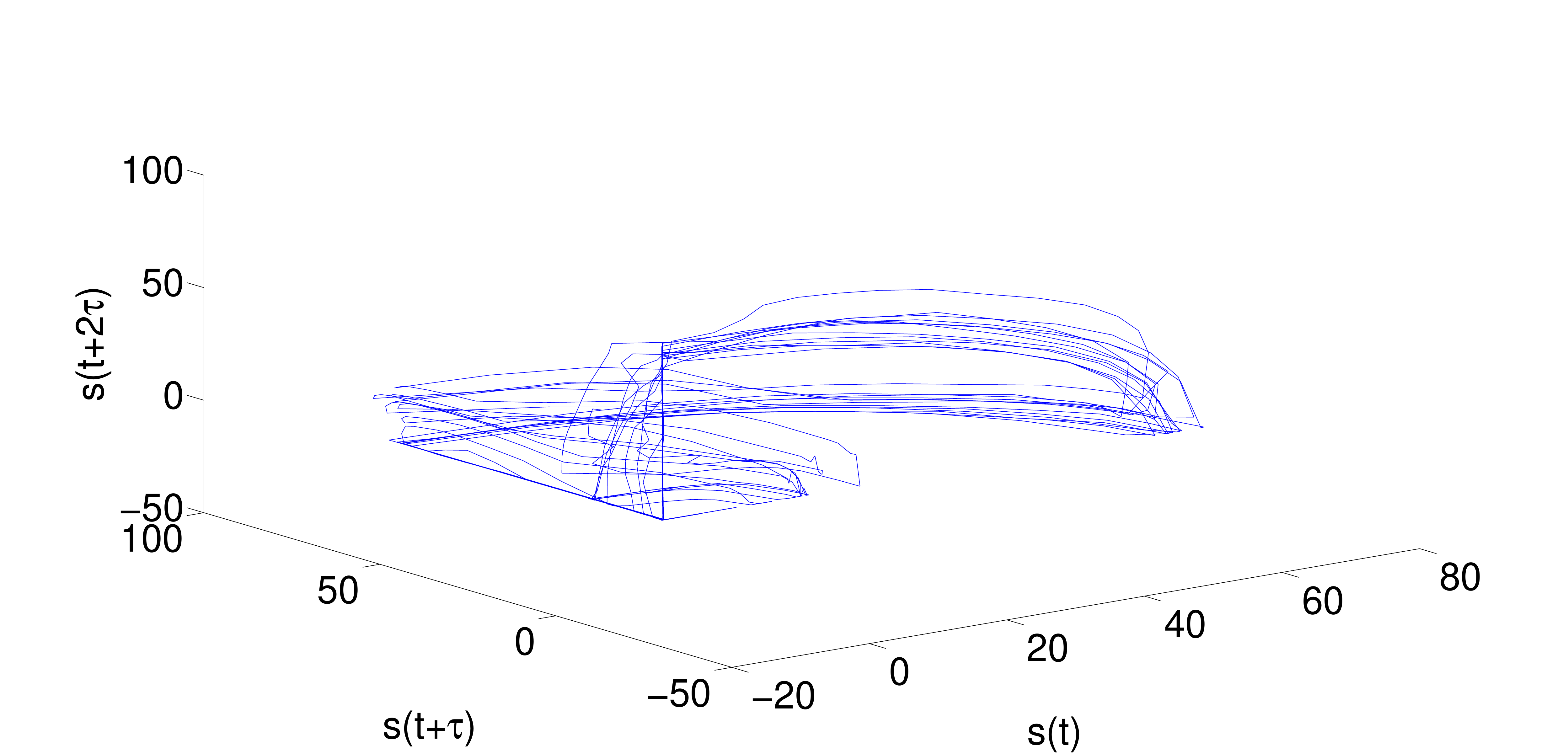}
        \end{subfigure}%
                		\begin{subfigure}[p]{0.2\textwidth}
                \includegraphics[width=\textwidth]{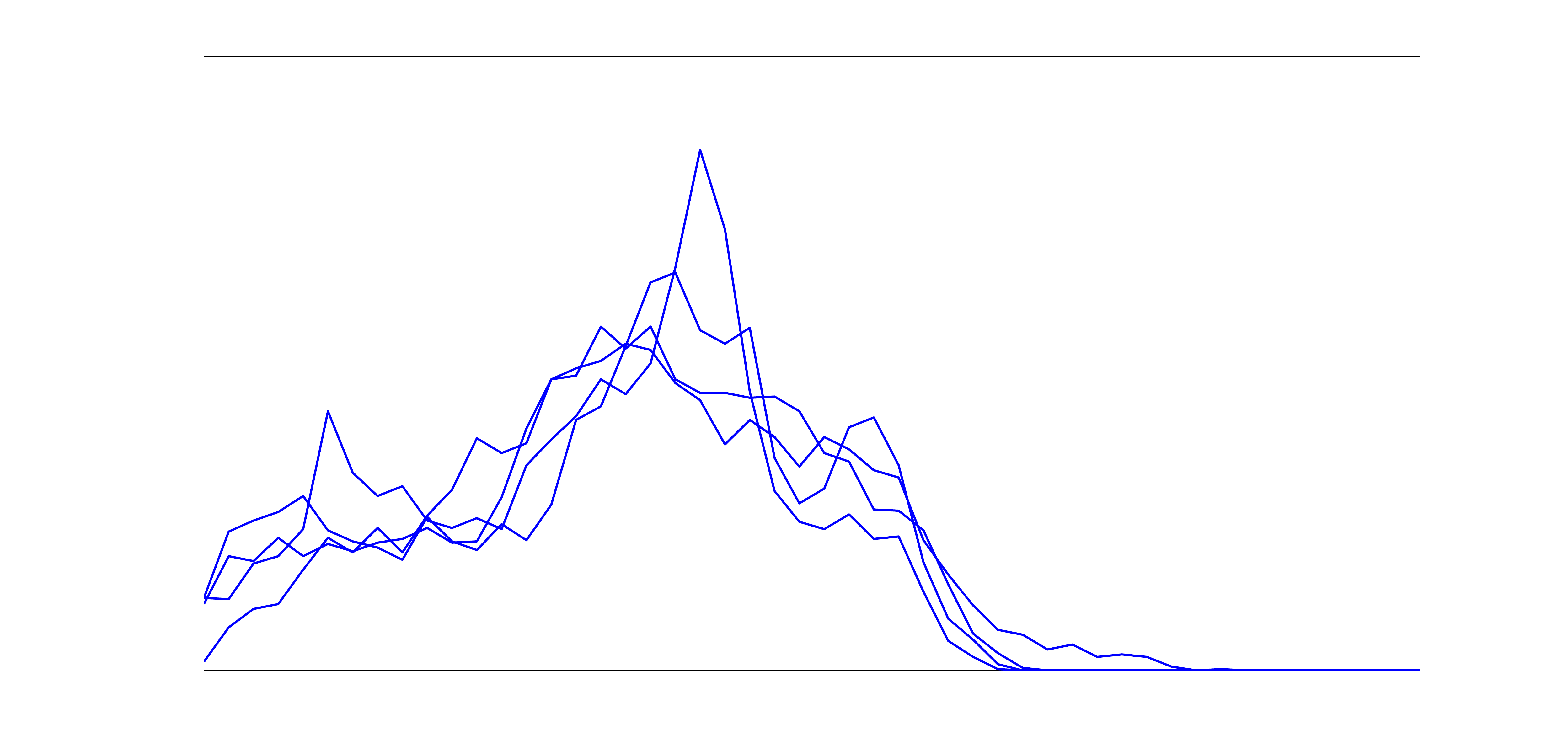}
        \end{subfigure}%
        
                \vspace{13pt}        
        		\begin{picture}(0,0)
				\put(5,0){{\mbox{\small Examples of phase space reconstruction of RightLeg X-rotation time-series for `Walk' action}}}
				\end{picture}				
				\begin{picture}(0,0)
				\put(405,0){{\mbox{\small Shape Distribution}}}
				\end{picture}
        \vspace{5pt}

                		\begin{subfigure}[p]{0.2\textwidth}
                \includegraphics[width=\textwidth]{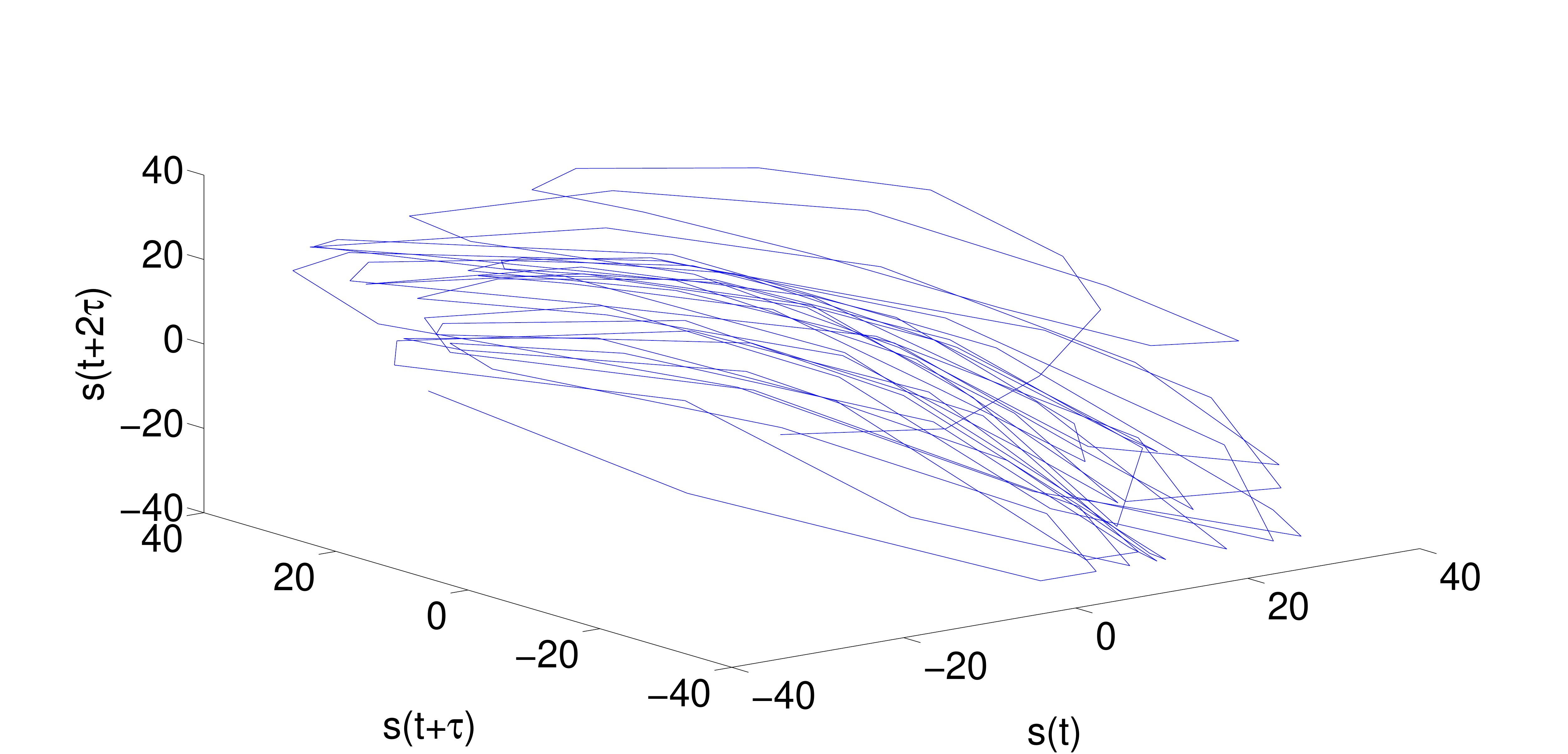}
        \end{subfigure}%
        		\begin{subfigure}[p]{0.2\textwidth}
                \includegraphics[width=\textwidth]{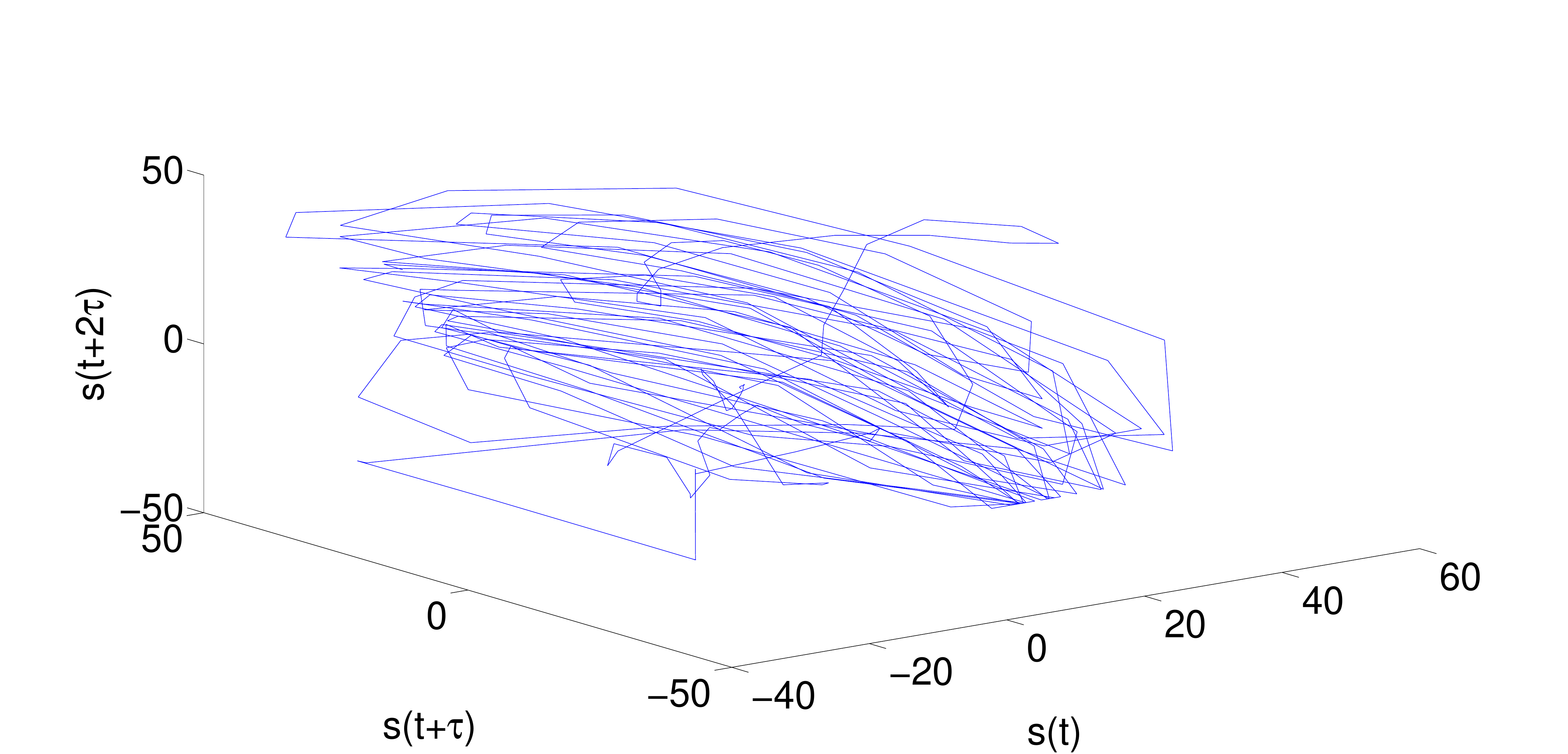}
        \end{subfigure}%
        		\begin{subfigure}[p]{0.2\textwidth}
                \includegraphics[width=\textwidth]{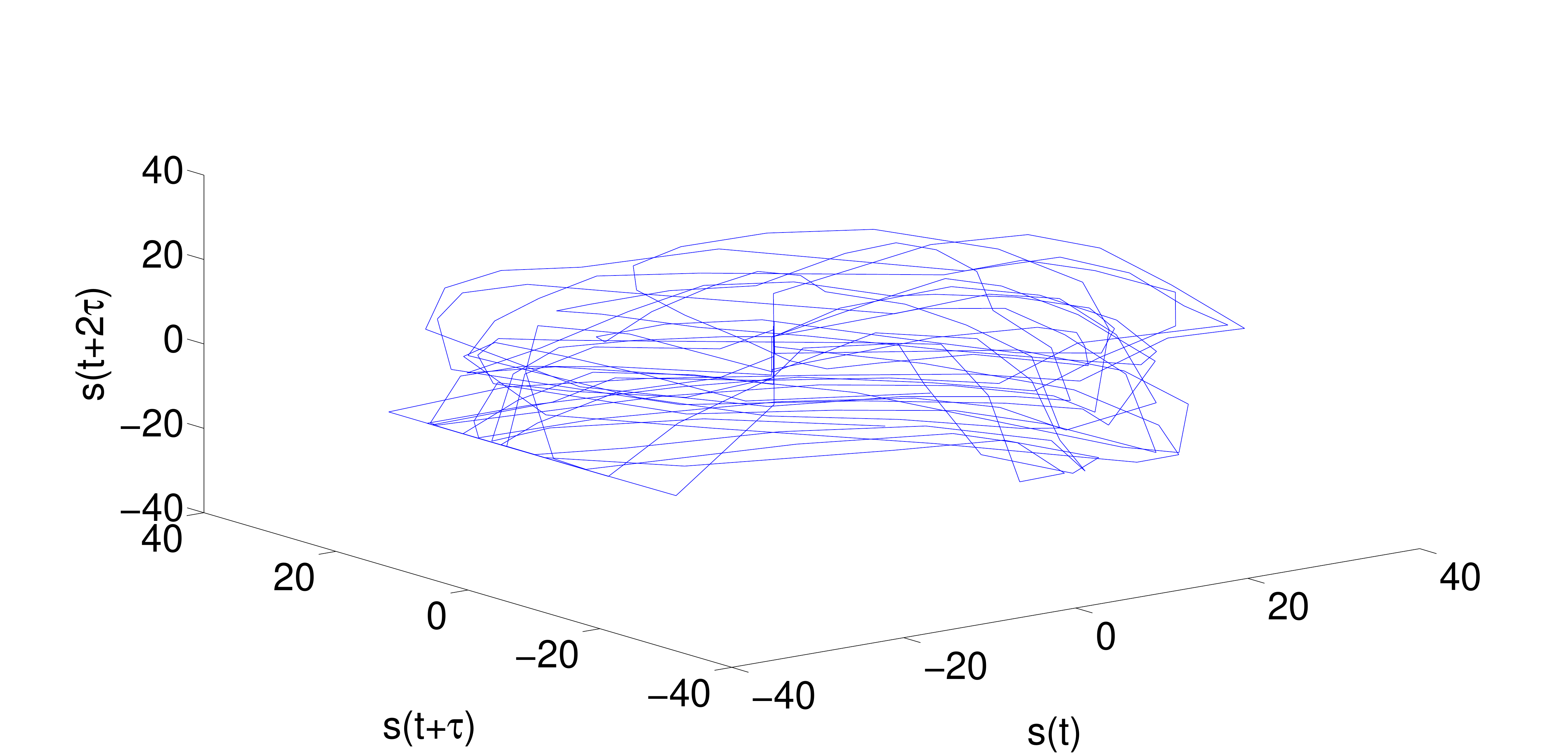}
        \end{subfigure}%
        		\begin{subfigure}[p]{0.2\textwidth}
                \includegraphics[width=\textwidth]{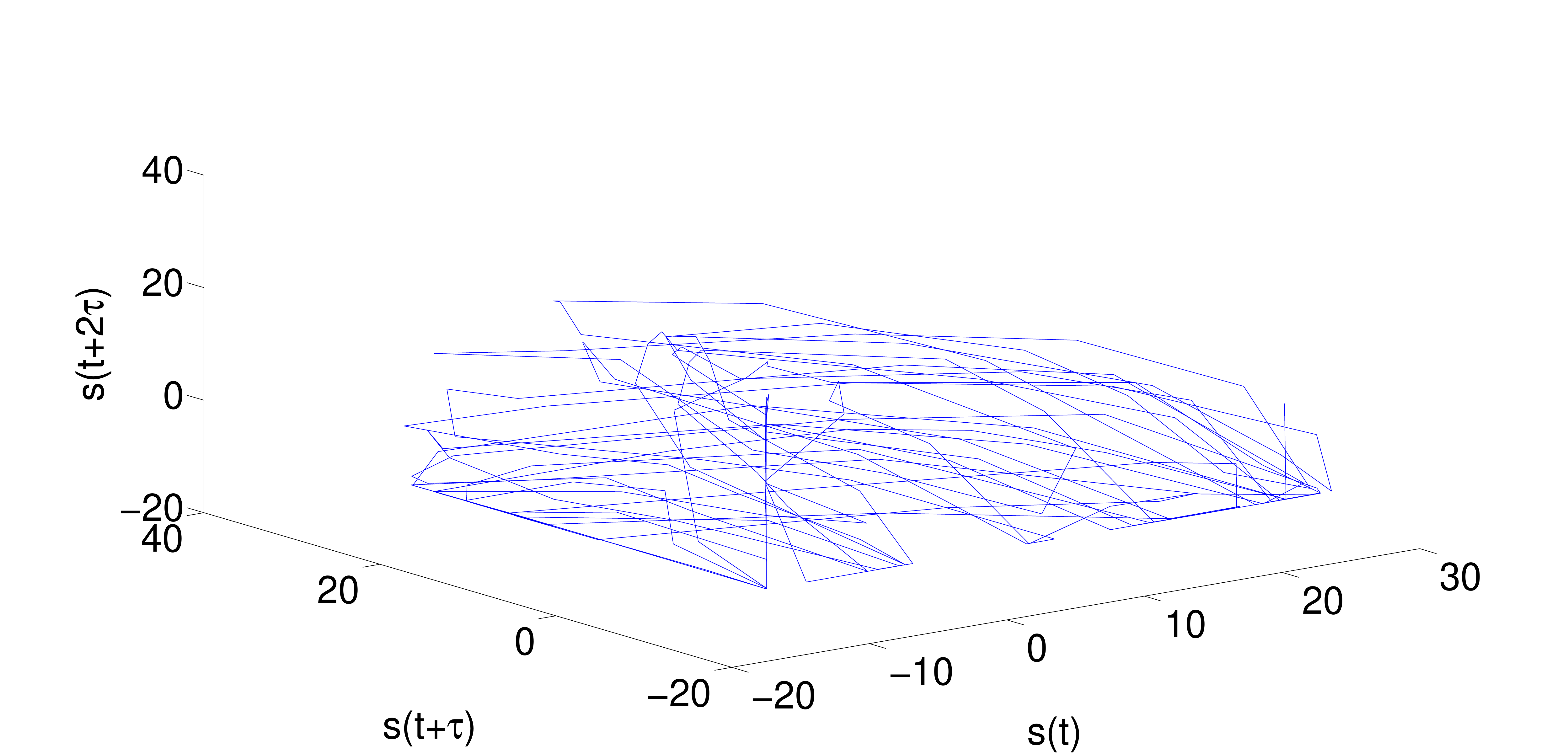}
        \end{subfigure}%
                		\begin{subfigure}[p]{0.2\textwidth}
                \includegraphics[width=\textwidth]{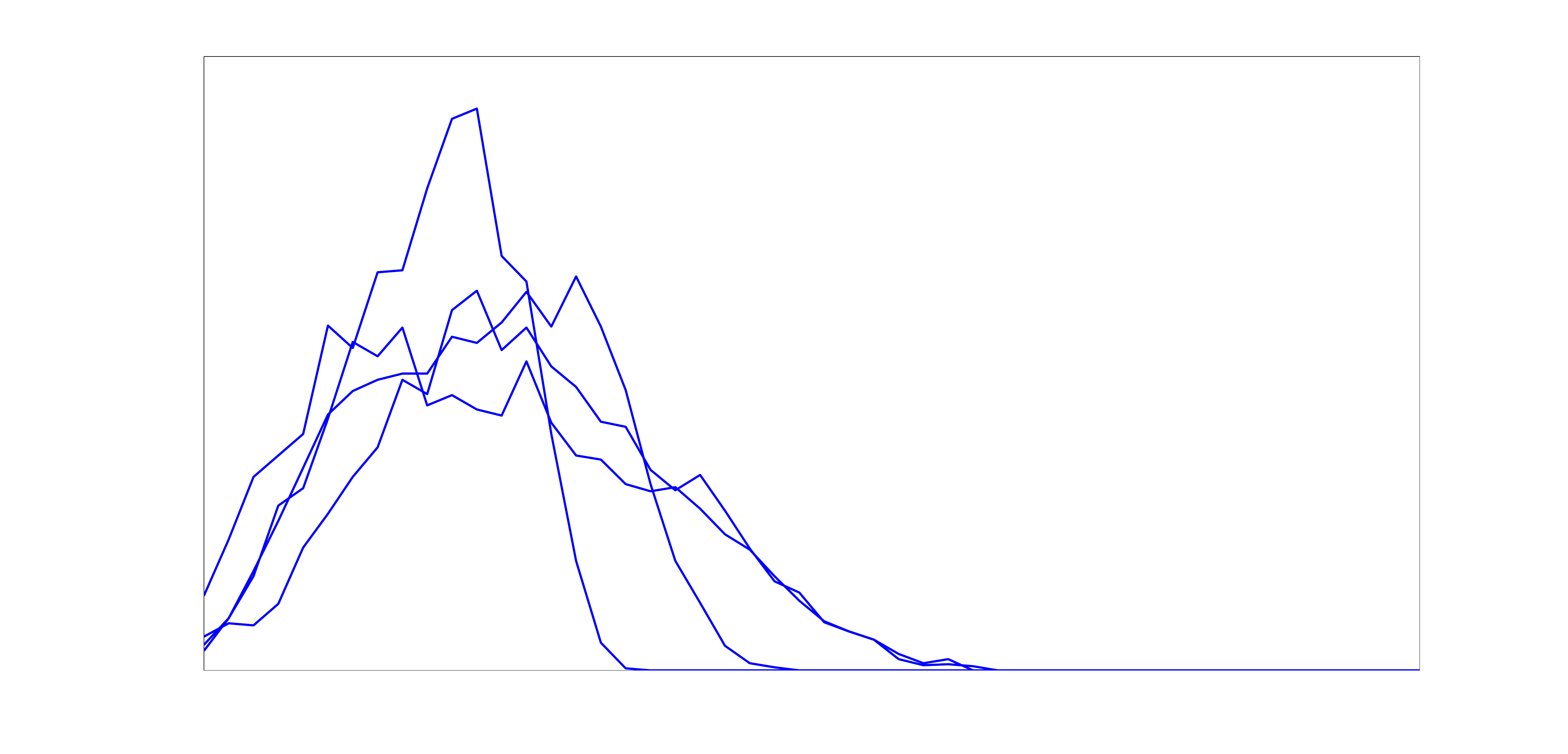}
        \end{subfigure}%
        \vspace{13pt}        
        		\begin{picture}(0,0)
				\put(5,0){{\mbox{\small Examples of phase space reconstruction of RightLeg X-rotation time-series for `Dance' action}}}
				\end{picture}				
				\begin{picture}(0,0)
				\put(405,0){{\mbox{\small Shape Distribution}}}
				\end{picture}
        \vspace{5pt}        
        \caption{Illustration of the phase space reconstruction and dynamical shape feature extraction (\textbf{D2} shape feature) using four examples of \textit{Run}, \textit{Walk} and \textit{Dance} action classes each from the motion capture dataset \cite{ali2007chaotic}. As an example, phase space reconstruction of X-rotation time-series from right leg of subjects performing these actions is shown. Embedding parameters, $m$ was selected to be $3$ and $\tau$ was calculated by method explained in section \ref{TauCalc}. It is evident from these examples that the `shape' of phase space is a representative feature for an action class and can be captured using shape distributions.}
        \label{fig:motioncapplots}
\end{figure*}

\textbf{Baseline:} The main contribution of our work is to propose a better way to encode dynamics compared to traditional chaotic invariants. To evaluate the effectiveness of our framework, we provide comparative results in each experiment with a feature vector \footnote{Code available at \\ http://www.physik3.gwdg.de/tstool/HTML/index.html} of traditional chaotic invariants obtained by concatenating the largest Lyapunov exponent, correlation dimension and correlation integral (for $8$ values of radius) resulting in a $10$-dimensional feature vector denoted as \textbf{\textit{Chaos}}. For a fair comparison, the embedding procedure is fixed as mentioned in earlier sections. 

\subsection{Motion Capture Dataset}
\label{Motioncapdata}
In the first experiment, we evaluate the performance of the proposed framework using $3$-dimensional motion capture sequences of body joints of subjects performing actions released by FutureLight, R\&D division of Santa Monica Studios \cite{ali2007chaotic}. The dataset is a collection of five actions: \textit{dance, jump, run, sit} and \textit{walk} with $31, 14, 30, 35$ and $48$ instances respectively. The classification problem on this dataset is shown to be challenging due to the presence of significant intra-class variations \cite{ali2007chaotic}. The data is in the form of trajectories of $3$D rotation angles from 18 body joints. We use all body joints except the hip joint, to remove any effects of translational movement of the body. The $3$D time-series from these $17$ body joints were divided into scalar time-series resulting in a $51$-dimensional vector representation for each action. Phase space reconstruction and dynamical shape feature extraction was performed. The results of the leave-one-out cross-validation approach using a nearest neighbor classifier (using Euclidean and $\chi^2$ distance metrics) are tabulated in TABLE \ref{tab:mocapres100}. The best classification performance we achieved was a mean accuracy of $99.37\%$ using \textbf{DT2} dynamical shape feature, in comparison with $89.7\%$ reported by Ali \textit{et al.} in \cite{ali2007chaotic} using traditional chaotic invariants. In addition, we see that the classification performance of each dynamical shape feature is significantly better than the results achieved by using traditional chaotic invariants (\textbf{\textit{Chaos}} with $m$ = $3$ \& $m = 5$). The proposed action modeling framework achieves near-perfect classification accuracy on the motion capture dataset even in the presence of significant intra-class variations indicating its stability. This is also evident from the examples shown in Fig. \ref{fig:motioncapplots}, where minor variations in the reconstructed phase space (in the form of intra-class variations) has not produced any significant effect on the dynamical shape feature indicating the stability of the proposed framework. From these results, we see that the dynamical shape features with temporal evolution information (\textbf{DT1} and \textbf{DT2}) performs better than the shape features \textbf{D1}, \textbf{D2} and \textbf{D3}, hence substantiating our hypothesis that shape functions with dynamical evolution information should only improve the recognition performance.

\begin{table}
\begin{center}
\caption{Classification rates for the various proposed dynamical shape features of phase space on the motion capture dataset with $m = 3$ (and $m = 5$ in parentheses). For comparison, we use Euclidean distance and chi-squared distance metrics as a measure of distance between probability distributions. We see that \textbf{DT2} achieves highest classification rate of $99.37\%$. The confusion table of the same is reported in TABLE \ref{tab:mocap confTable}.}
\begin{tabular}{| c | c | c |}
\hline
\multirow{2}{*}{\textbf{Dynamical Shape Feature}} & \multicolumn{2}{c|}{\textbf{Distance Measure}}  \\ 
\cline{2-3} & \textbf{$L_2$} & \textbf{$\chi^2$} \\ \hline \hline
\textbf{Chaos} & 80.38 (82.28) & 83.54 (85.54)	\\ \hline
\textbf{Ali} \textit{et al}. & 89.70 & - \\ \hline
\textbf{D1}  & 94.30 (94.30) & 98.10 (98.10)  	\\	\hline
\textbf{D2} & 96.84 (96.20) & 96.84 (96.20)	\\	\hline
\textbf{D3}  & 97.47 (96.84) & 97.47 (97.47) 	\\	\hline
\textbf{DT1} & 97.47 (96.20) & 98.73 (98.10) 	\\	\hline
\textbf{DT2}  & 96.84 (96.20) & 99.37 (99.37) 	\\	\hline
\end{tabular}
\label{tab:mocapres100}
\end{center}
\end{table}

\begin{table}
\begin{center}
  \caption{Confusion table for motion capture dataset using \textbf{DT2} as the dynamical shape feature achieving mean classification rate of $99.37\%$ when compared to $89.7\%$ reported by Ali \textit{et al}. in \cite{ali2007chaotic}.}
  \begin{tabular}{| c | c | c | c | c | c |}
    \hline
    \textbf{\textit{Action}} & \textbf{Dance} & \textbf{Jump} & \textbf{Run} & \textbf{Sit} & \textbf{Walk}   	\\ \hline	\hline
    \textbf{Dance} & \textbf{30} & 1 & 0 & 0 & 0 	\\ \hline
    \textbf{Jump} & 0 & \textbf{14} & 0 & 0 & 0 		\\ \hline
    \textbf{Run} & 0 & 0 & \textbf{30} & 0 & 0 		\\ \hline
    \textbf{Sit} & 0 & 0 & 0 & \textbf{35} & 0		\\ \hline
    \textbf{Walk} & 0 & 0 & 0 & 0 & \textbf{48}		\\ \hline
  \end{tabular}  
  \label{tab:mocap confTable}
\end{center}
\end{table}

\begin{figure*}
\framebox{
		\begin{subfigure}[t]{0.1\textwidth}
                \centering
                \includegraphics[scale=0.75]{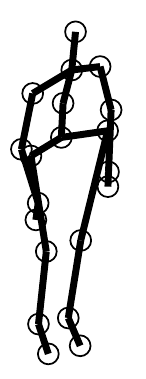}
        \end{subfigure}
        \begin{subfigure}[t]{0.15\textwidth}
                \centering
                \includegraphics[scale=0.75]{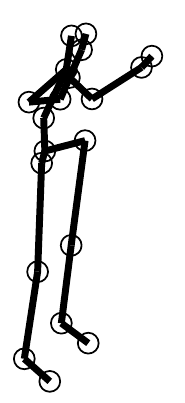}
        \end{subfigure}                
        \begin{subfigure}[t]{0.1\textwidth}
                \centering
                \includegraphics[scale=0.75]{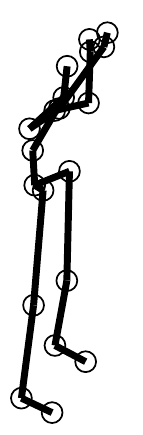}
        \end{subfigure}
        \begin{subfigure}[t]{0.1\textwidth}
                \centering
                \includegraphics[scale=0.75]{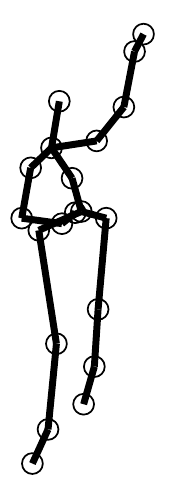}
        \end{subfigure}                
        \begin{subfigure}[t]{0.15\textwidth}
                \centering
                \includegraphics[scale=0.75]{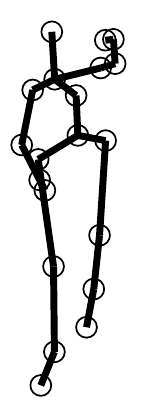}
        \end{subfigure}       
                \begin{subfigure}[t]{0.1\textwidth}
                \centering
                \includegraphics[scale=0.75]{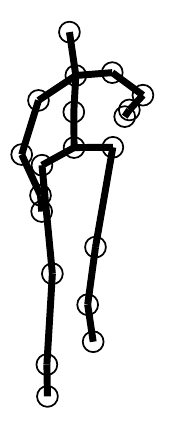}
        \end{subfigure} 
        \begin{subfigure}[t]{0.1\textwidth}
                \centering
                \includegraphics[scale=0.75]{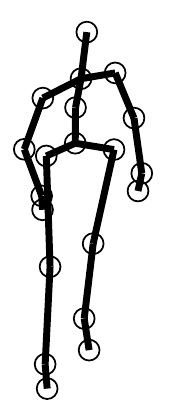}
        \end{subfigure} 
        \begin{subfigure}[t]{0.1\textwidth}
                \centering
                \includegraphics[scale=0.75]{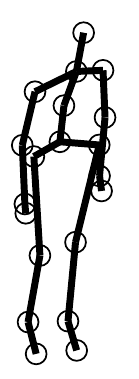}
        \end{subfigure}         
}
\\ \\
		\begin{subfigure}[t]{0.19\textwidth}
                \includegraphics[width=\textwidth]{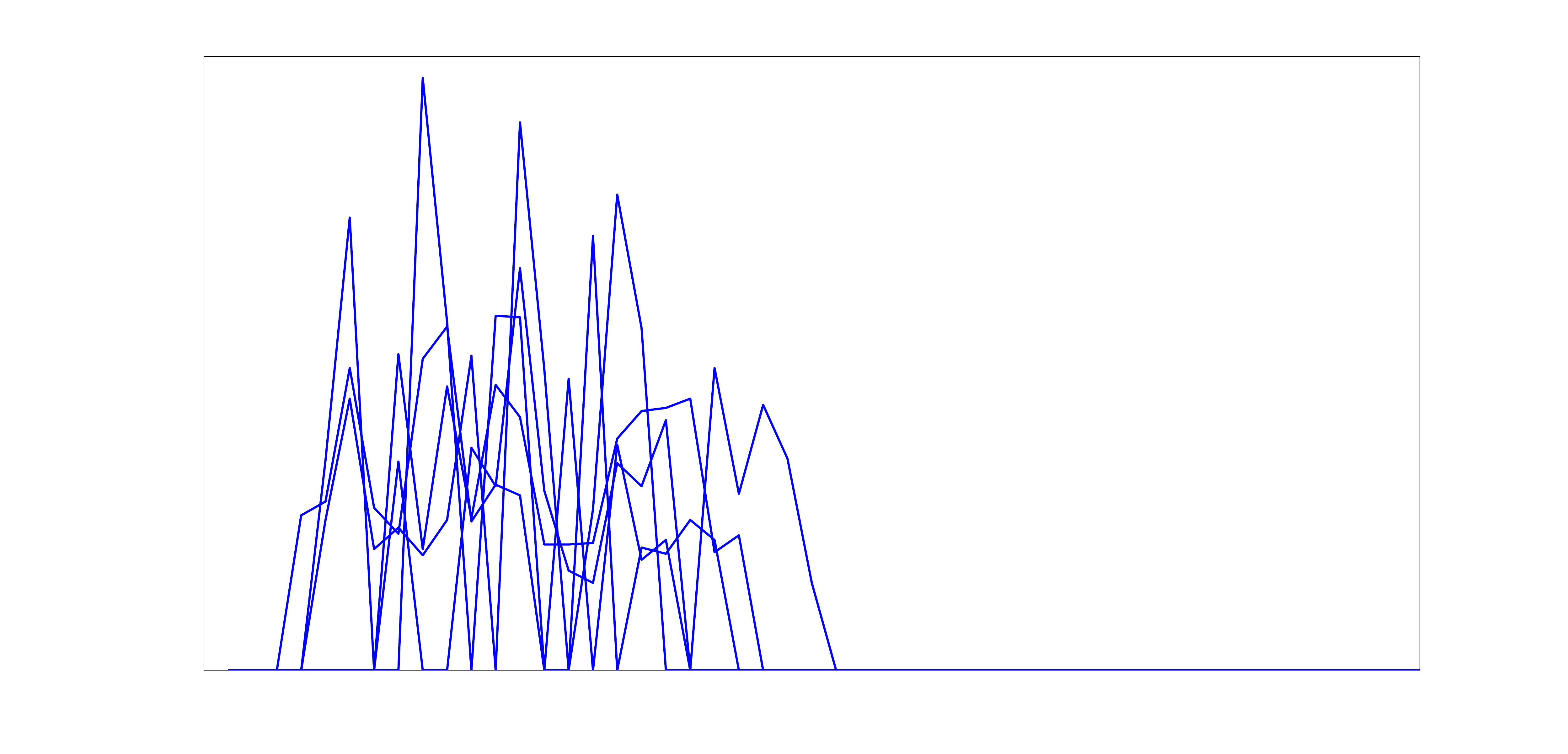}
                \caption{\textbf{D1}}
        \end{subfigure}
        		\begin{subfigure}[t]{0.19\textwidth}
                \includegraphics[width=\textwidth]{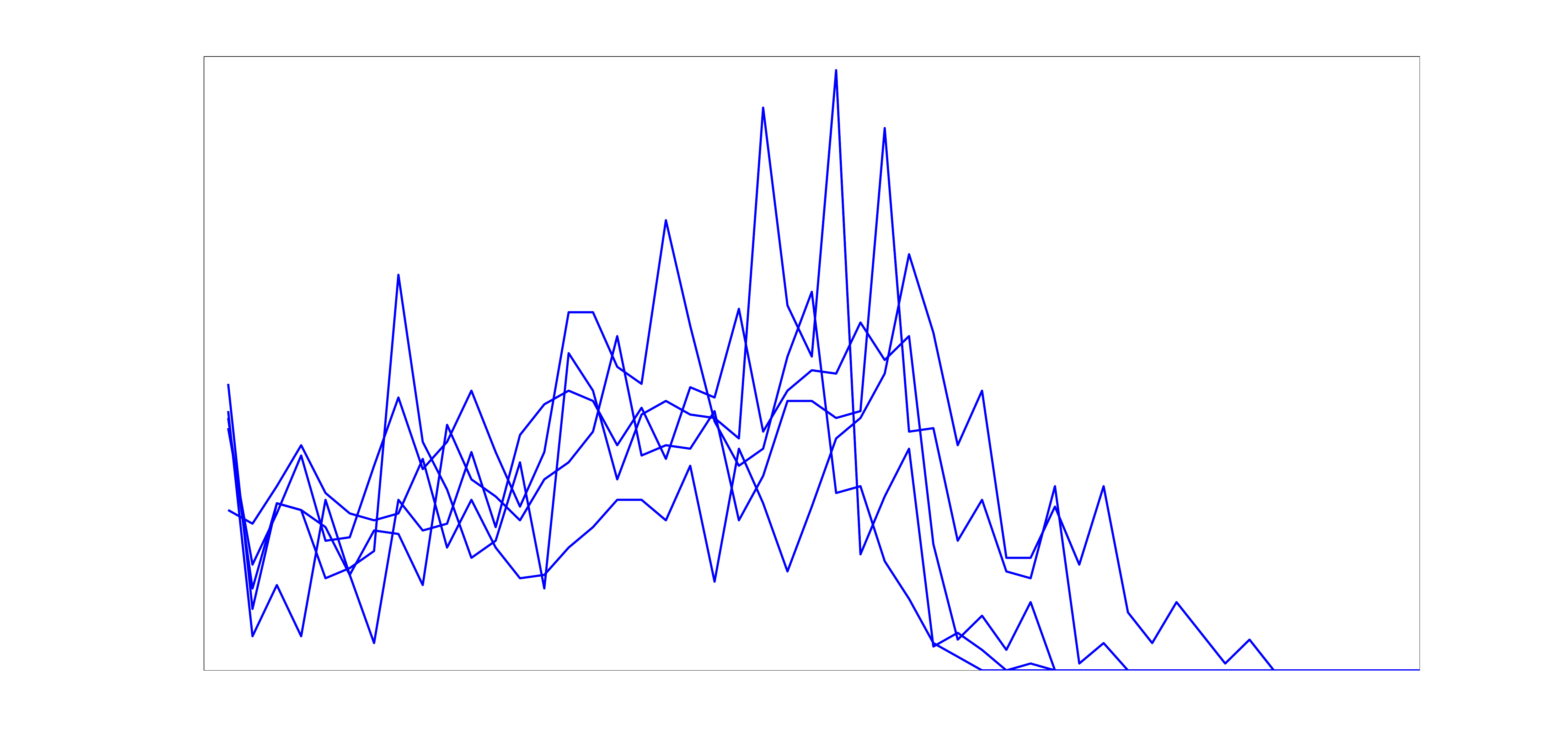}
                \caption{\textbf{D2}}
        \end{subfigure}
        		\begin{subfigure}[t]{0.19\textwidth}
                \includegraphics[width=\textwidth]{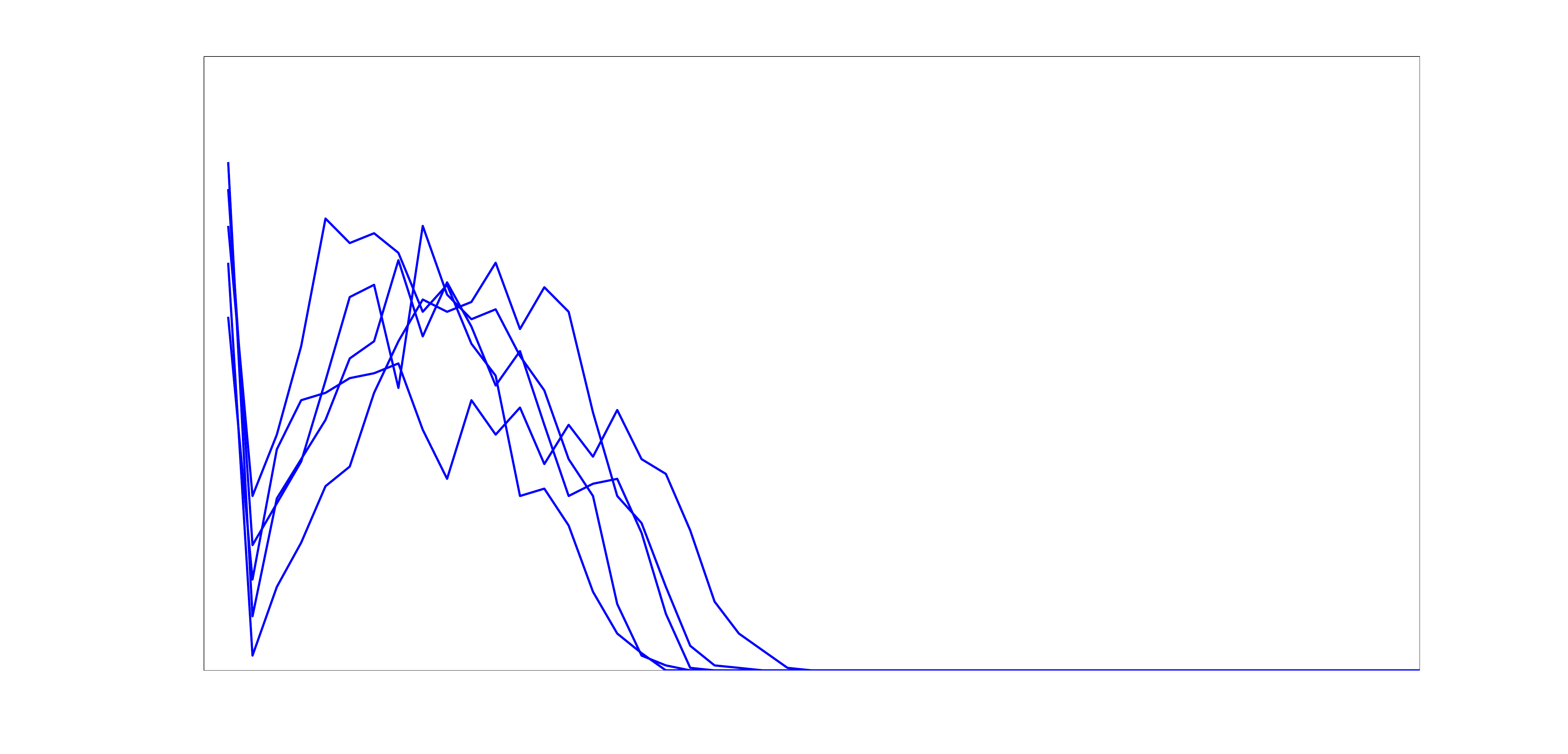}
                \caption{\textbf{D3}}
        \end{subfigure}
        		\begin{subfigure}[t]{0.19\textwidth}
                \includegraphics[width=\textwidth]{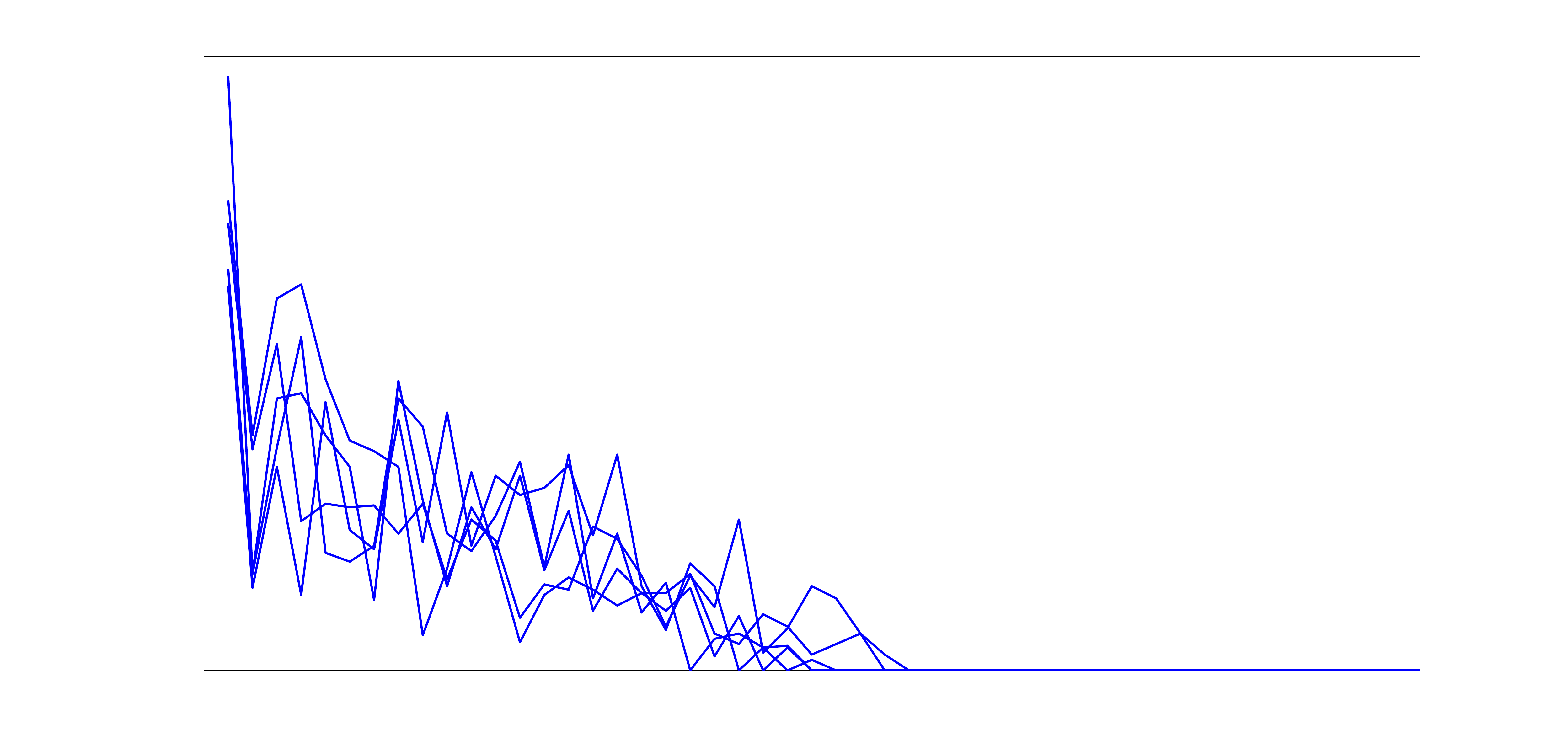}
                \caption{\textbf{DT1}}
        \end{subfigure}
        		\begin{subfigure}[t]{0.19\textwidth}
                \includegraphics[width=\textwidth]{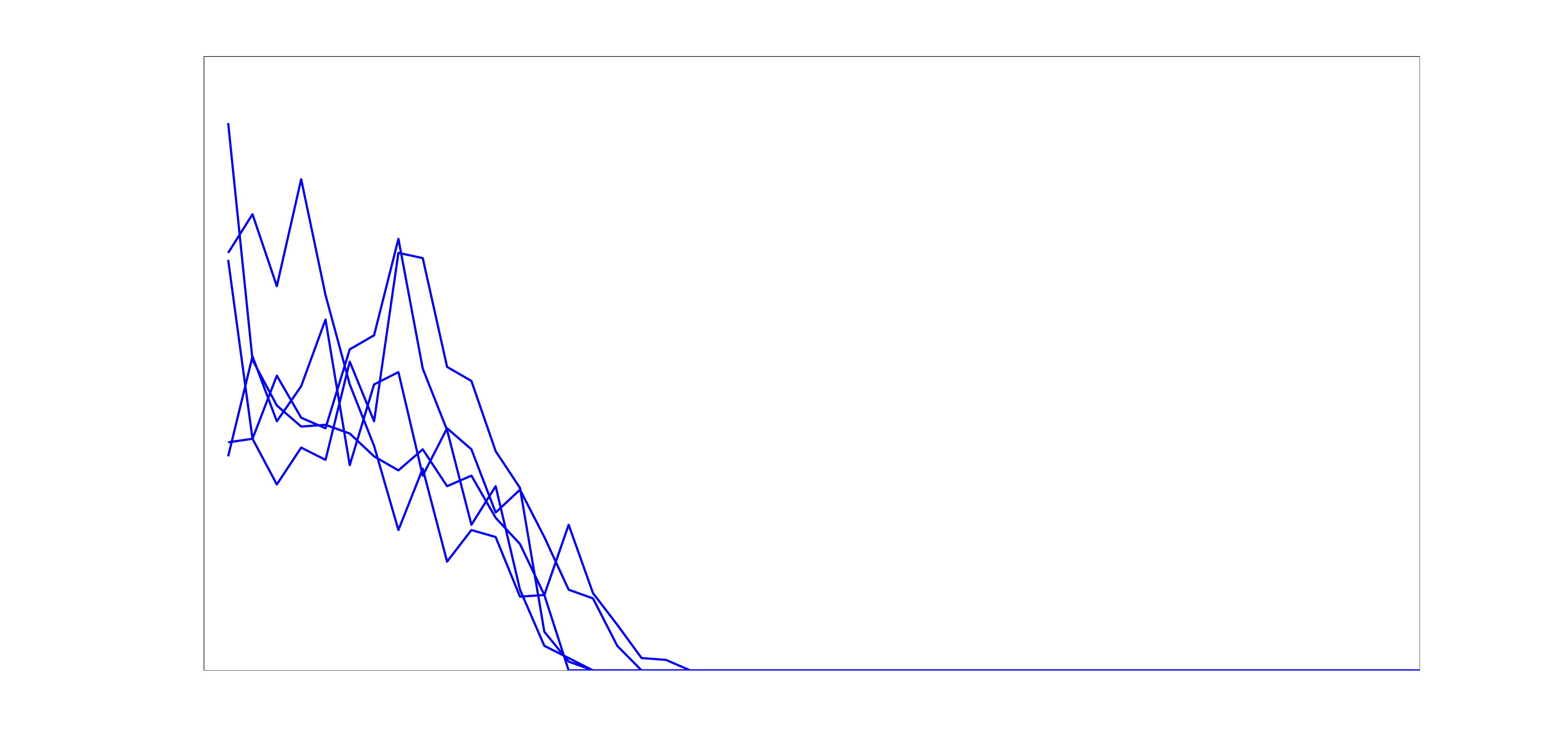}
                \caption{\textbf{DT2}}
        \end{subfigure}
\\ \\       
\framebox{
\begin{subfigure}[t]{0.1\textwidth}
               \centering
                \includegraphics[scale=0.75]{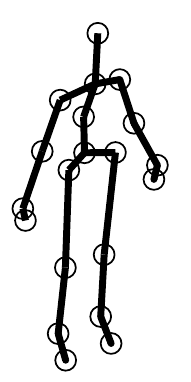}
        \end{subfigure}
        \begin{subfigure}[t]{0.15\textwidth}
                \centering
                \includegraphics[scale=0.75]{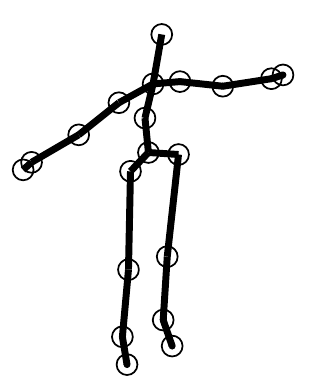}
        \end{subfigure}                
        \begin{subfigure}[t]{0.1\textwidth}
                \centering
                \includegraphics[scale=0.75]{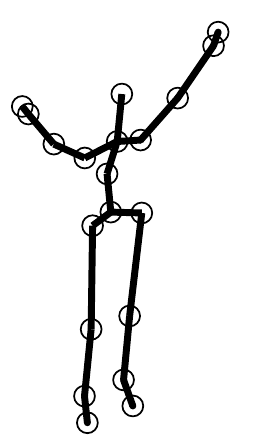}
        \end{subfigure}
        \begin{subfigure}[t]{0.1\textwidth}
                \centering
                \includegraphics[scale=0.75]{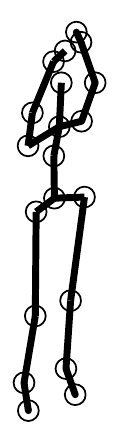}
        \end{subfigure}                
        \begin{subfigure}[t]{0.1\textwidth}
                \centering
                \includegraphics[scale=0.75]{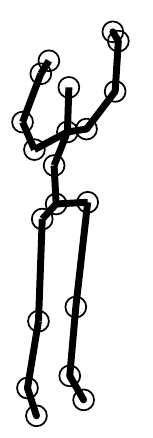}
        \end{subfigure}       
                \begin{subfigure}[t]{0.15\textwidth}
                \centering
                \includegraphics[scale=0.75]{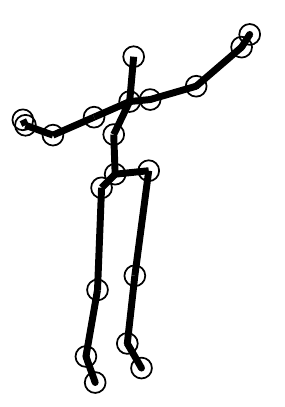}
        \end{subfigure} 
        \begin{subfigure}[t]{0.1\textwidth}
                \centering
                \includegraphics[scale=0.75]{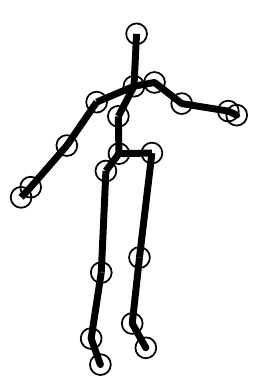}
        \end{subfigure} 
                \begin{subfigure}[t]{0.1\textwidth}
                \centering
                \includegraphics[scale=0.75]{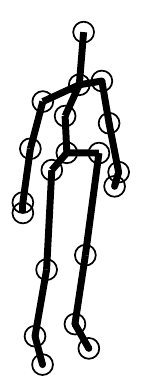}
        \end{subfigure} 
        }
\\ \\         
		\begin{subfigure}[t]{0.19\textwidth}
                \includegraphics[width=\textwidth]{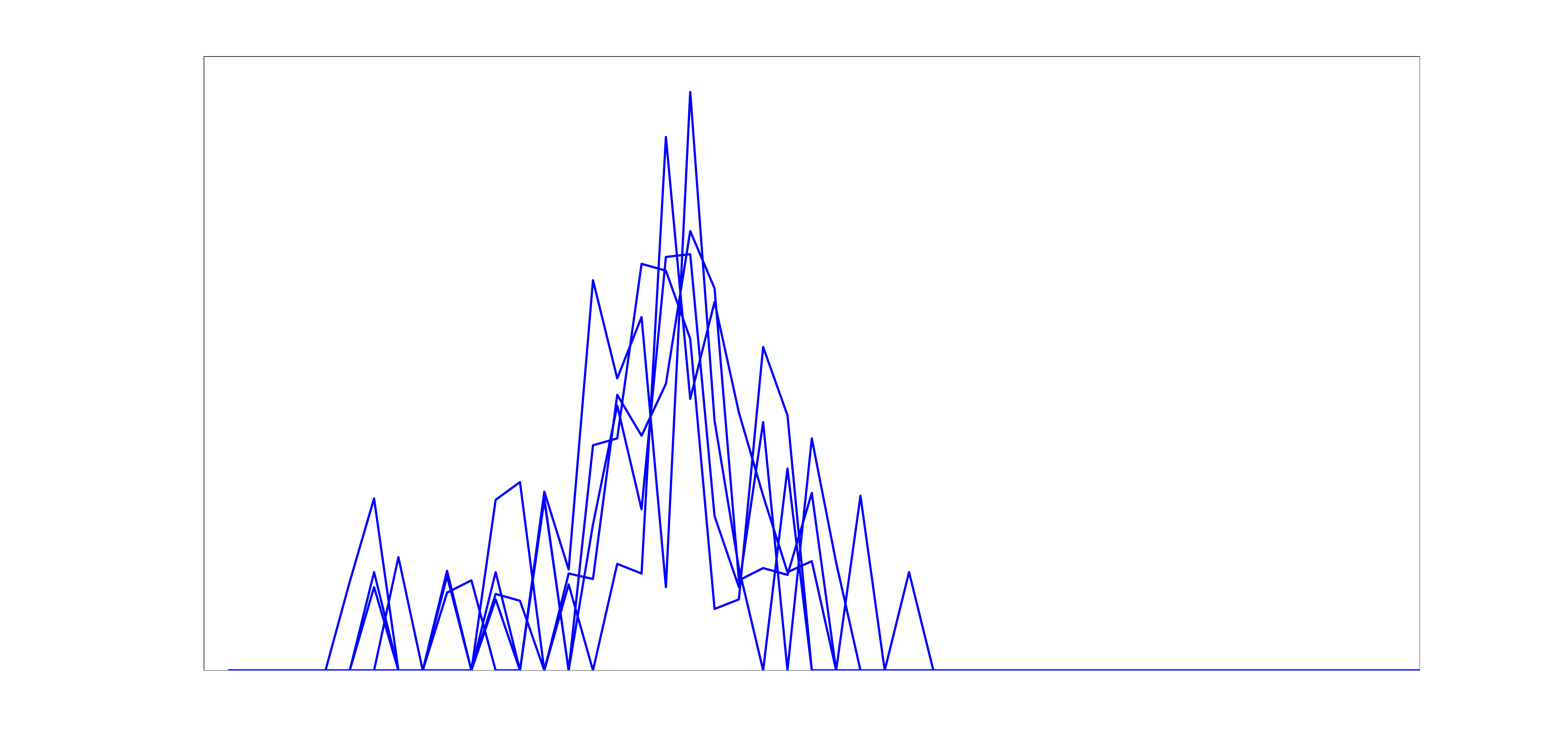}
                \caption{\textbf{D1}}
        \end{subfigure}
        		\begin{subfigure}[t]{0.19\textwidth}
                \includegraphics[width=\textwidth]{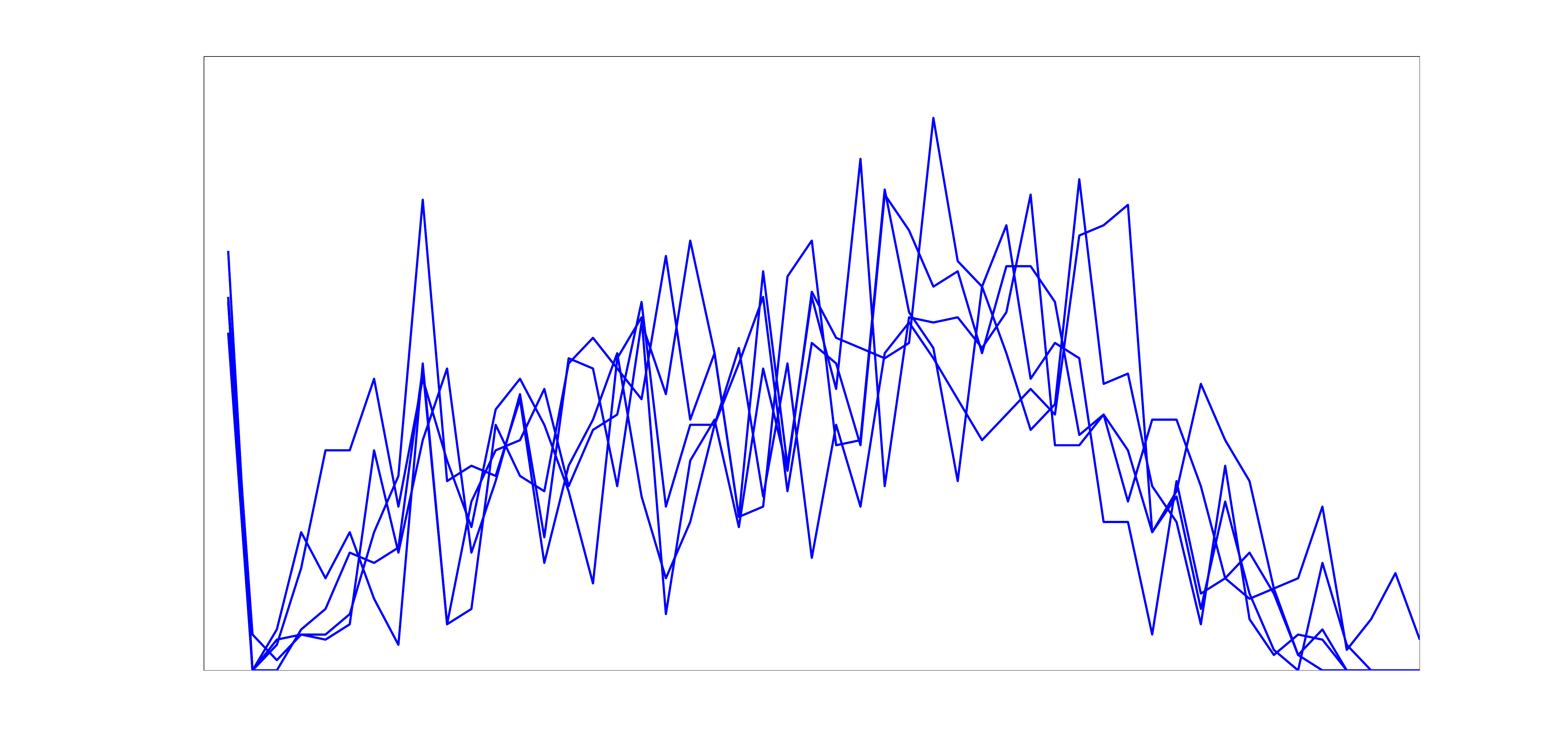}
                \caption{\textbf{D2}}
        \end{subfigure}
        		\begin{subfigure}[t]{0.19\textwidth}
                \includegraphics[width=\textwidth]{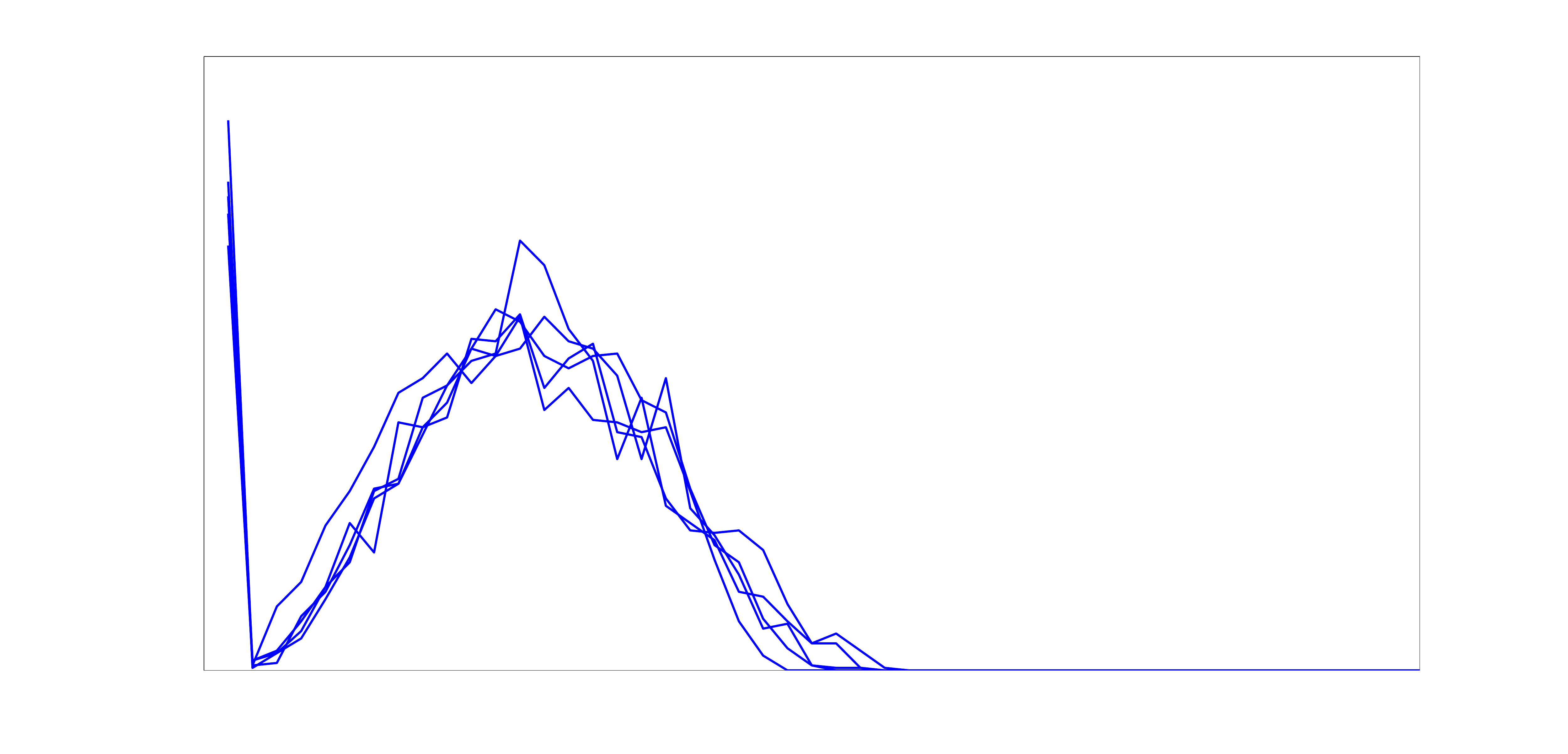}
                \caption{\textbf{D3}}
        \end{subfigure}
        		\begin{subfigure}[t]{0.19\textwidth}
                \includegraphics[width=\textwidth]{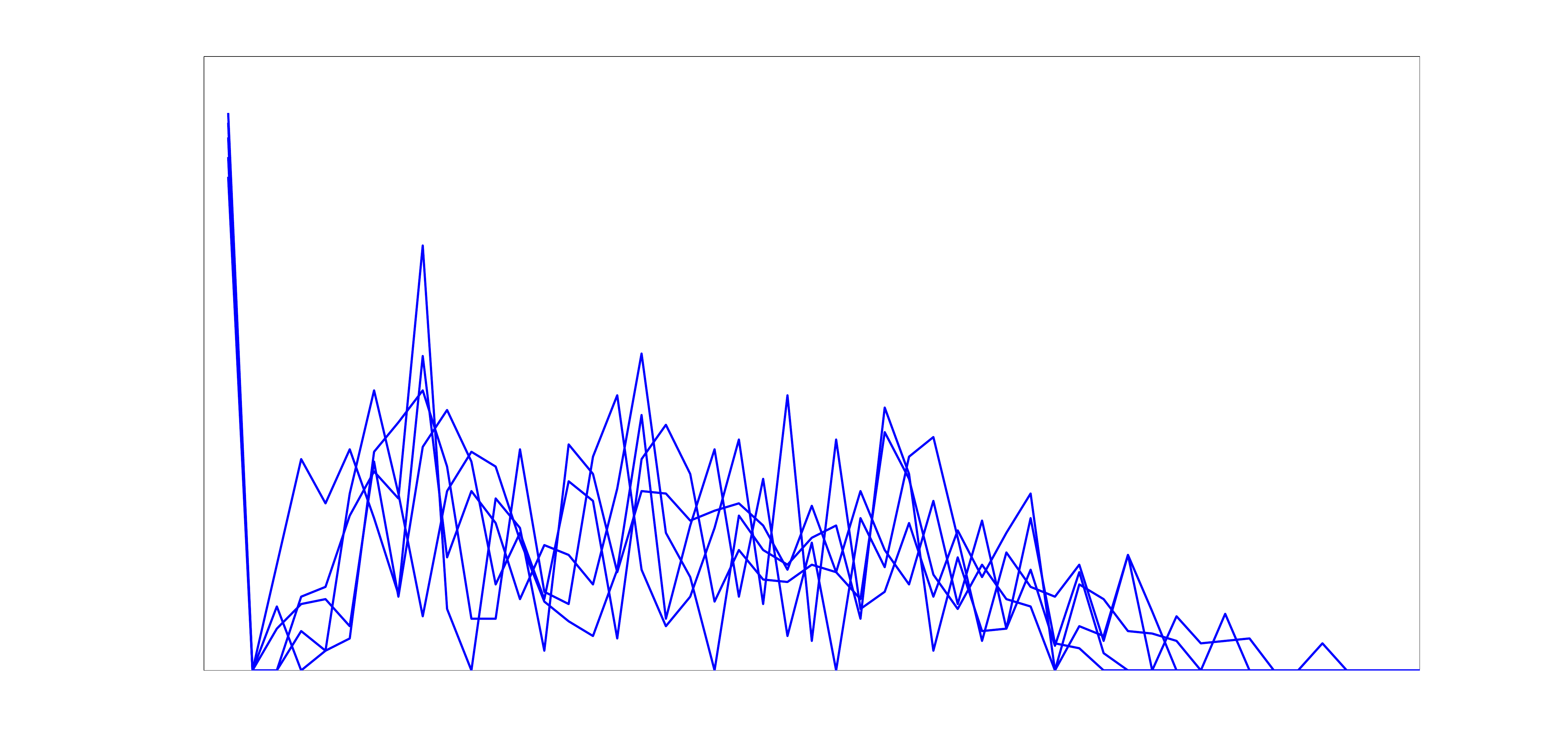}
                \caption{\textbf{DT1}}
        \end{subfigure}
        		\begin{subfigure}[t]{0.19\textwidth}
                \includegraphics[width=\textwidth]{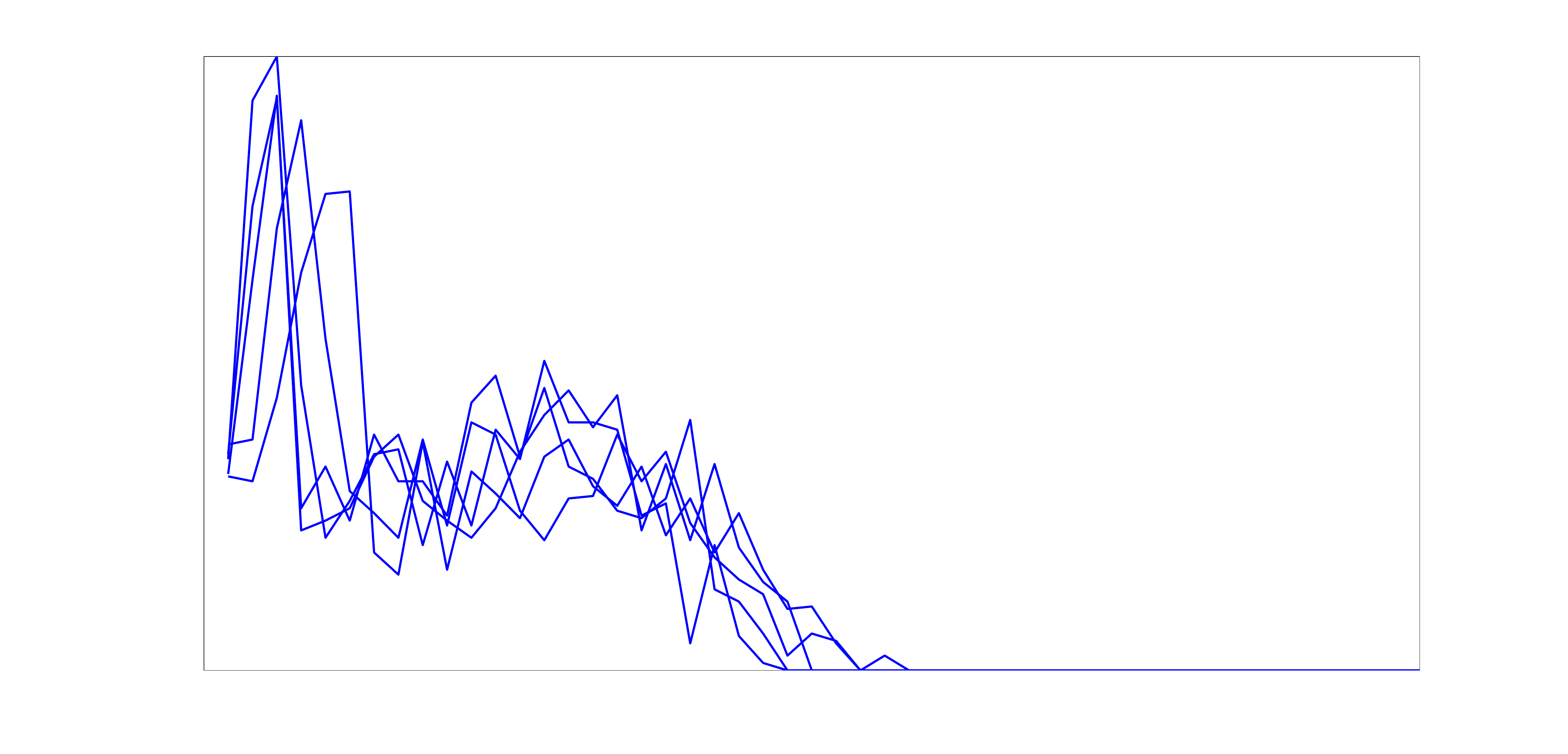}
                \caption{\textbf{DT2}}
        \end{subfigure}        
        \caption{Example actions from action class \textit{Tennis serve} (a) and \textit{Two hand wave} (b) from the MSR Action3D dataset. Skeleton data of 20 joints provided in the dataset will be used in our action recognition experiment. Shape distributions from reconstructed phase space using the hand trajectory from five instances each of tennis serve and two hand wave actions is shown here to illustrate the insensitivity of the framework to inter-class similarities.}
        \label{fig:kinectdataset}
\end{figure*}

\begin{table*}[!]
\begin{center}
\scriptsize
  \caption{Classification results for cross-subject test setting where $50\%$ subjects were used for training and the remaining $50\%$ subjects for testing in proposed method using linear SVM with $m = 3$ (and $m = 5$ in parentheses).}
  \begin{tabular}{| c | c | c | c | c | c | c | c |}
    \hline 
& \multicolumn{5}{c|}{\textbf{\textit{Shape Distribution}} } & \multicolumn{1}{c|}{\textbf{\textit{Chaos}}}  \\ \hline
\textbf{Set}  & \textbf{D1} & \textbf{D2} & \textbf{D3} & \textbf{DT1} & \textbf{DT2} &  \\ \hline \hline
\textbf{AS1} & 88.35 (86.14) & 89.32 (87.13) & 87.13 (86.41) & 88.57 (87.38) & 90.48 (89.58) & 72.28 (74.56)\\ \hline
\textbf{AS2} & 69.72 (63.39) & 72.65 (69.75) & 71.43 (72.32) & 73.21 (73.50) & 74.11 (70.00) & 51.85 (52.40)\\ \hline
\textbf{AS3} & 90.74 (84.68) & 96.40 (93.69) & 98.20 (96.43) & 98.25 (92.92) & 99.09 (96.49) & 76.36 (78.86)\\ \hline	\hline
\textbf{Avg.} & \textbf{82.94 (78.07)} & \textbf{86.12 (83.52)} & \textbf{85.59 (85.05)} & \textbf{86.68 (84.60)} & \textbf{87.89 (85.34)} & 66.83 (68.61)\\ \hline
\end{tabular}  
  \label{tab:kinectdatabse 3 sets}
\end{center}
\end{table*}

\subsection{Kinect Dataset}
\label{Kinectdata}
The framework was also evaluated on a more comprehensive dataset released by Microsoft Research called \textit{MSR Action3D} dataset \cite{li2010action} having $20$ action classes: \textit{high arm wave, horizontal arm wave, hammer, hand catch, forward punch, high throw, draw x, draw tick, draw circle, hand clap, two hand wave, side boxing, bend, forward kick, side kick, jogging, tennis swing, tennis serve, golf swing, pick up \& throw} with $10$ subjects performing each action thrice (see Fig. \ref{fig:kinectdataset} for example actions). The action classes in this dataset were selected to ensure the use of arms, legs and torso by subjects to simulate interaction with gaming consoles. High similarity between classes (e.g., forward punch and hammer, high throw and pickup \& throw) makes this a challenging dataset. The $20$ action classes were further divided into $3$ Action Sets: \textit{AS1}, \textit{AS2} and \textit{AS3} in \cite{li2010action} to account for the large amount of computation involved in classification of these actions. The action sets $1$ and $2$ were intended to group actions with similar movement and action set $3$ to group complex movements. The dataset provides $3$D joint positions on which phase space reconstruction and extraction of shape distribution were carried out individually on every dimension ($x, y$ \& $z$). These shape distributions were concatenated to form our feature vector representative of any given action. The classification results on the cross-subject test setting using a linear SVM are tabulated in TABLE \ref{tab:kinectdatabse 3 sets} and as seen, the proposed framework performs better than the traditional chaotic invariants. Examples shown in Fig. \ref{fig:kinectdataset} further support our hypothesis that shape distributions can be used as discriminative feature of reconstructed phase space representative of actions. In order to illustrate the proposed framework's stability to intra-class variations and insensitivity to inter-class similarities, we compare the dynamical shape features of hand trajectory for five instances of \textit{tennis serve} and \textit{two hand wave} action classes. Evident from these examples is that even actions using similar hand movements are represented by dynamical shape features with enough differences to successfully recognize these actions. Furthermore, from results in TABLE \ref{tab:kinectdatabse 3 sets}, we see that the dynamical shape feature \textbf{DT2} has the highest overall classification accuracy, indicating that the shape distribution based on temporal evolution of phase space is better than traditional global shape representations. We have also provided classification results using a nearest neighbor classifier in TABLE \ref{tab:kinectdatabse 3 sets nn} for a comprehensive comparison of the proposed shape distributions. Our results indicate that we achieve similar performance with both $m=3$ and $m=5$. In further evaluation experiments, we use $m=3$.  
 
\begin{table}
\begin{center}
\scriptsize
  \caption{Classification results for cross-subject test setting where $50\%$ subjects were used for training and the remaining $50\%$ subjects for testing in proposed method using nearest-neighbor classifier.}
  \begin{tabular}{| c | c | c | c | c | c | c | c |}
    \hline 
    & \multicolumn{5}{c|}{\textbf{\textit{Shape Distribution}} ($m = 3$)} & \multicolumn{2}{c|}{\textbf{\textit{Chaos}}}  \\ \hline
\textbf{Set}  & \textbf{D1} & \textbf{D2} & \textbf{D3} & \textbf{DT1} & \textbf{DT2} & $m = 3$ & $m = 5$ \\ \hline \hline
\textbf{AS1} & 67.00 & 74.62 & 75.73 & 75.05 & 78.43 & 52.30 & 55.67\\ \hline
\textbf{AS2} & 59.63 & 67.66 & 65.77 & 64.47 & 68.21 & 42.53 & 49.23\\ \hline
\textbf{AS3} & 87.83 & 89.96 & 89.66 & 88.11 & 91.13 & 53.45 & 60.59\\ \hline	\hline
\textbf{Avg.} & \textbf{71.49} & \textbf{77.41} & \textbf{77.05} & \textbf{75.87} & \textbf{79.25} & 49.43	& 55.16\\ \hline
\end{tabular}  
  \label{tab:kinectdatabse 3 sets nn}
\end{center}
\end{table}

\begin{figure*}
\centering
\begin{picture}(0,0)
\put(-10,-18){\rotatebox{90}{\mbox{Unimpaired}}}
\end{picture}
		\begin{subfigure}[p]{0.24\textwidth}
                \includegraphics[width=\textwidth]{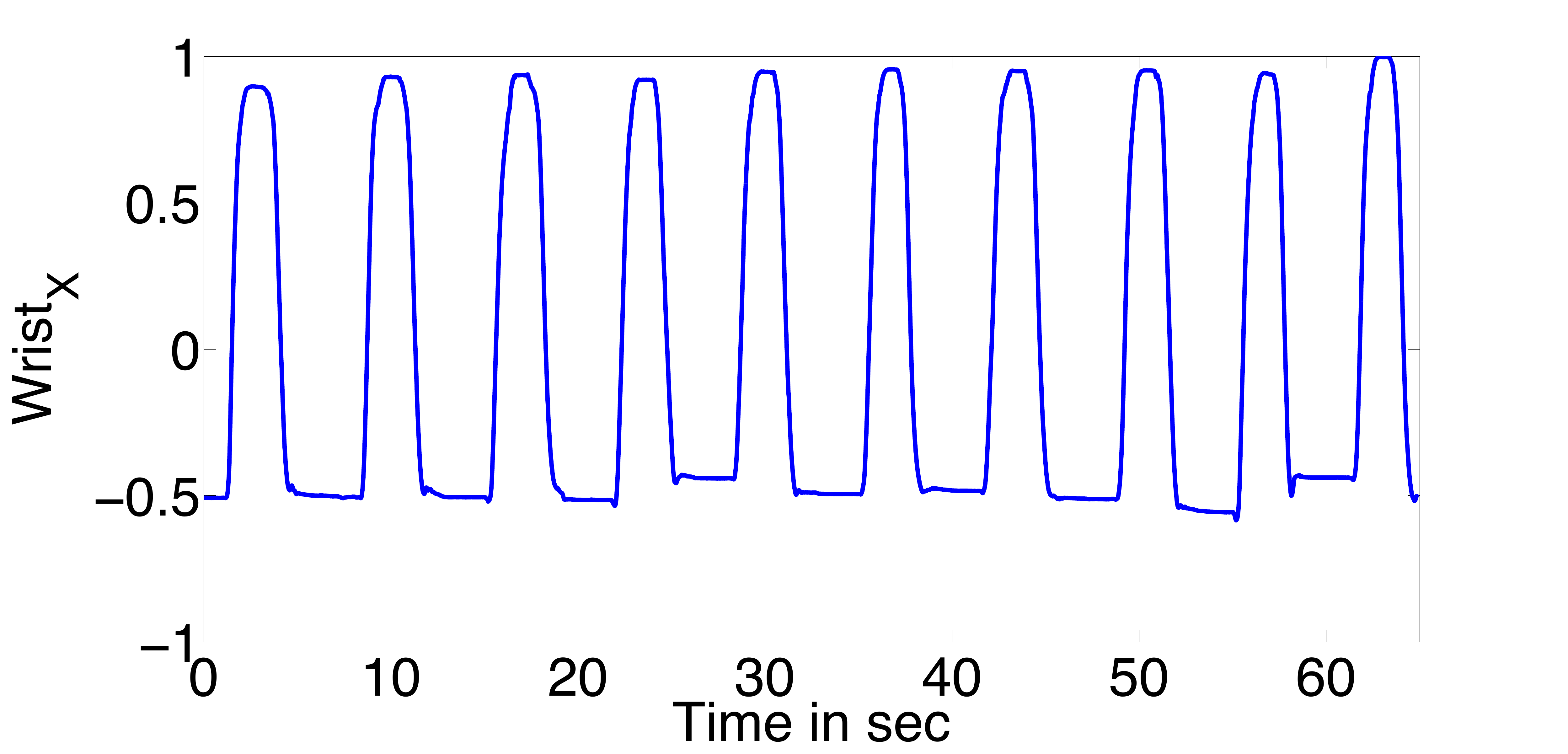}
        \end{subfigure}%
        \begin{subfigure}[p]{0.24\textwidth}
                \includegraphics[width=\textwidth]{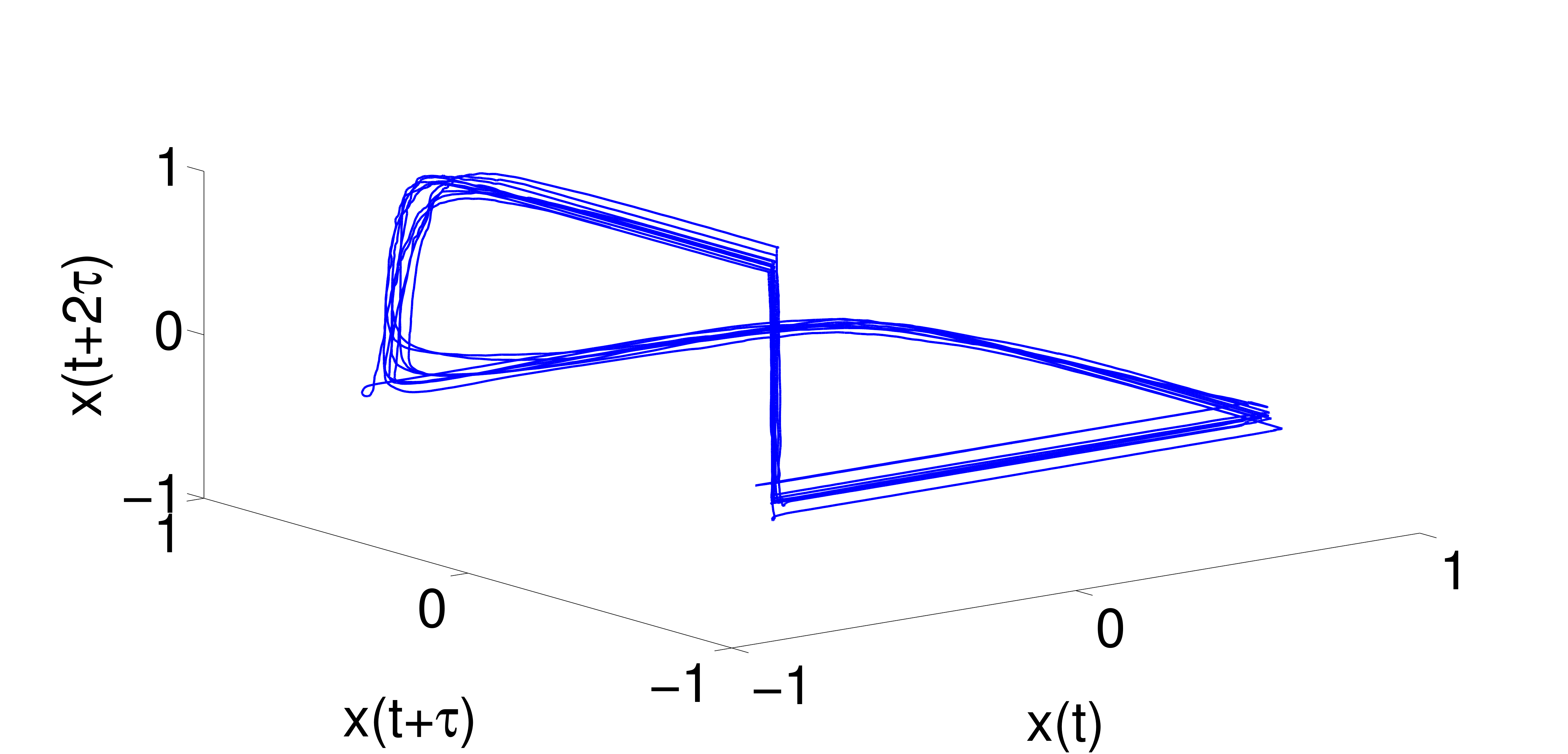}
        \end{subfigure}        
        \begin{subfigure}[p]{0.24\textwidth}
                \includegraphics[width=\textwidth]{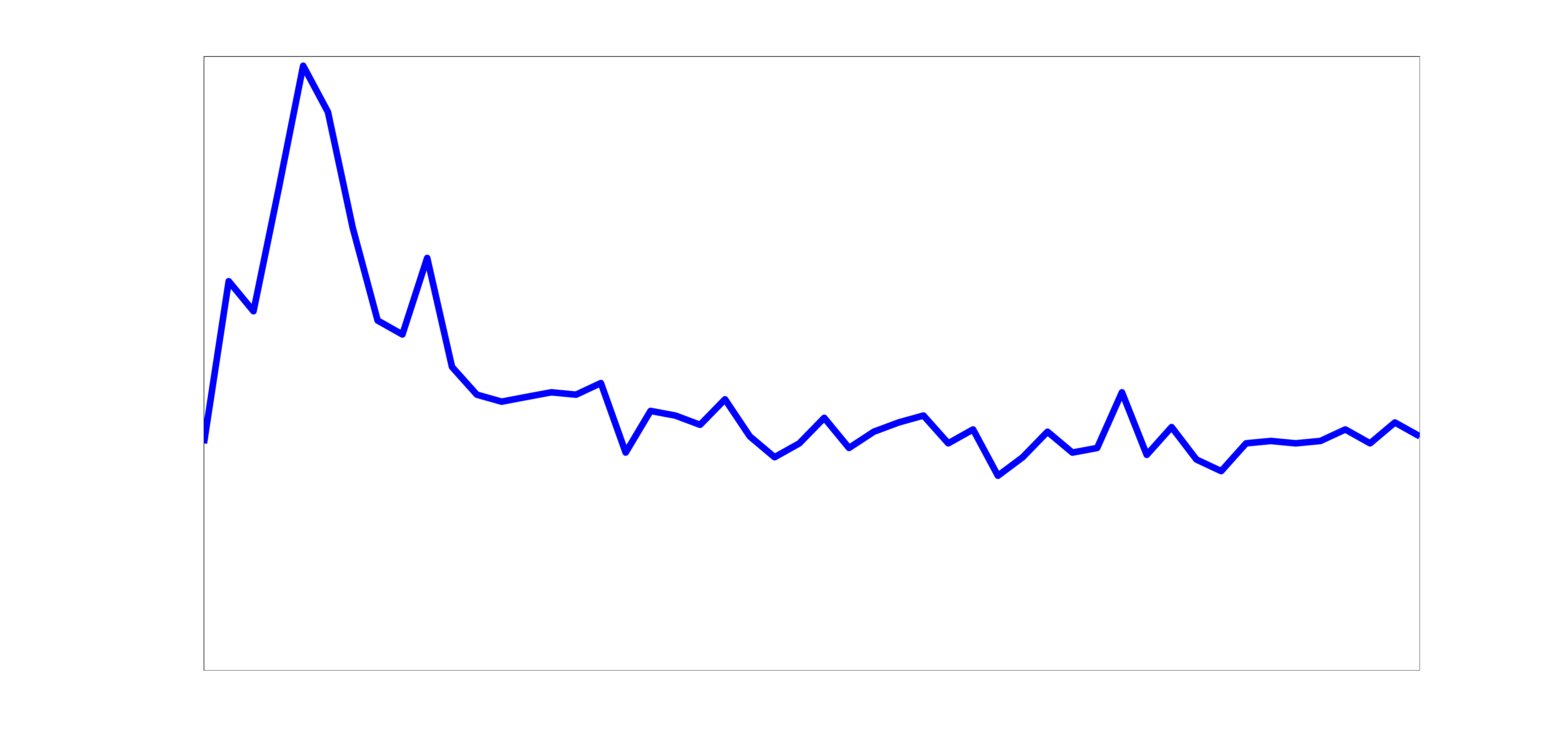}
        \end{subfigure} 
        \begin{picture}(100,40)
        \thicklines
			\put(10,0){\vector(1,-1){30}}           
		\end{picture}
		\begin{picture}(0,-38)
		\put(-50,-35){\fbox{\parbox{0.1\textwidth}{\centering
			Similarity\\			
			Measure
			}}}
		\end{picture}
		
		\begin{picture}(0,0)
		\put(-7,-4){\rotatebox{90}{\mbox{Impaired}}}
		\end{picture}
		\begin{subfigure}[p]{0.24\textwidth}
                \includegraphics[width=\textwidth]{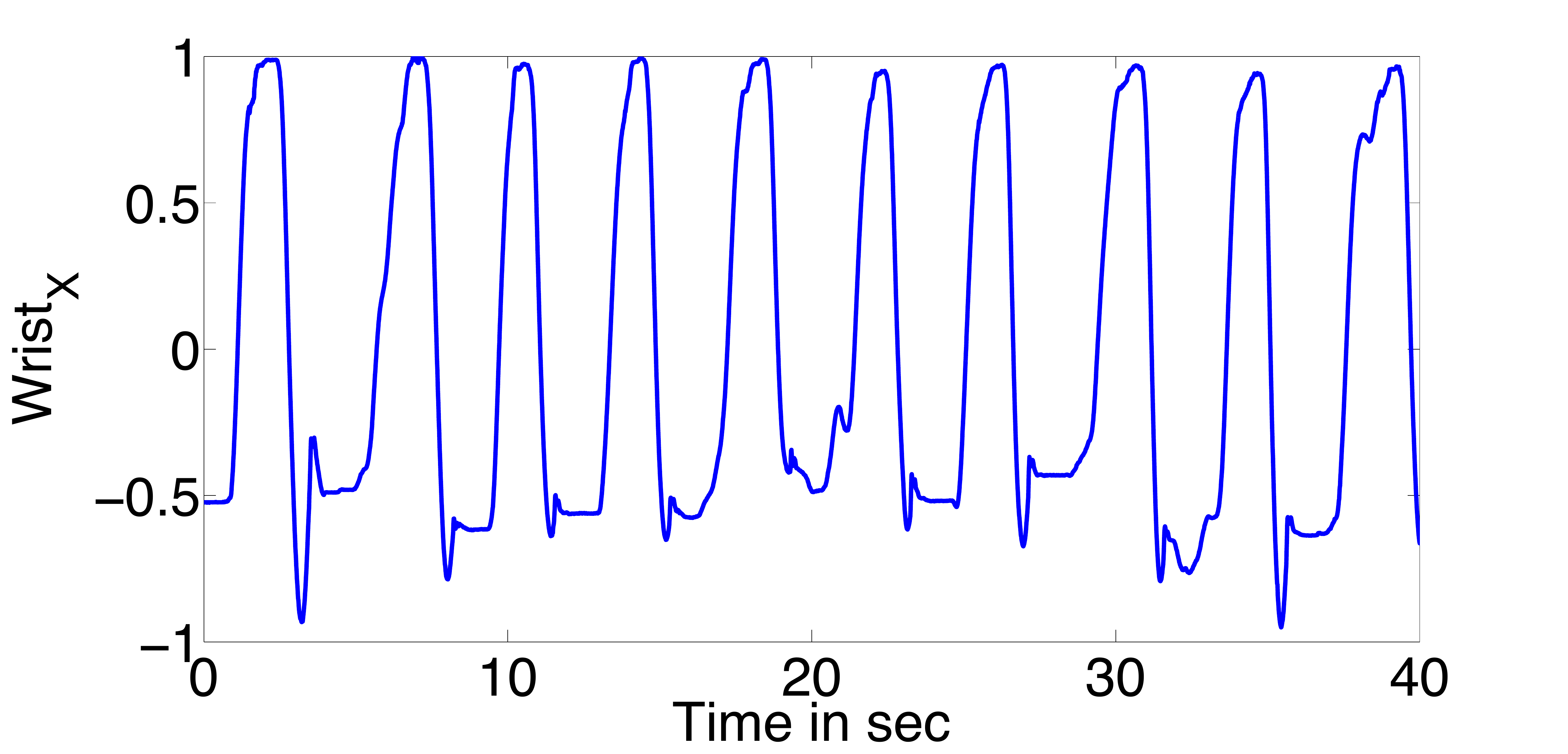}
                \caption{Time-series data}
        \end{subfigure} 
		\begin{subfigure}[p]{0.24\textwidth}
                \includegraphics[width=\textwidth]{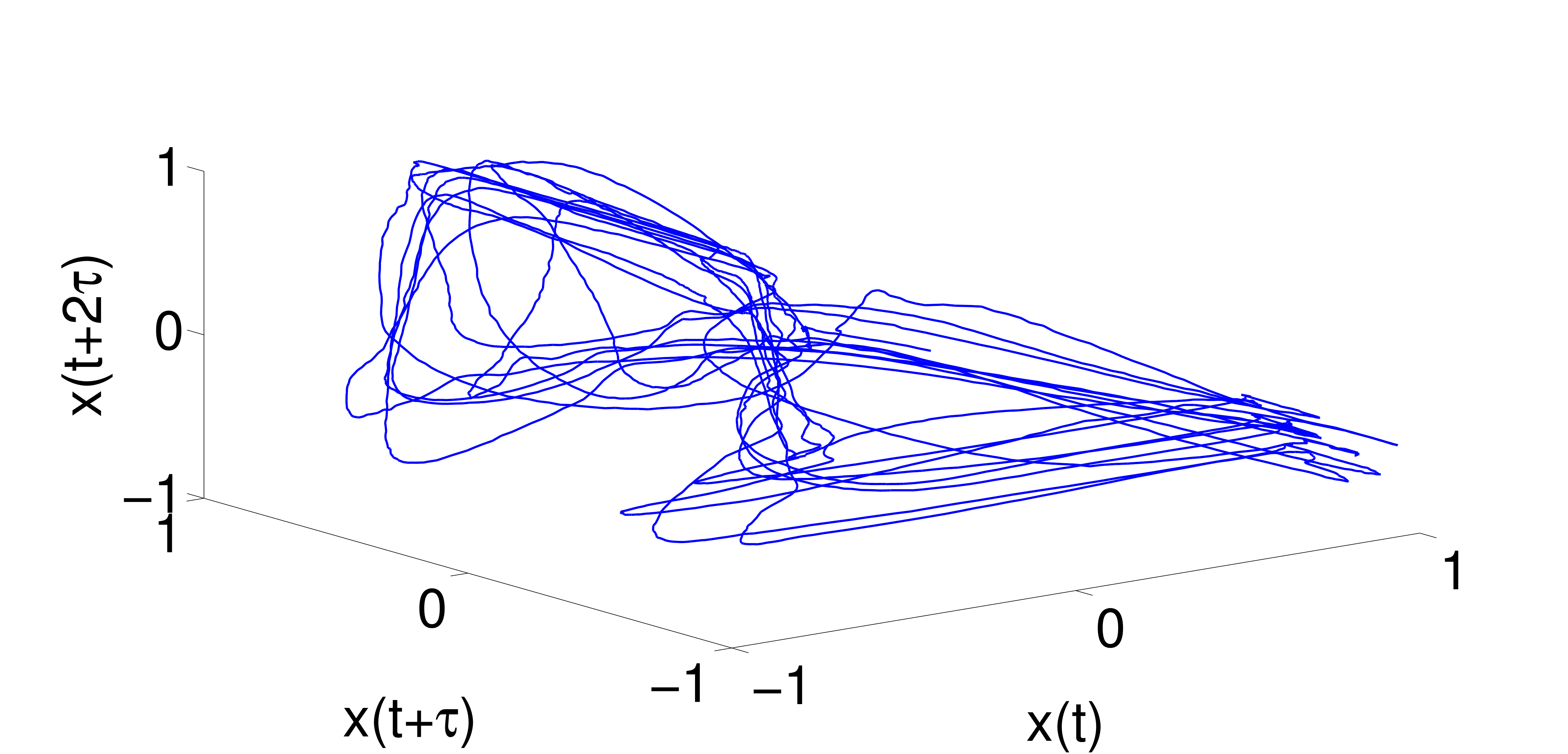}
                \caption{Reconstructed phase space}                
        \end{subfigure} 
		\begin{subfigure}[p]{0.24\textwidth}
                \includegraphics[width=\textwidth]{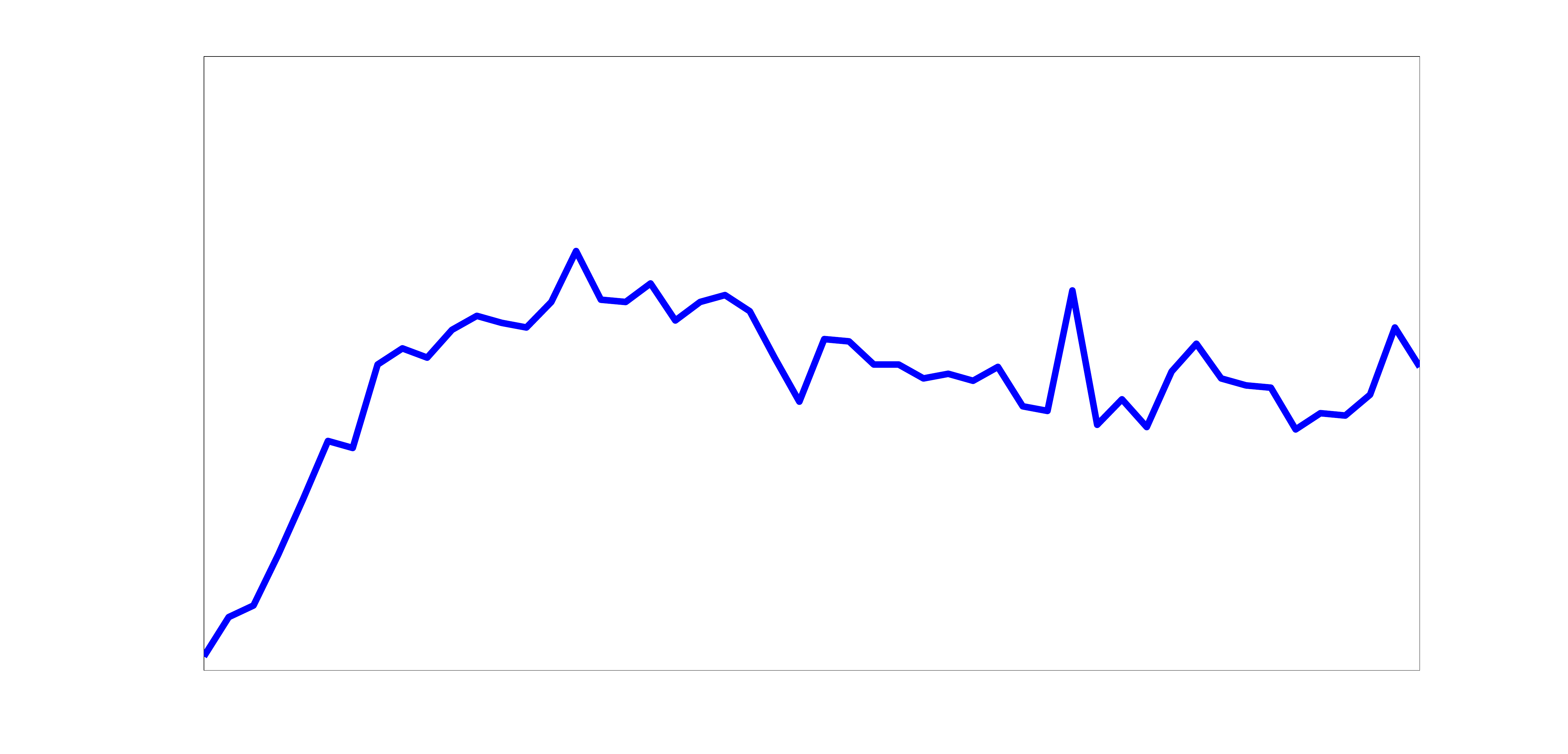}
                \caption{Shape distribution}                
        \end{subfigure}      
        \begin{picture}(100,40)
        \thicklines
			\put(10,0){\vector(1,1){30}}           
		\end{picture}
		\begin{picture}(2,2)
			\put(15,0){\line(1,1){0}}           
		\end{picture}
        \caption{Proposed framework for movement quality assessment and action recognition by extraction of dynamical shape feature from reconstructed phase space. (a) shows the time-series of $x$-location of wrist marker; its respective reconstructed phase space is shown in (b). These two exemplar trajectories are collected from the stroke rehabilitation dataset \cite{chen2011computational} and belong to unimpaired and impaired subjects respectively. The corresponding dynamical shape feature represented by shape distribution is shown in (c). Similarity measure (e.g., Euclidean distance) can be used to classify these trajectories.}
        \label{fig:PhaseSpaceStrokeData}        
\end{figure*}
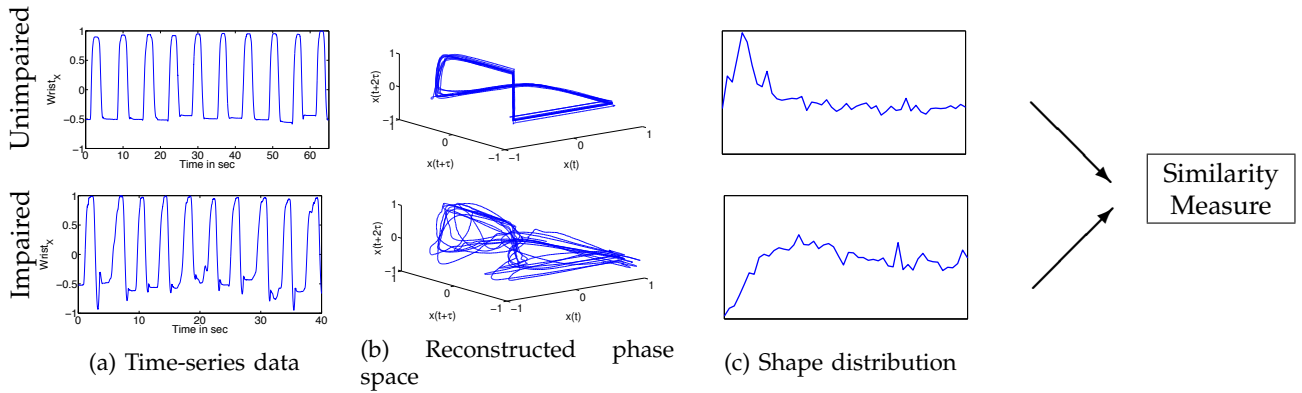

\subsection{Activity Quality for Stroke Rehabilitation}
\label{strokedata}
Our aim in this experiment is two-fold: a) to classify movements of unimpaired (neurologically normal) and impaired (stroke survivors) subjects, b) to quantitatively assess the quality of movement performed by the impaired subjects during repetitive task therapy. Fig. \ref{fig:PhaseSpaceStrokeData} illustrates the differences in shape of reconstructed phase space between unimpaired and impaired subjects using trajectories from the wrist marker (reflective marker placed on the subject's wrist). 
The experimental data was collected using a heavy marker-based system ($14$ markers on the right hand, arm and torso) in a hospital setting. Seven unimpaired and $15$ impaired subjects perform multiple repetitions of reach and grasp movements, both on-table and elevated (the subject must move against gravity to reach the target). Each subject would perform $4$ sets of reach and grasp movements to different target locations, with each set having $10$ repetitions. To account for a small number of training examples, we adopt leave-one-reach-out cross validation scheme where one set of reach movement was used as testing example and rest as training examples. The stroke survivors were also evaluated by the Wolf Motor Function Test (WMFT) \cite{wolf2001assessing} on the day of recording, which evaluates the subject's functional ability on a scale of $1-5$ (with $5$ being least impaired and $1$ being most impaired) based on predefined functional tasks. Since our focus is on development of quantitative measures of movement quality for a home-based rehabilitation system that would use a single marker on the wrist, we only use the data corresponding to the single marker on the wrist from the heavy marker-based hospital system. 

The focus of traditional methods for quantitative assessment of movement quality has been towards kinematics. Hence, in TABLE \ref{tab:strkcomptab}, we compare our results with an approach which uses kinematic analysis on the same dataset \cite{chen2011computational}. We also compare our results with the performance of traditional chaotic invariants. It is evident from these results that our framework performs better than the two promising quantitative measures for movement analysis in the field of stroke rehabilitation.

\begin{table}[t]
\begin{center}
\caption{Comparison of classification rates for different methods using leave-one-reach-out cross-validation and nearest neighbor classifier on the stroke rehabilitation dataset.}
\begin{tabular}{|c|c|}
\hline
\textbf{Method} & \textbf{Classification Rate (\%)} 	\\ \hline \hline
 KIM \cite{chen2011computational} & 85.2 	\\ 
Chaos ($m = 3$) & 81.82 	\\ 
Chaos ($m = 5$) & 83.43 	\\ \hline 
\textbf{D1} ($m = 3$) & \textbf{84.32}		\\ 
\textbf{D2} ($m = 3$) & \textbf{88.60} 		\\ 
\textbf{D3} ($m = 3$) & \textbf{86.04} 		\\ 
\textbf{DT1} ($m = 3$) & \textbf{87.65} 		\\ 
\textbf{DT2} ($m = 3$) & \textbf{92.05} 		\\ \hline
\end{tabular}
\label{tab:strkcomptab}
\end{center}
\end{table}

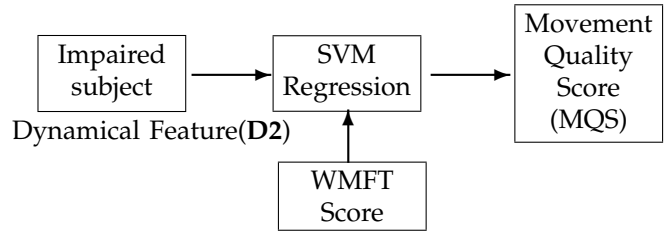
\begin{figure}[t]
\vspace{25pt}
\begin{picture}(0,0)
		\put(3,0){\fbox{\parbox{0.1\textwidth}{\centering
			Impaired\\			
			subject
			}}}
\end{picture}
\begin{picture}(0,0)
		\put(-10,-22){{\mbox{Dynamical Feature \\
		 (\textbf{D2})}}}
\end{picture}
\begin{picture}(0,0)
        \thicklines
			\put(54,2){\vector(1,0){30}}           
\end{picture}
\begin{picture}(0,0)
		\put(81,0){\fbox{\parbox{0.1\textwidth}{\centering
			SVM\\			
			Regression
			}}}
\end{picture}
\begin{picture}(0,0)
        \thicklines
			\put(106,-31){\vector(0,0){20}}           
\end{picture}
\begin{picture}(0,0)
		\put(76,-46){\fbox{\parbox{0.1\textwidth}{\centering
			WMFT\\			
			Score
			}}}
\end{picture}
\begin{picture}(0,0)
        \thicklines
			\put(130,2){\vector(1,0){30}}           
\end{picture}
\begin{picture}(0,0)
\thicklines
		\put(158,0){\fbox{\parbox{0.1\textwidth}{\centering
			Movement Quality\\			
			Score (MQS)
			}}}
\end{picture}
\vspace{2.cm}
\caption{Block diagram representation for learning a regressor for movement quality assessment using Functional Activity Score (FAS) from the Wolf Motor Function Test (WMFT).}
\label{fig:regression}
\end{figure}

\begin{figure}
\begin{center}
	\includegraphics[width=0.5\textwidth]{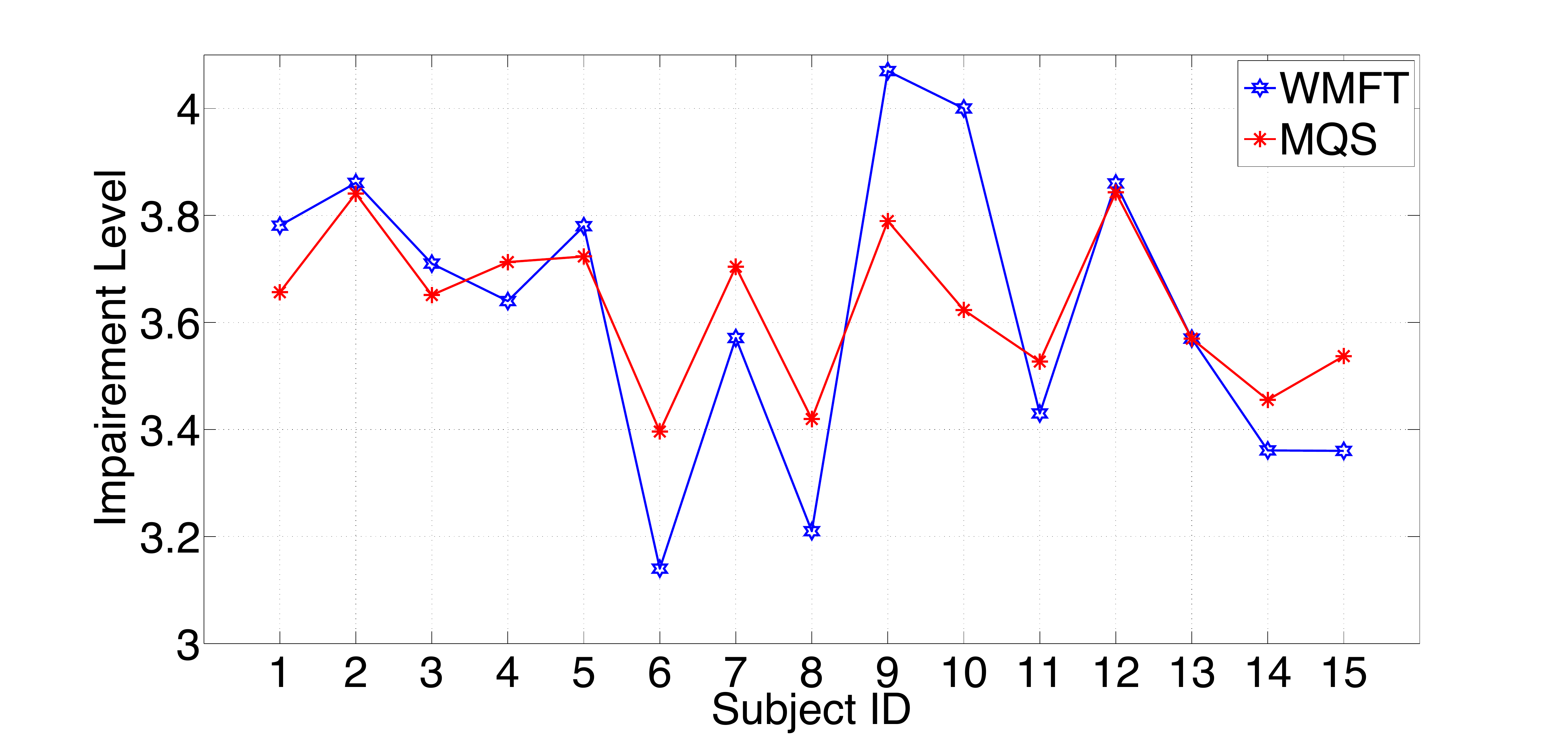}
	\caption{Comparison between impairment level (with $5$ being least impaired and $1$ being most impaired) given by actual WMFT score and MQS for $15$ impaired subjects. The Pearson correlation coefficient was found to be $0.8527$ with a two-tail P-value of $5.35\times 10^{-5}$, proving its statistical significance.}	
\label{fig:corr_stroke}
\end{center}
\end{figure}

We also propose a framework for movement quality assessment (shown in Fig. \ref{fig:regression}) for stroke rehabilitation. Using the WMFT scores of impaired subjects, we learn a regression function using SVM to compute a movement quality score from dynamical shape feature (using \textbf{D2} shape distribution). The regressor was trained using leave-one-reach-out cross-validation technique. The outputs of the regressor were averaged per subject to get the Movement Quality Score (MQS). Fig. \ref{fig:corr_stroke} shows a comparison between the actual WMFT score and the quality assessment score by the proposed method (MQS). The Pearson correlation coefficient between the MQS and the Function Activity Score (FAS) of the WMFT was found to be $0.8527$. When we repeat the same experiment with kinematic attributes on a single wrist marker, the correlation coefficient was found to be $0.6481$. In comparison, kinematic analysis of data from all $14$ markers gave a correlation coefficient of $0.9041$. This experiment clearly shows that the proposed framework achieves comparable results obtained by the heavy marker-based system even when using a single wrist marker, which is facilitated by the phase space reconstruction and robust feature extraction from phase space using shape distribution.

The WMFT scores are based on several functional tasks (e.g., folding a towel, picking up a pencil) and not on evaluation of the actual movements during repetitive therapy treatment (reach and grasp movements). In the above experiment, we utilize these WMFT scores as an approximate high-level quantitative measure for movement quality of impaired subjects performing reach and grasp movements, as both WMFT evaluation and $3$D marker data on the wrist were obtained on the same day. 

To address this conflict in collection of ground truth (movement quality labels) and trajectory data, we have collected a dataset from eight stroke survivors performing reach and grasp movement tasks and have developed a rating scale for movement quality in collaboration with physical therapists. Within this scale, physical therapists would provide us an overall rating on a scale of $1-5$ based on the therapist's impression of the participant's performance. A score of $1$ denotes that the participant could not complete the task (most impaired) and a $5$ denotes that the participant performed the task with the same quality of performance as the therapist if he/she were to perform it (least impaired or unimpaired). We have collected both $3$D position of the wrist and physical therapist ratings in order to make comparisons among the kinematics, our proposed measure, and the therapist ratings, across the same reach action. Utilizing the expert knowledge of the therapist ratings for these rated actions will also help us better contextualize the data to better shape our framework as a therapy tool. Using the same framework for regression as earlier, we see from TABLE \ref{HomeSysResults} that the proposed framework (using \textbf{DT2}) performs better than the traditional methods for movement quality assessment in terms of correlation coefficient and mean squared error. It should be noted that the proposed framework does not require data collected from unimpaired subjects for generating MQS, while kinematic methods like KIM \cite{chen2011computational} does, making the framework more suitable to model complex tasks during therapy treatment.  

\begin{table}
\begin{center}
\caption{Comparison of performance of the proposed dynamical shape features with the performance of traditional methods used for movement quality analysis.}
\begin{tabular}{| c | c | c |}
\hline
\textbf{Method} & \textbf{Correlation Coefficient} & \textbf{MSE}	\\ \hline \hline
\textbf{KIM} \cite{chen2011computational} & 0.4918 & 0.0066 \\ \hline
\textbf{Chaos ($m = 3$)} & 0.4717 & 0.0101 \\ \hline
\textbf{Chaos ($m = 5$)} & 0.5089 & 0.0100 \\ \hline
\textbf{D1} ($m = 3$)& 0.3877 & 0.1190 \\ \hline
\textbf{D2} ($m = 3$) & 0.5029 & 0.0078 \\ \hline
\textbf{D3} ($m = 3$) & 0.4935 & 0.0061 \\ \hline
\textbf{DT1} ($m = 3$) & 0.4582 & 0.0100 \\ \hline
\textbf{DT2} ($m = 3$) & 0.5510 & 0.0057 \\ \hline
\end{tabular}
\label{HomeSysResults}
\end{center}
\end{table}


\begin{table}
\tiny
\begin{center}
\caption{Comparison of classification rates for various approaches on the Maryland ``in-the-wild'' dataset (with $m = 3$).}
\begin{tabular}{| c || c | c | c | c | c | c | c |}
\hline
\textbf{Class} & \textbf{Chaos} \cite{movingvistas} & \textbf{Chaos} (our) \tablefootnote{Here ``our'' refers to our implementation of traditional chaotic invariants using the OpenTSTOOL package.} & \textbf{D1} & \textbf{D2} & \textbf{D3} & \textbf{DT1} & \textbf{DT2} \\ \hline \hline
avalanche & 30 & 40 & 0 & 0 & 20 & 10 & 0\\ \hline
b. water & 30 & 40 & 30 & 40 & 20 & 30 & 30\\ \hline
c. traffic & 50  & 30 & 80 & 100 & 50 & 60 & 90\\ \hline
f. fire & 30 & 20 & 10 & 30 & 30 & 30 & 30\\ \hline
fountain & 20 & 0 & 40 & 30 & 30 & 30 & 40\\ \hline
i. collapse & 10 & 0 & 10 & 0 & 0 & 10 & 0\\ \hline
landslide & 10 & 50 & 0 & 10 & 20 & 10 & 20 \\ \hline
s. traffic & 20 & 20 & 20 & 30 & 30 & 40 & 30\\ \hline
tornado & 60 & 10 & 40 & 70 & 60 & 50 & 60\\ \hline
v. eruption & 70 & 0 & 60 & 70 & 60 & 40 & 70\\ \hline
waterfall & 30 & 20 & 10 & 40 & 20 & 20 & 30\\ \hline
waves & 80 & 40 & 70 & 80 & 80 & 90 & 80 \\ \hline
whirlpool & 30 & 20 & 40 & 50 & 30 & 70 & 50\\ \hline \hline
\textbf{Avg. }(\%) & 36 & 22.31 & 31.54 & 42.31 & 34.62 & 37.69 & 40.77\\ \hline
\end{tabular}
\label{sceneclassresults1}
\end{center}
\end{table}

\begin{table}
\tiny
\begin{center}
\caption{Comparison of classification rates for various approaches on the Yupenn ``stabilized'' dynamic dataset (with $m = 3$).}
\begin{tabular}{| c || c | c | c | c | c | c | c |}
\hline
\textbf{Class} & \textbf{Chaos} \cite{movingvistas} & \textbf{Chaos} (our)$^2$ & \textbf{D1} & \textbf{D2} & \textbf{D3} & \textbf{DT1} & \textbf{DT2} \\ \hline \hline
beach & 27 & 17 & 77 & 80 & 77 & 83 & 77 \\ \hline
c. street & 17 & 70 & 3 & 87 & 90 & 100 & 93 \\ \hline
elevator & 40 & 17 & 7 & 37 & 10 & 23 & 17 \\ \hline
f. fire & 50 & 10 & 40 & 50 & 57 & 40 & 50 \\ \hline
fountain & 7 & 10 & 0 & 27 & 17 & 47 & 0 \\ \hline
highway & 17 & 17 & 77 & 47 & 53 & 33 & 60 \\ \hline
l. storm & 37 & 97 & 97 & 97 & 93 & 97 & 100 \\ \hline
ocean & 43 & 30 & 60 & 70 & 80 & 87 & 77\\ \hline
railway & 3 & 17 & 60 & 57 & 23 & 40 & 60\\ \hline
r. river & 3 & 87 & 60 & 90 & 83 & 87 & 77\\ \hline
sky & 33 & 23 & 30 & 47 & 43 & 50 & 57\\ \hline
snowing & 10 & 77 & 73 & 80 & 90 & 90 & 93\\ \hline
waterfall & 10 & 17 & 50 & 37 & 30 & 37 & 37\\ \hline
w. farm & 17 & 03 & 30 & 13 & 20 & 10 & 33\\ \hline \hline
\textbf{Avg.} (\%) & 22.43 & 35.14 & 48.64 & 58.50 & 54.71 & 58.85 & 59.35\\ \hline
\end{tabular}
\label{sceneclassresults2}
\end{center}
\end{table}

\subsection{Dynamic Scene Recognition}
Natural dynamic scene recognition has been gaining interest in recent years \cite{movingvistas,derpanis2012dynamic}. In an attempt to test the generality of the proposed framework to dynamical modeling for applications in video analysis, we evaluate its performance on dynamical scene classification. In this experiment, we use the Maryland ``in-the-wild'' dataset \cite{movingvistas} which is a collection of $13$ classes with $10$ examples per class and a larger Yupenn stabilized dynamic dataset \cite{derpanis2012dynamic} which is a collection of $14$ classes with $30$ examples per class. The former has videos collected from video hosting websites with no control over recording process leading to a dataset with large variations in illumination, view and scale \cite{movingvistas}. The latter dataset was recently released to emphasize only the scene-specific temporal information rather than camera-induced ones. In addition, the scene classes in the datasets were selected to illustrate potential failure of static scene representations leading to confusion between classes (e.g., chaotic traffic and smooth traffic). 

Recent research on dynamical modeling of scenes have shown that temporal (motion) information can provide better classification performance than traditional feature representations (e.g., GIST \cite{oliva2001modeling}) on static scenes \cite{movingvistas,derpanis2012dynamic}. The GIST feature is based on the hypothesis that humans recognize scenes by holistic understanding of a scene \cite{oliva2001modeling,biederman1987recognition}, thereby providing a global spatial representation of a scene. Shroff \textit{et al}. employed traditional chaotic invariants to model the dynamics in the time-series of the $960$-dimensional GIST descriptor extracted from each video and will be treated as our baseline. Similarly, we compare the performance of our proposed shape distribution features estimated on the $960$-dimensional GIST descriptor to further support our hypothesis that proposed shape-based features can perform better than traditional chaotic invariants in video-based inference tasks.

The average classification accuracy for all the proposed dynamical shape features in comparison with traditional chaotic invariants using a nearest neighbor classifier are tabulated in TABLE \ref{sceneclassresults1} and \ref{sceneclassresults2}. It is evident from these results that the proposed dynamical shape features (\textbf{D2} and \textbf{DT2}) perform better than the traditional chaotic invariants used in literature for dynamical scene classification. Evidently it is possible to improve classification performance further by fusion of dynamical and  spatial features as in \cite{movingvistas}, but here we restrict ourselves to comparison with core dynamical approaches.

\section{Conclusion and Future Directions}
In this paper, we have proposed a shape theoretic dynamical analysis framework for applications in action and gesture recognition, movement quality assessment for stroke rehabilitation and dynamical scene classification. We address the drawbacks of traditional measures from chaos theory for modeling the dynamics by proposing a framework combining the concepts of nonlinear time-series analysis and shape theory to extract robust and discriminative features from the reconstructed phase space. Our experiments on nonlinear dynamical models and joint trajectory data from motion capture support our hypothesis that the $shape$ of the reconstructed phase space can be used as feature representation for the above discussed applications. Furthermore, the wide range of experimental analysis on publicly available datasets for recognition of actions, gestures and scenes validate our claims. The framework was also tested on movement analysis on a finer scale, where we were interested in quantifying the \textit{movement quality} (level of impairment) for applications in stroke rehabilitation. Our experiments using a single marker indicate that with combination of dynamical features and machine learning tools, we are able to achieve comparable performance levels to a heavy marker-based system in movement quality assessment.

In this work, we perform phase space reconstruction on every dimension independently (univariate phase space reconstruction). Our future directions will be towards employing techniques for multi-variate phase space reconstruction \cite{cao1998dynamics}. It has been shown in \cite{basharat2009time} that multi-variate phase space reconstruction method provides better modeling than univariate phase space reconstruction, and hence lower error in predictions for human motion. We would also like to explore the use of approximate entropy \cite{pincus1991approximate}, a dynamical measure quantifying regularity in a time-series. The suggested number of data samples required for computation of approximate entropy is between $50$ and $5000$ \cite{pincus1991approximate}, which makes it more a suitable feature representation for applications in video-based inferences.

\section{Acknowledgments}
This work was supported by the National Science Foundation (NSF) CAREER grant 1452163.

{
\bibliographystyle{IEEEtran}
\bibliography{bare_jrnl_compsoc}

\begin{thebibliography}{10}
\providecommand{\url}[1]{#1}
\csname url@samestyle\endcsname
\providecommand{\newblock}{\relax}
\providecommand{\bibinfo}[2]{#2}
\providecommand{\BIBentrySTDinterwordspacing}{\spaceskip=0pt\relax}
\providecommand{\BIBentryALTinterwordstretchfactor}{4}
\providecommand{\BIBentryALTinterwordspacing}{\spaceskip=\fontdimen2\font plus
\BIBentryALTinterwordstretchfactor\fontdimen3\font minus
  \fontdimen4\font\relax}
\providecommand{\BIBforeignlanguage}[2]{{%
\expandafter\ifx\csname l@#1\endcsname\relax
\typeout{** WARNING: IEEEtran.bst: No hyphenation pattern has been}%
\typeout{** loaded for the language `#1'. Using the pattern for}%
\typeout{** the default language instead.}%
\else
\language=\csname l@#1\endcsname
\fi
#2}}
\providecommand{\BIBdecl}{\relax}
\BIBdecl

\bibitem{venkataraman2013attractor}
V.~Venkataraman, P.~Turaga, N.~Lehrer, M.~Baran, T.~Rikakis, and S.~L. Wolf,
  ``Attractor-shape for dynamical analysis of human movement: Applications in
  stroke rehabilitation and action recognition,'' in \emph{IEEE Conference on
  Computer Vision and Pattern Recognition Workshops}, June 2013, pp. 514--520.

\bibitem{aggarwal2011human}
J.~Aggarwal and M.~S. Ryoo, ``Human activity analysis: A review,'' \emph{ACM
  Computing Surveys (CSUR)}, vol.~43, no.~3, p.~16, 2011.

\bibitem{movingvistas}
N.~Shroff, P.~Turaga, and R.~Chellappa, ``Moving vistas: Exploiting motion for
  describing scenes,'' in \emph{IEEE Conference on Computer Vision and Pattern
  Recognition}, June 2010, pp. 1911--1918.

\bibitem{stergiou2011human}
N.~Stergiou and L.~M. Decker, ``Human movement variability, nonlinear dynamics,
  and pathology: is there a connection?'' \emph{Human Movement Science},
  vol.~30, no.~5, pp. 869--888, 2011.

\bibitem{ali2007chaotic}
S.~Ali, A.~Basharat, and M.~Shah, ``Chaotic invariants for human action
  recognition,'' in \emph{IEEE International Conference on Computer Vision},
  Oct. 2007, pp. 1--8.

\bibitem{junejo2011view}
I.~N. Junejo, E.~Dexter, I.~Laptev, and P.~P{\'e}rez, ``View-independent action
  recognition from temporal self-similarities,'' \emph{IEEE Transactions on
  Pattern Analysis and Machine Intelligence}, vol.~33, no.~1, pp. 172--185,
  2011.

\bibitem{perc2005dynamics}
M.~Perc, ``The dynamics of human gait,'' \emph{European journal of physics},
  vol.~26, no.~3, pp. 525--534, 2005.

\bibitem{dingwell2000nonlinear}
J.~B. Dingwell and J.~P. Cusumano, ``Nonlinear time series analysis of normal
  and pathological human walking,'' \emph{Chaos: An Interdisciplinary Journal
  of Nonlinear Science}, vol.~10, no.~4, pp. 848--863, 2000.

\bibitem{dingwell2007differences}
J.~B. Dingwell and H.~G. Kang, ``Differences between local and orbital dynamic
  stability during human walking,'' \emph{Journal of Biomechanical
  Engineering}, vol. 129, no.~4, pp. 586--593, 2007.

\bibitem{harbourne2009movement}
R.~T. Harbourne and N.~Stergiou, ``Movement variability and the use of
  nonlinear tools: principles to guide physical therapist practice,''
  \emph{Physical Therapy}, vol.~89, no.~3, pp. 267--282, 2009.

\bibitem{miller2006improved}
D.~J. Miller, N.~Stergiou, and M.~J. Kurz, ``An improved surrogate method for
  detecting the presence of chaos in gait,'' \emph{Journal of biomechanics},
  vol.~39, no.~15, pp. 2873--2876, 2006.

\bibitem{ralaivola2003dynamical}
L.~Ralaivola, F.~d'Alch{\'e} Buc \emph{et~al.}, ``Dynamical modeling with
  kernels for nonlinear time series prediction,'' in \emph{Neural Information
  Processing Systems}, vol.~4, 2003, pp. 129--136.

\bibitem{bissacco2001recognition}
A.~Bissacco, A.~Chiuso, Y.~Ma, and S.~Soatto, ``Recognition of human gaits,''
  in \emph{IEEE Conference on Computer Vision and Pattern Recognition}, 2001,
  pp. 52--57.

\bibitem{wolf1985determining}
A.~Wolf, J.~B. Swift, H.~L. Swinney, and J.~A. Vastano, ``Determining lyapunov
  exponents from a time series,'' \emph{Physica D: Nonlinear Phenomena},
  vol.~16, no.~3, pp. 285--317, 1985.

\bibitem{eckmann1985ergodic}
J.-P. Eckmann and D.~Ruelle, ``Ergodic theory of chaos and strange
  attractors,'' \emph{Reviews of modern physics}, vol.~57, no.~3, pp. 617--656,
  1985.

\bibitem{sano1985measurement}
M.~Sano and Y.~Sawada, ``Measurement of the lyapunov spectrum from a chaotic
  time series,'' \emph{Physical review letters}, vol.~55, no.~10, pp.
  1082--1085, 1985.

\bibitem{farmer1987predicting}
J.~D. Farmer and J.~J. Sidorowich, ``Predicting chaotic time series,''
  \emph{Physical review letters}, vol.~59, no.~8, pp. 845--848, 1987.

\bibitem{lyaprosen}
M.~Rosenstein, J.~Collins, and C.~De~Luca, ``A practical method for calculating
  largest lyapunov exponents from small data sets,'' \emph{Physica D: Nonlinear
  Phenomena}, vol.~65, no.~1, pp. 117--134, 1993.

\bibitem{tenbroek2007lyapunov}
T.~TenBroek, R.~Van~Emmerik, C.~Hasson, and J.~Hamill, ``Lyapunov exponent
  estimation for human gait acceleration signals,'' \emph{Journal of
  Biomechanics}, vol.~40, no.~2, p. 210, 2007.

\bibitem{iasemidis2003adaptive}
L.~D. Iasemidis, D.-S. Shiau, W.~Chaovalitwongse, J.~C. Sackellares, P.~M.
  Pardalos, J.~C. Principe, P.~R. Carney, A.~Prasad, B.~Veeramani, and
  K.~Tsakalis, ``Adaptive epileptic seizure prediction system,'' \emph{IEEE
  Transactions on Biomedical Engineering}, vol.~50, no.~5, pp. 616--627, 2003.

\bibitem{gavrila1999visual}
D.~M. Gavrila, ``The visual analysis of human movement: A survey,''
  \emph{Computer Vision and Image Understanding}, vol.~73, no.~1, pp. 82--98,
  1999.

\bibitem{rabiner1989tutorial}
L.~Rabiner, ``A tutorial on hidden markov models and selected applications in
  speech recognition,'' \emph{Proceedings of the IEEE}, vol.~77, no.~2, pp.
  257--286, 1989.

\bibitem{casti1986linear}
J.~L. Casti, \emph{Linear Dynamical Systems}.\hskip 1em plus 0.5em minus
  0.4em\relax Academic Press Professional, Inc., 1986.

\bibitem{yamato1992recognizing}
J.~Yamato, J.~Ohya, and K.~Ishii, ``Recognizing human action in time-sequential
  images using hidden markov model,'' in \emph{IEEE Conference on Computer
  Vision and Pattern Recognition}, June 1992, pp. 379--385.

\bibitem{wilson1995learning}
A.~D. Wilson and A.~F. Bobick, ``Learning visual behavior for gesture
  analysis,'' in \emph{IEEE International Symposium on Computer Vision}, Nov.
  1995, pp. 229--234.

\bibitem{vaswani2005shape}
N.~Vaswani, A.~K. Roy-Chowdhury, and R.~Chellappa, ``Shape activity: a
  continuous-state hmm for moving/deforming shapes with application to abnormal
  activity detection,'' \emph{IEEE Transactions on Image Processing}, vol.~14,
  no.~10, pp. 1603--1616, 2005.

\bibitem{cuntoor2007epitomic}
N.~P. Cuntoor and R.~Chellappa, ``Epitomic representation of human
  activities,'' in \emph{IEEE Conference on Computer Vision and Pattern
  Recognition}, June 2007, pp. 1--8.

\bibitem{kale2004identification}
A.~Kale, A.~Sundaresan, A.~Rajagopalan, N.~P. Cuntoor, A.~K. Roy-Chowdhury,
  V.~Kruger, and R.~Chellappa, ``Identification of humans using gait,''
  \emph{IEEE Transactions on Image Processing}, vol.~13, no.~9, pp. 1163--1173,
  2004.

\bibitem{liu2006improved}
Z.~Liu and S.~Sarkar, ``Improved gait recognition by gait dynamics
  normalization,'' \emph{IEEE Transactions on Pattern Analysis and Machine
  Intelligence}, vol.~28, no.~6, pp. 863--876, 2006.

\bibitem{bregler1997learning}
C.~Bregler, ``Learning and recognizing human dynamics in video sequences,'' in
  \emph{IEEE Conference on Computer Vision and Pattern Recognition}, June 1997,
  pp. 568--574.

\bibitem{chen2011computational}
Y.~Chen, M.~Duff, N.~Lehrer, H.~Sundaram, J.~He, S.~L. Wolf, and T.~Rikakis,
  ``A computational framework for quantitative evaluation of movement during
  rehabilitation,'' in \emph{AIP Conference Proceedings-American Institute of
  Physics}, vol. 1371, 2011, pp. 317--326.

\bibitem{fugl1975post}
A.~Fugl-Meyer, L.~J{\"a}{\"a}sk{\"o}, I.~Leyman, S.~Olsson, S.~Steglind
  \emph{et~al.}, ``The post-stroke hemiplegic patient. 1. a method for
  evaluation of physical performance.'' \emph{Scandinavian journal of
  rehabilitation medicine}, vol.~7, no.~1, pp. 13--31, 1975.

\bibitem{wolf2001assessing}
S.~L. Wolf, P.~A. Catlin, M.~Ellis, A.~L. Archer, B.~Morgan, and A.~Piacentino,
  ``Assessing wolf motor function test as outcome measure for research in
  patients after stroke,'' \emph{Stroke}, vol.~32, no.~7, pp. 1635--1639, 2001.

\bibitem{fei2005bayesian}
L.~Fei-Fei and P.~Perona, ``A bayesian hierarchical model for learning natural
  scene categories,'' in \emph{IEEE Conference on Computer Vision and Pattern
  Recognition}, June 2005, pp. 524--531.

\bibitem{xiao2010sun}
J.~Xiao, J.~Hays, K.~A. Ehinger, A.~Oliva, and A.~Torralba, ``Sun database:
  Large-scale scene recognition from abbey to zoo,'' in \emph{IEEE Conference
  on Computer Vision and Pattern Recognition}, June 2010, pp. 3485--3492.

\bibitem{oliva2006building}
A.~Oliva and A.~Torralba, ``Building the gist of a scene: The role of global
  image features in recognition,'' \emph{Progress in brain research}, vol. 155,
  pp. 23--36, 2006.

\bibitem{oliva2001modeling}
A.~Oliva and Torralba, ``Modeling the shape of the scene: A holistic
  representation of the spatial envelope,'' \emph{International Journal of
  Computer Vision}, vol.~42, no.~3, pp. 145--175, 2001.

\bibitem{soatto2001dynamic}
S.~Soatto, G.~Doretto, and Y.~N. Wu, ``Dynamic textures,'' in \emph{IEEE
  International Conference on Computer Vision}, vol.~2, 2001, pp. 439--446.

\bibitem{doretto2003dynamic}
G.~Doretto, A.~Chiuso, Y.~N. Wu, and S.~Soatto, ``Dynamic textures,''
  \emph{International Journal of Computer Vision}, vol.~51, no.~2, pp. 91--109,
  2003.

\bibitem{derpanis2012dynamic}
K.~G. Derpanis, M.~Lecce, K.~Daniilidis, and R.~P. Wildes, ``Dynamic scene
  understanding: The role of orientation features in space and time in scene
  classification,'' in \emph{IEEE Conference on Computer Vision and Pattern
  Recognition}, June 2012, pp. 1306--1313.

\bibitem{osada2002shape}
R.~Osada, T.~Funkhouser, B.~Chazelle, and D.~Dobkin, ``Shape distributions,''
  \emph{ACM Transactions on Graphics}, vol.~21, no.~4, pp. 807--832, 2002.

\bibitem{bissacco2005modeling}
A.~Bissacco, ``Modeling and learning contact dynamics in human motion,'' in
  \emph{IEEE Conference on Computer Vision and Pattern Recognition}, June 2005,
  pp. 421--428.

\bibitem{williams1997chaos}
G.~P. Williams, \emph{Chaos theory tamed}.\hskip 1em plus 0.5em minus
  0.4em\relax Joseph Henry Press, 1997.

\bibitem{abarbanel1996analysis}
H.~D. Abarbanel, \emph{Analysis of observed chaotic data}.\hskip 1em plus 0.5em
  minus 0.4em\relax New York: Springer-Verlag, 1996.

\bibitem{Takens}
F.~Takens, ``Detecting strange attractors in turbulence,'' \emph{Dynamical
  Systems and Turbulence}, vol. 898, pp. 366--381, 1981.

\bibitem{kennel1992determining}
M.~B. Kennel, R.~Brown, and H.~D. Abarbanel, ``Determining embedding dimension
  for phase-space reconstruction using a geometrical construction,''
  \emph{Physical review A}, vol.~45, no.~6, p. 3403, 1992.

\bibitem{small2005applied}
M.~Small, \emph{Applied nonlinear time series analysis: applications in
  physics, physiology and finance}.\hskip 1em plus 0.5em minus 0.4em\relax
  World Scientific Publishing Company Incorporated, 2005, vol.~52.

\bibitem{tucker1999lorenz}
W.~Tucker, ``The lorenz attractor exists,'' \emph{Comptes Rendus de
  l'Acad{\'e}mie des Sciences-Series I-Mathematics}, vol. 328, no.~12, pp.
  1197--1202, 1999.

\bibitem{bhattacharyya1943measure}
A.~Bhattacharyya, ``On a measure of divergence between two statistical
  populations defined by their probability distributions,'' \emph{Indian
  Journal of Statistics}, vol.~35, no. 99-109, p.~4, 1943.

\bibitem{srivastava2007riemannian}
A.~Srivastava, I.~Jermyn, and S.~Joshi, ``Riemannian analysis of probability
  density functions with applications in vision,'' in \emph{IEEE Conference on
  Computer Vision and Pattern Recognition}, June 2007, pp. 1--8.

\bibitem{rubner1998metric}
Y.~Rubner, C.~Tomasi, and L.~J. Guibas, ``A metric for distributions with
  applications to image databases,'' in \emph{IEEE International Conference on
  Computer Vision}, Jan. 1998, pp. 59--66.

\bibitem{li2010action}
W.~Li, Z.~Zhang, and Z.~Liu, ``Action recognition based on a bag of 3d
  points,'' in \emph{IEEE Conference on Computer Vision and Pattern Recognition
  Workshops}, Jun. 2010, pp. 9--14.

\bibitem{baran2011design}
M.~Baran, N.~Lehrer, D.~Siwiak, Y.~Chen, M.~Duff, T.~Ingalls, and T.~Rikakis,
  ``Design of a home-based adaptive mixed reality rehabilitation system for
  stroke survivors,'' in \emph{IEEE Conference on Engineering in Medicine and
  Biological Society}, Aug. 2011, pp. 7602--7605.

\bibitem{biederman1987recognition}
I.~Biederman, ``Recognition-by-components: a theory of human image
  understanding,'' \emph{Psychological review}, vol.~94, no.~2, pp. 115--147,
  1987.

\bibitem{cao1998dynamics}
L.~Cao, A.~Mees, and K.~Judd, ``Dynamics from multivariate time series,''
  \emph{Physica D: Nonlinear Phenomena}, vol. 121, no.~1, pp. 75--88, 1998.

\bibitem{basharat2009time}
A.~Basharat and M.~Shah, ``Time series prediction by chaotic modeling of
  nonlinear dynamical systems,'' in \emph{IEEE International Conference on
  Computer Vision}, 2009, pp. 1941--1948.

\bibitem{pincus1991approximate}
S.~M. Pincus, ``Approximate entropy as a measure of system complexity.''
  \emph{Proceedings of the National Academy of Sciences}, vol.~88, no.~6, pp.
  2297--2301, 1991.

\end{thebibliography}
}

\begin{IEEEbiography}
[{\includegraphics[width=1in,height=1.25in,clip,keepaspectratio]{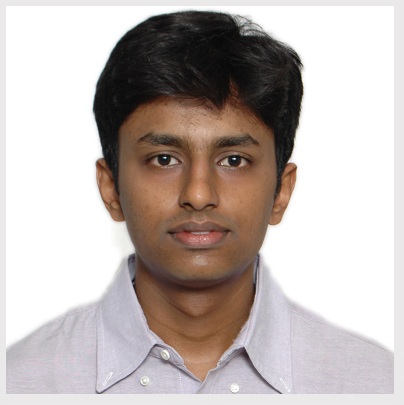}}]{Vinay Venkataraman}
received his M.S. degree in Electrical Engineering from Arizona State University in 2012. He is currently a doctoral student in the department of Electrical Engineering at Arizona State University. His research interests are in nonlinear dynamical analysis, computer vision and biomedical signal processing. He is a student member of IEEE. 
\end{IEEEbiography}

\begin{IEEEbiography}
[{\includegraphics[width=1in,height=1.25in,clip,keepaspectratio]{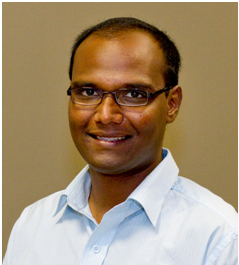}}]{Pavan Turaga}
(S'05, M'09, SM'14) is Assistant Professor in the School of Arts, Media, Engineering, and Electrical Engineering at Arizona State University. He received the B.Tech. degree in electronics and communication engineering from the Indian Institute of Technology Guwahati, India, in 2004, and the M.S. and Ph.D. degrees in electrical engineering from the University of Maryland, College Park in 2008 and 2009 respectively. He then spent two years as a research associate at the Center for Automation Research, University of Maryland, College Park. His research interests are in computer vision and computational imaging with applications in activity analysis, and dynamic scene analysis, with a focus on non-Euclidean techniques for these applications. He was awarded the Distinguished Dissertation Fellowship in 2009. He was selected to participate in the Emerging Leaders in Multimedia Workshop by IBM, New York, in 2008. He received the National Science Foundation CAREER award in 2015.

\end{IEEEbiography}

%
%
%

\end{document}